%% file: iclr2027_conference.tex
\documentclass{article} 
\usepackage{iclr2027_conference,times}

\input{math_commands.tex}

\usepackage{hyperref}
\hypersetup{hidelinks}
\usepackage{url}
\usepackage{graphicx}
\usepackage{subcaption}
\usepackage{multirow}
\usepackage{enumitem}
\usepackage{booktabs} 
\usepackage{wrapfig}
\usepackage{float}
\usepackage{makecell}
\usepackage{colortbl}
\usepackage{xcolor}
\definecolor{projlink}{RGB}{200,30,120}

\newcommand{\blfootnote}[1]{%
  \begingroup
  \renewcommand\thefootnote{}\footnote{#1}%
  \addtocounter{footnote}{-1}%
  \endgroup
}

\title{GRACE: Generation-Aware Latent\\ Compression for Efficient Video Generation}
\author{%
Jiyoung Kim$^{1}$ \quad Paul Hyunbin Cho$^{1}$ \quad Jisu Nam$^{1}$ \quad Donghoon Lee$^{2}$ \\
\bf Hyunsung Go$^{2}$ \quad Yeonkyeong Lee$^{2}$ \quad Hansaem Kim$^{2}$ \quad Seungryong Kim$^{1}$ \\[2pt]
$^{1}$KAIST AI \qquad $^{2}$Kakao Corp. \\[2pt]
Project page: \textcolor{projlink}{\url{https://cvlab-kaist.github.io/GRACE/}} \\
}

\iclrfinalcopy 
\begin{document}

\maketitle
\blfootnote{This work was done while the first three authors were interns at Kakao Corp.}

\input{sec/0_Abstract}
\input{sec/1_Introduction}

\input{sec/2_RelatedWorks} 
\input{sec/3_Preliminaries} 
\input{sec/4_Method}

\input{sec/5_Experiments} 
\input{sec/6_Conclusion}

\subsection*{AI use statement}

In this work, we used generative AI tools for writing assistance, including polishing prose, shortening captions, and formatting results into LaTeX tables, and for assistance with implementing our method. We have not used generative AI tools for research ideation, methodology design, or experiment design. We used GPT-4o to expand the short VBench-I2V prompts into the evaluation prompts, as described in Appendix~\ref{app:protocol}; formulating mathematical claims and assisting with proofs are not applicable to this work. Additionally, we used generative AI tools to search for related work, whose relevance and content we verified by reading the cited papers ourselves. We have reviewed all AI-assisted work: AI-assisted code was read and tested by the authors, every number reported in this paper was produced by our own experiments and transcribed into the tables by the authors, and all AI-assisted text was rewritten or approved by the authors. We take responsibility for the final content of this work, including text, claims or artifacts produced with the aid of generative AI.





\subsection*{Reproducibility statement}

The architecture and training setup of our autoencoder are described in Appendix~\ref{app:ae_details}, and those of the DiT adaptation in Appendix~\ref{app:diffusion_details}, with all hyperparameters listed in the corresponding tables. Appendix~\ref{app:eval_details} describes how each compared autoencoder is adapted to the pretrained DiT and how every model is evaluated, including the sampling configuration of each model and the protocol used to measure latency.

\bibliography{iclr2027_conference}
\bibliographystyle{iclr2027_conference}

\input{sec/X_suppl}

\end{document}

%% file: math_commands.tex
\usepackage{amsmath,amsfonts,bm}

\def\eqref#1{equation~\ref{#1}}

\def\1{\bm{1}}

\DeclareMathAlphabet{\mathsfit}{\encodingdefault}{\sfdefault}{m}{sl}
\SetMathAlphabet{\mathsfit}{bold}{\encodingdefault}{\sfdefault}{bx}{n}



%% file: sec/0_Abstract.tex
\input{figures/1_teaser}
\begin{abstract}
Highly compressed video autoencoders offer an effective way to accelerate video diffusion models, since the Diffusion Transformer (DiT) then operates on far fewer tokens. However, such autoencoders are difficult to obtain: a higher compression ratio degrades reconstruction quality, and recovering it requires more latent channels, which is known to slow DiT convergence. Moreover, the compressed latent space differs from the one the DiT was trained on, so the pretrained DiT must either be retrained from scratch or adapted at considerable cost. Further compressing the autoencoder that the DiT was trained with may seem to preserve compatibility, yet optimizing it for reconstruction alone still shifts the latent away from the distribution the DiT has learned. To address this, we propose \textbf{Generation-Aware Latent Compression for Efficient Video Generation (GRACE)}, a two-stage framework that compresses a pretrained video autoencoder while keeping it compatible with the pretrained DiT. In the first stage, we retain the base latent produced by the frozen pretrained encoder and learn a residual latent that captures the information lost under stronger compression. We further align the compressed latent with the pretrained latent in the feature space of the frozen DiT, so that the autoencoder is optimized for generation rather than reconstruction alone. In the second stage, we adapt the DiT through lightweight fine-tuning and asymmetric denoising, in which the base latent is denoised ahead of the residual latent. GRACE reduces the token count of Wan2.1-I2V-14B by nearly {\boldmath$8\times$} and its latency by {\boldmath$11.1\times$} at $480{\times}832$ resolution with 81 frames, while matching the VBench generation quality of the pretrained pipeline before compression.
\end{abstract}

%% file: figures/1_teaser.tex
\begin{figure}[htbp]
    \centering
    \includegraphics[width=\linewidth]{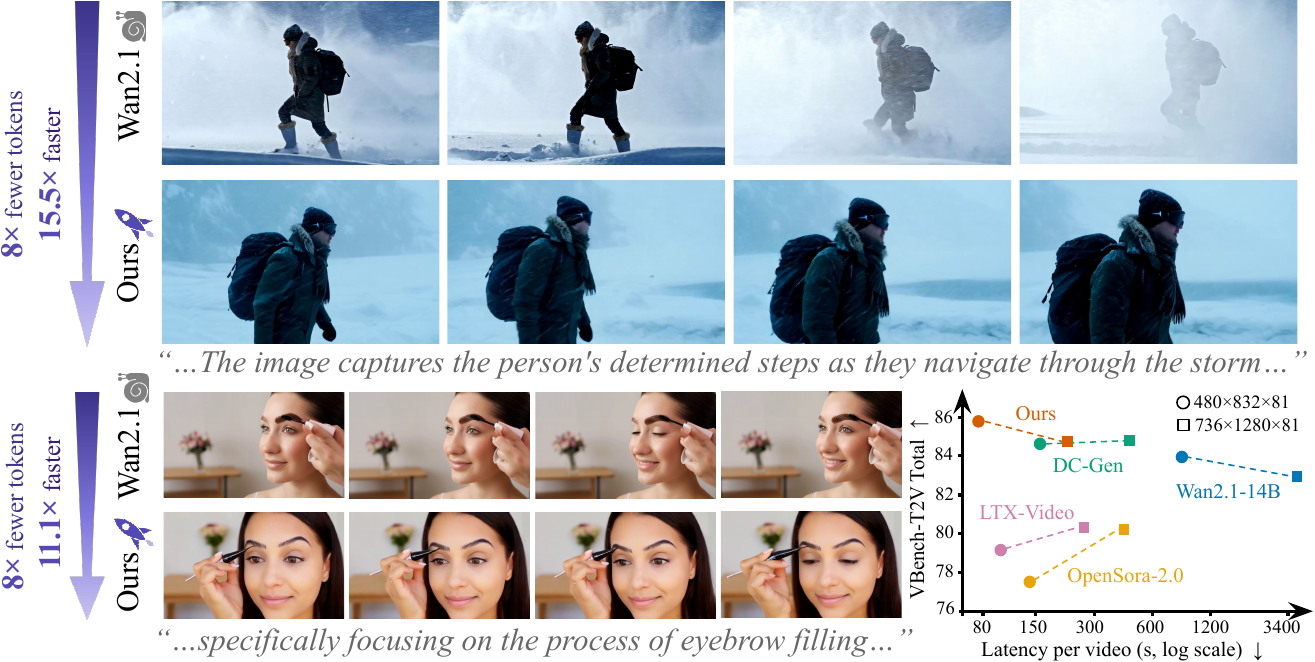}
    \caption{\textbf{Teaser.} We present \textbf{GRACE}, a novel latent compression technique that fine-tunes pretrained video generation models, such as Wan2.1-14B~\citep{wan2025wanopenadvancedlargescale}, to generate from substantially fewer latent tokens (e.g., nearly {\boldmath$8\times$} fewer). \textbf{(Top)} Text-to-video (T2V) samples at 736p from Wan2.1-14B before compression and from GRACE, using the same prompt. \textbf{(Bottom left)} The same comparison at 480p. \textbf{(Bottom right)} On VBench-T2V~\citep{huang2023vbenchcomprehensivebenchmarksuite, huang2024vbenchcomprehensiveversatilebenchmark}, GRACE preserves the generation quality of Wan2.1-14B before compression and achieves higher scores than existing high-compression autoencoders~\citep{hacohen2024ltxvideorealtimevideolatent, zheng2026opensora20trainingcommerciallevel, he2026dc}, while generating {\boldmath$11.1\times$} faster at 480p and {\boldmath$15.5\times$} faster at 736p. All models in the plot are evaluated at matched resolutions with 50 sampling steps. The plot reports text-to-video, and the speedups above are measured on image-to-video. }
\label{fig:teaser}
\end{figure}

%% file: sec/1_Introduction.tex
\section{Introduction}
\label{sec:intro}
Recent advances in video diffusion models~\citep{wan2025wanopenadvancedlargescale, ma2025stepvideot2vtechnicalreportpractice, hacohen2024ltxvideorealtimevideolatent} have demonstrated remarkable generation quality, yet their computational cost remains a significant bottleneck for both training and inference. This computational cost largely depends on the number of tokens processed by the Diffusion Transformer (DiT)~\citep{peebles2023scalablediffusionmodelstransformers}, whose attention scales quadratically with sequence length.

The number of input tokens scales with the spatio-temporal resolution of the latent produced by the video autoencoder~\citep{wan2025wanopenadvancedlargescale, hacohen2024ltxvideorealtimevideolatent}. Therefore, increasing the autoencoder compression ratio provides a straightforward way to reduce computation. However, fewer tokens limit the representational capacity of the latent, leading to a significant drop in reconstruction quality~\citep{yao2025reconstructionvsgenerationtaming}.

To preserve reconstruction quality with fewer tokens, previous works~\citep{chen2025deepcompressionautoencoderefficient, chen2025dcae15acceleratingdiffusion, hacohen2024ltxvideorealtimevideolatent, zheng2026opensora20trainingcommerciallevel, ma2025stepvideot2vtechnicalreportpractice} increase the channel dimension of each token. However, higher-dimensional latents are known to compromise generation quality, introducing a trade-off between reconstruction and generation quality~\citep{yao2025reconstructionvsgenerationtaming}, and they converge more slowly, raising the training cost. Moreover, modifying the autoencoder architecture typically requires retraining the autoencoder from scratch, making it difficult to fully leverage pretrained knowledge and incurring substantial training cost. Reusing a released high-compression autoencoder avoids such cost but passes the burden to the pretrained DiT, which must adapt to a latent space never seen during pretraining and may not fully converge even on 160 GPUs~\citep{zheng2026opensora20trainingcommerciallevel}. A natural alternative is to leverage a pretrained autoencoder by adding a lightweight compression block to its architecture. However, we empirically find that this approach still degrades generation quality (Tabs.~\ref{tab:recon} and~\ref{tab:ablation}). Optimizing the autoencoder and the additional compression block solely for reconstruction shifts the resulting latent distribution away from that learned by the pretrained DiT, degrading generation quality~\citep{he2026dc,zhao2024cvvaecompatiblevideovae}.

These observations point to two key challenges in efficient latent compression: (1) efficiently adapting pretrained autoencoder and DiT models while fully leveraging their pretrained knowledge, and (2) reducing the gap between reconstruction and generation quality introduced by compression. Our key insight is that a pretrained autoencoder and DiT already form a well-aligned generative pipeline, and this alignment can be exploited during compression. We therefore propose \textbf{Generation-Aware Latent Compression for Efficient Video Generation (GRACE)}, a training framework that compresses a pretrained video autoencoder while explicitly accounting for both reconstruction and generation quality.

First, to fully leverage the pretrained autoencoder, we introduce a dual-latent representation that separates the pretrained representation from the  information lost to compression. The base latent is obtained from the pretrained autoencoder using a spatially and temporally downsampled input, preserving the representation already learned by the pretrained autoencoder. The residual latent then captures the spatial detail and inter-frame motion that the low-resolution base latent cannot represent. This design allows the compressed autoencoder to retain the pretrained representation while using the residual latent only to complement what the downsampled input leaves out.

Second, we introduce asymmetric denoising to exploit this separation during generation. The base latent is denoised at an earlier timestep than the residual latent, allowing the denoised base to provide a stable anchor for the residual to recover the missing information. Both latents are denoised within a single pass, introducing no additional sampling cost.

Finally, to preserve compatibility with the pretrained DiT, we introduce generation-aware alignment, which matches the compressed autoencoder latent with the pretrained latent in the DiT's feature space. Rather than relying solely on conventional reconstruction objectives such as L1/L2 and perceptual losses~\citep{zhang2018unreasonableeffectivenessdeepfeatures}, we use intermediate DiT features to measure the distance between the compressed and pretrained latents. This generation-aware alignment encourages the compressed latent to remain compatible with the distribution learned by the pretrained DiT, thereby reducing the gap between reconstruction and generation quality.

Our method reduces the token count of Wan2.1-14B by nearly $\mathbf{8}\times$ and end-to-end generation latency by $\mathbf{11.1}\times$ at $480{\times}832{\times}81$, while matching the generation quality of the pretrained pipeline before compression on VBench~\citep{huang2023vbenchcomprehensivebenchmarksuite,huang2024vbenchcomprehensiveversatilebenchmark} (Fig.~\ref{fig:teaser}), compressing both axes at once to $16\times$ spatial and $8\times$ temporal. The gain grows with resolution, reaching $\mathbf{15.5}\times$ at $736{\times}1280{\times}81$. Compared with our single-latent baseline and existing high-compression video autoencoders on the same generation backbone, our method achieves better generation quality at the same or fewer tokens, even though its reconstruction on Panda-70M~\citep{chen2024panda70mcaptioning70mvideos} is not the best among them, showing that reconstruction fidelity alone does not determine generation quality.

We summarize our contributions as follows:
\begin{itemize}[leftmargin=*,itemsep=3pt,topsep=4pt,parsep=0pt,partopsep=0pt]
\item We propose GRACE, a diffusion training framework that efficiently compresses a pretrained video autoencoder while preserving the pretrained autoencoder--DiT pipeline and narrowing the reconstruction--generation gap.

\item We introduce a dual-latent representation with asymmetric denoising to preserve the pretrained autoencoder representation while recovering the spatial detail and inter-frame motion that the downsampled input leaves out.

\item We introduce a generation-aware alignment objective in the feature space of the pretrained DiT to preserve its compatibility with the compressed autoencoder.

\item Our method reduces the token count by nearly $8\times$, compressing both axes at once to $16\times$ spatial and $8\times$ temporal, while matching the generation quality of the pretrained pipeline before compression on VBench with an $11.1\times$ reduction in latency at $480{\times}832{\times}81$ and a $15.5\times$ reduction at $736{\times}1280{\times}81$.
\end{itemize}

%% file: sec/2_RelatedWorks.tex
\section{Related Work}
\label{sec:related}

\noindent\textbf{Latent compression for efficient video generation.}
Video diffusion models are computationally expensive since they repeatedly denoise a large number of latent tokens. Existing acceleration methods either reduce the number of denoising steps~\citep{song2022denoisingdiffusionimplicitmodels, lu2022dpmsolverfastodesolver, zhao2023unipcunifiedpredictorcorrectorframework, wang2024animatelcmcomputationefficientpersonalizedstyle, li2024t2vturbobreakingqualitybottleneck}, often at the cost of quality, or reduce the computation per step~\citep{xi2025sparsevideogenacceleratingvideo, sun2025asymrnrvideodiffusiontransformers}. Several works~\citep{hacohen2024ltxvideorealtimevideolatent, zheng2026opensora20trainingcommerciallevel, tian2025reduciogenerating1kvideo, chen2025dcvideogenefficientvideogeneration, ma2025stepvideot2vtechnicalreportpractice} instead propose to compress the latent beyond the common setting of $8\times$ spatial and $4\times$ temporal compression (f8t4), which significantly lowers the token count by a fixed ratio. Existing approaches increase spatial compression through autoencoder architectural changes~\citep{chen2025deepcompressionautoencoderefficient, tian2025reduciogenerating1kvideo}, or progressively increase temporal compression~\citep{mahapatra2025progressive}. A complementary strategy assigns more channels to each token to preserve reconstruction quality at high compression ratios, but wider latents can degrade generation quality, which recent work addresses by structuring the channel dimension~\citep{chen2025dcae15acceleratingdiffusion, cai2026davaepluginlatentcompression}. Our work compresses the video autoencoder in both space and time by fine-tuning a pretrained autoencoder and DiT rather than training either from scratch, keeping training cost low while mainly addressing the generation quality and convergence issues introduced by increased channel capacity.\vspace{+5pt}

\noindent\textbf{Autoencoder adaptation for pretrained generators.}
A few recent works redesign or improve the autoencoder in generation pipelines~\citep{zheng2025diffusiontransformersrepresentationautoencoders, chen2025dcae15acceleratingdiffusion, cai2026davaepluginlatentcompression}. One line of work regularizes the latent with a pretrained vision foundation model so that a diffusion model learns it more easily~\citep{yao2025reconstructionvsgenerationtaming}, but trains both the autoencoder and the generator from scratch. Changing the latent space in this way typically requires retraining the DiT from scratch at prohibitive cost, since the new latent distribution no longer matches the one on which the DiT was trained. To avoid this cost, other works reuse the pretrained DiT and close the resulting mismatch during training~\citep{zheng2026opensora20trainingcommerciallevel}. Some constrain the new latent to remain reconstructable by the pretrained decoder~\citep{zhao2024cvvaecompatiblevideovae}, while others align patch embeddings between the pretrained and modified pipelines after the latent space has already diverged~\citep{he2026dc}. We instead optimize the autoencoder latent with intermediate features from the frozen DiT, keeping it close to the representation space where the pretrained generator operates.

%% file: sec/3_Preliminaries.tex
\section{Preliminaries}
In this section, we briefly review latent video diffusion models, which consist of a video autoencoder that compresses videos into a latent space and a diffusion transformer that generates within that space. The autoencoder encodes a video into a latent and decodes a latent back into a video, while the diffusion transformer learns to denoise noisy latents.

\noindent\textbf{Video autoencoders.}
Given an input video $\mathbf{x} \in \mathbb{R}^{3\times(1+L)\times H\times W}$ with $1+L$ frames of height $H$ and width $W$, a causal video autoencoder encodes it as $\mathbf{z} = \mathcal{E}(\mathbf{x}) \in \mathbb{R}^{C\times(1+\frac{L}{t})\times\frac{H}{f}\times\frac{W}{f}}$, where $C$ is the number of latent channels, and $f$ and $t$ are the spatial and temporal compression factors. The decoder reconstructs $\hat{\mathbf{x}} = \mathcal{D}(\mathbf{z})$. Modern video autoencoders compress space and time jointly with 3D causal convolutions~\citep{wan2025wanopenadvancedlargescale, hacohen2024ltxvideorealtimevideolatent}, commonly at f8t4.

\noindent\textbf{Diffusion transformers.}
A diffusion transformer $\mathbf{v}_\theta$~\citep{peebles2023scalablediffusionmodelstransformers} generates in this latent space, patchifying the latent with a patch size $p$, so the token count is set by $f$, $t$, and $p$ together. Following the rectified flow formulation~\citep{esser2024scalingrectifiedflowtransformers}, a schedule time $u \in [0,1]$ is mapped to the flow matching timestep $\tau$ by a shift function $\psi_s$,
\begin{equation}
\label{eq:shift}
    \tau = \psi_s(u) = \frac{s\,u}{1 + (s-1)\,u} \> ,
\end{equation}
where larger $s$ places more of the schedule at higher noise. Writing $\mathbf{z}_0$ for the clean latent $\mathbf{z}$, the noisy latent is
\begin{equation}
\label{eq:add_noise}
    \mathbf{z}_\tau = (1-\tau) \mathbf{z}_0 + \tau \boldsymbol{\epsilon},
    \quad \boldsymbol{\epsilon} \sim \mathcal{N}(0, \mathbf{I}).
\end{equation}
A timestep embedding $\phi(\tau)$ is mapped by a projection $W_\text{mod}$ to the modulation $\mathbf{m}$ that conditions every transformer block through adaptive layer normalization (AdaLN). Conditioned on $\mathbf{m}$ and text embeddings $\mathbf{c}$, $\mathbf{v}_\theta$ predicts $(\boldsymbol{\epsilon} - \mathbf{z}_0)$:
\begin{equation}
\label{eq:flow_matching}
    \mathcal{L}_\text{velocity}
    = \mathbb{E}_{\mathbf{z}_0, \boldsymbol{\epsilon}, \tau}
    \|\mathbf{v}_\theta(\mathbf{z}_\tau, \tau, \mathbf{c})
    - (\boldsymbol{\epsilon} - \mathbf{z}_0)\|^2.
\end{equation}

%% file: sec/4_Method.tex
\section{Method}
\subsection{Overview}
\label{sec:overview}

We propose \textbf{GRACE}, a two-stage framework that compresses the latent of a pretrained pipeline while preserving generation quality (Fig.~\ref{fig:main_architecture}). \textbf{Stage 1} fine-tunes the autoencoder into a dual latent of a frozen base and a learned residual, guided by the frozen DiT (Section~\ref{sec:ae}). \textbf{Stage 2} adapts the DiT to that latent, denoising the base ahead of the residual (Section~\ref{sec:dit}). In both stages, the pretrained pipeline before compression guides its own compression.

\subsection{Stage 1: Video Autoencoder Training}
\label{sec:ae}

\input{figures/3_main_architecture}

We make full use of the pretrained autoencoder, keeping its architecture and weights, which keeps the latent close to the pretrained DiT's latent space and avoids training from scratch. The straightforward way to reach a higher ratio from there is to add compression blocks inside the encoder and train under the pretrained reconstruction objective $\mathcal{L}_\text{recon}$~\citep{wan2025wanopenadvancedlargescale}, a weighted sum of L1, LPIPS~\citep{zhang2018unreasonableeffectivenessdeepfeatures}, and KL~\citep{kingma2022autoencodingvariationalbayes} terms (Tab.~\ref{tab:hp_ae}). This already gives a strong baseline, reconstructing within the range of high-compression autoencoders trained from scratch (Tab.~\ref{tab:recon}). However, reconstruction quality does not carry over to generation, even with the pretrained initialization. The latent drifts to optimize the objective, leaving the DiT with more to adapt to in Stage 2. 

Stage 1 addresses this in two ways: (i) \emph{dual-latent representation}, which anchors the compressed latent in the pretrained latent space, and (ii) \emph{generation-aware alignment}, which matches the compressed latent to the pretrained latent inside the frozen DiT.

\noindent\textbf{Dual-latent representation.} We first design the compressed latent as a dual representation, with a base latent and a residual latent. The base latent, taken from the frozen pretrained encoder $\mathcal{E}$, anchors the compressed latent in the pretrained latent space, so the DiT has less to adapt to in Stage 2. Since $\mathcal{E}$ compresses only by $f$ and $t$, reaching the higher ratio requires reducing the input before encoding, as in prior work~\citep{mahapatra2025progressive,zhao2024cvvaecompatiblevideovae}: we downsample it spatially by $r_s$ and subsample it temporally by $r_t$, producing $\mathbf{x}_\text{low}$ within the resolution range $\mathcal{E}$ was trained on~\citep{wan2025wanopenadvancedlargescale}. Then $\mathcal{E}$ processes $\mathbf{x}_\text{low}$ with its original weights as
\begin{equation}
\mathbf{z}_\text{base} = \mathcal{E}(\mathbf{x}_\text{low}) \in \mathbb{R}^{C \times (1+\frac{L}{t \cdot r_t}) \times \frac{H}{f \cdot r_s} \times \frac{W}{f \cdot r_s}} \> .
\end{equation}

This reduced input lowers the token count, which limits the reconstruction capacity of $\mathbf{z}_\text{base}$~\citep{chen2025deepcompressionautoencoderefficient, chen2025dcae15acceleratingdiffusion}, since the decoder cannot restore what never reached the latent. We therefore introduce a residual encoder $\mathcal{E}_\text{res}$, initialized from the pretrained weights, that takes the video at full resolution and carries the missing information in $C'$ additional channels,
\begin{equation}
\mathbf{z}_\text{res} = \mathcal{E}_\text{res}(\mathbf{x}) \in \mathbb{R}^{C' \times (1+\frac{L}{t \cdot r_t}) \times \frac{H}{f \cdot r_s} \times \frac{W}{f \cdot r_s}} \> .
\end{equation}
To match the spatial and temporal size of $\mathbf{z}_\text{base}$, $\mathcal{E}_\text{res}$ includes an additional downsampling stage, paired with a parameter-free shortcut that folds space and time into channels~\citep{chen2025deepcompressionautoencoderefficient}. The two latents are concatenated along the channel axis into the compressed latent,

\begin{equation}
\mathbf{z} = [\mathbf{z}_\text{base}; \mathbf{z}_\text{res}] \in \mathbb{R}^{(C+C') \times (1+\frac{L}{t \cdot r_t}) \times \frac{H}{f \cdot r_s} \times \frac{W}{f \cdot r_s}} \> .
\end{equation}
Separating the two parts organizes the latent, which prior work finds to help generation at high latent dimensions~\citep{chen2025dcae15acceleratingdiffusion}. As shown in Fig.~\ref{fig:latent_analysis}, the dual latent distorts the body less than a single latent and keeps the colors natural, raising the total VBench score and the uniformity of the latent distribution. The decoder is fully fine-tuned, with the residual channels zero-initialized so that training starts at the pretrained reconstruction quality. For image-to-video (I2V), we also pass the first frame to the decoder to restore detail the compressed latent cannot carry. 

\input{figures/11_latent_analysis_merged}
\noindent\textbf{Generation-aware alignment in pretrained DiT feature space.}
\label{sec:alignment} The dual-latent representation alone narrows the gap between the compressed latent and the pretrained latent space, but we find it insufficient at higher compression ratios, where the residual latent must carry more information missing from the base. The residual channels are optimized only through the reconstruction objective, which improves fidelity but does not directly account for how easily the diffusion model can learn the resulting latent.  We introduce generation-aware alignment, a regularizer that supervises the compressed latent inside the feature space of the pretrained DiT without modifying the autoencoder architecture. Our approach builds on the same insight as prior autoencoder adaptation methods: reconstruction fidelity alone does not necessarily yield a compressed latent that the diffusion backbone can learn effectively~\citep{yao2025reconstructionvsgenerationtaming, chen2025dcae15acceleratingdiffusion}.
Since the pretrained DiT is already adapted to the original latent space, we compare the two latents in its intermediate feature space. We perturb both latents with a shared $\tau$ drawn from the pretrained pipeline's schedule and independently sampled $\boldsymbol{\epsilon}$. The alignment loss is then defined by passing both noised latents through this frozen DiT and matching their intermediate representations.
For the pretrained latent, the frozen encoder encodes the full-resolution video, and $\mathbf{v}_\theta$ patchifies the resulting latent through the original input projection $W_\text{in}$ before the transformer blocks:
\begin{equation}
    \mathbf{z}_\text{pre} = \mathcal{E}(\mathbf{x})
    \in \mathbb{R}^{C \times (1+\frac{L}{t}) \times \frac{H}{f} \times \frac{W}{f}},
    \qquad
    \mathbf{h}^l_\text{pre} = \mathbf{v}_\theta^l(\mathbf{z}_{\text{pre},\tau}, \tau, \mathbf{c}) \> ,
\end{equation}
where $\mathbf{c}$ is the caption embedding, $\mathbf{z}_{\text{pre},\tau}$ and $\mathbf{z}_\tau$ are the pretrained and compressed latents noised by Eq.~\ref{eq:add_noise}, and $\mathbf{v}_\theta^l$ is the $l$-th layer feature of $\mathbf{v}_\theta$. For image-to-video, we also pass the first frame to $\mathbf{v}_\theta$ as its image condition.
For the compressed latent $\mathbf{z}$, we use $\mathbf{v}_{\theta,+}$, which shares the frozen transformer body of $\mathbf{v}_\theta$ but extends the input projection along the channel axis with a zero-initialized $W^\text{res}_\text{in}$ to handle the residual channels.
Since this branch runs at a lower latent resolution, its features contain fewer tokens. We reshape them to their spatio-temporal grid, trilinearly upsample them to the full-resolution grid, and apply a zero-initialized per-layer residual projection $P_l$:
\begin{equation}
\mathbf{h}^l_\text{cmp} = \mathbf{v}_{\theta,+}^l(\mathbf{z}_\tau, \tau, \mathbf{c}) \> , \qquad
\hat{\mathbf{h}}^l_\text{cmp} = \text{interp}(\mathbf{h}^l_\text{cmp}) + P_l\big(\text{interp}(\mathbf{h}^l_\text{cmp})\big) \> .
\end{equation}
where $\text{interp}(\cdot)$ denotes this reshaping and trilinear upsampling.

We then compare the two features at each supervised layer:
\begin{equation}
\mathcal{L}_\text{align}
=
\sum_{l \in S}
\Big[
1 -
\text{sim}\big(
\mathbf{h}^l_\text{pre},
\hat{\mathbf{h}}^l_\text{cmp}
\big)
\Big] \> ,
\end{equation}
where $\text{sim}(\cdot, \cdot)$ is the cosine similarity averaged over tokens and $S$ is the set of supervised layers, the first 10 of the 40 DiT blocks (see Appendix~\ref{app:align} for this choice). The features from $\mathbf{z}_\text{pre}$ are kept fixed, so $\mathcal{L}_\text{align}$ only optimizes the compressed representation.

Since $\mathcal{L}_\text{align}$ and $\mathcal{L}_\text{recon}$ operate on different scales, a fixed weight makes training unstable. Following prior work~\citep{yao2025reconstructionvsgenerationtaming}, we set the weight as the ratio of their gradient norms with respect to the last convolutional layer of $\mathcal{E}_\text{res}$, so that the two losses contribute at a comparable scale without manual hyperparameter tuning. The final training objective is
\begin{equation}
\mathcal{L} = \mathcal{L}_\text{recon} + w_\text{adaptive}\,\mathcal{L}_\text{align} \> , \qquad
w_\text{adaptive} = \frac{\|\nabla \mathcal{L}_\text{recon}\|}{\|\nabla \mathcal{L}_\text{align}\|} \> .
\end{equation}
We analyze the resulting latent in Fig.~\ref{fig:latent_analysis} from two complementary perspectives. Following prior work that relates the uniformity of the latent distribution to generation quality~\citep{yao2025reconstructionvsgenerationtaming}, we fit a kernel density estimate to the t-SNE~\citep{JMLR:v9:vandermaaten08a} projection and measure its coefficient of variation, Gini coefficient, and normalized entropy (see Appendix~\ref{app:protocol} for details). All three improve with the alignment loss, and fewer artifacts remain in the samples, with the total VBench score following. These metrics summarize the latent as a whole, while the PCA visualization in Fig.~\ref{fig:latent_analysis} shows how it varies within and across frames: the dual latent makes the components less noisy, but they stay weak until the alignment loss makes them follow object regions and hold structure across frames.

\subsection{Stage 2: Video Diffusion Transformer Adaptation}
\label{sec:dit}
In this section, we adapt the pretrained DiT to the compressed latent while keeping the autoencoder frozen. Since the compressed latent adds $C'$ unseen channels, we extend the DiT input and output projections from $C$ to $C+C'$ channels. We fully fine-tune these projections and apply LoRA~\citep{hu2021loralowrankadaptationlarge} to the transformer blocks, keeping the adaptation cost low while preserving the pretrained generation capability~\citep{he2026dc}. The DiT is trained with the flow matching loss $\mathcal{L}_\text{velocity}$, applied to the base and the residual latent at different noise levels, which we describe next.

\noindent\textbf{Asymmetric denoising.}
\label{sec:base_ahead}
Our key idea is to denoise the two parts of the latent asymmetrically, keeping the base ahead of the residual so that the residual builds on a reliable foundation throughout generation. The residual encoder is trained to carry what the base does not, so what the residual should contain becomes clearer as the base settles, which makes the residual easier to generate.

\input{figures/7_offset_qual}
During training, we sample a base-ahead offset $\delta \sim \mathcal{U}(\delta_\text{min}, \delta_\text{max})$ (Tab.~\ref{tab:hp_dit}) and a schedule time $\hat{u} \sim \mathcal{U}(0, 1+\delta)$ per video, and set
\begin{equation}
\label{eq:offset}
u_\text{base} = \max(\hat{u} - \delta, 0) \> , \qquad u_\text{res} = \min(\hat{u}, 1) \> ,
\end{equation}
where $u = 1$ corresponds to pure noise and $u = 0$ to a clean latent, so the base is always the less corrupted of the two. We obtain the noise levels $\tau_\text{base}$ and $\tau_\text{res}$ from Eq.~\ref{eq:shift}, and noise the base and the residual separately with Eq.~\ref{eq:add_noise}:
\begin{equation}
\begin{split}
\mathbf{z}_\tau = \big[ &(1-\tau_\text{base}) \mathbf{z}_\text{base} + \tau_\text{base} \boldsymbol{\epsilon}_\text{base} \> ; \\
&(1-\tau_\text{res}) \mathbf{z}_\text{res} + \tau_\text{res} \boldsymbol{\epsilon}_\text{res} \big] \> ,
\end{split}
\end{equation}
where $\boldsymbol{\epsilon}_\text{base}, \boldsymbol{\epsilon}_\text{res} \sim \mathcal{N}(0, \mathbf{I})$. We sample $\delta$ rather than fixing it, which trains one model across offsets. The two timesteps share $\phi$ and enter through separate projections, which keeps the conditioning path closer to the pretrained path. We apply the pretrained $W_\text{mod}$ to the base and a zero-initialized $W_\text{mod}^\text{res}$ to the residual, so adaptation begins from the pretrained behavior,
\begin{equation}
\mathbf{m} = W_\text{mod}\,\phi(\tau_\text{base}) + W_\text{mod}^\text{res}\,\phi(\tau_\text{res}) \> .
\end{equation}
The modulation $\mathbf{m}$ conditions every transformer block, and the DiT predicts a velocity for each part,
\begin{equation}
[\hat{\mathbf{v}}_\text{base}, \hat{\mathbf{v}}_\text{res}] = \mathbf{v}_\theta(\mathbf{z}_\tau, [\tau_\text{base}, \tau_\text{res}], \mathbf{c}) \> ,
\end{equation}
where $\mathbf{v}_\theta$ here denotes the adapted DiT with the extended projections and LoRA, and the training objective is the average of the flow matching loss of Eq.~\ref{eq:flow_matching} over the base and the residual. At inference, we fix the offset to $\delta = 0.15$ and run the same number of sampling steps over $\hat{u} \in [0, 1+\delta]$, so the base leads the residual by $\delta$ throughout sampling (Appendix~\ref{app:diffusion_implementation}). Both parts are denoised in the same forward pass, rather than one after the other, so the number of function evaluations remains unchanged. Fig.~\ref{fig:denoising_order} shows that the asymmetric schedule recovers detail on the subject's face that a shared schedule degrades where motion is large.
\input{tables/0_recon_tab}

%% file: figures/3_main_architecture.tex
\begin{figure*}[t!]
  \centering
\includegraphics[width=\linewidth]{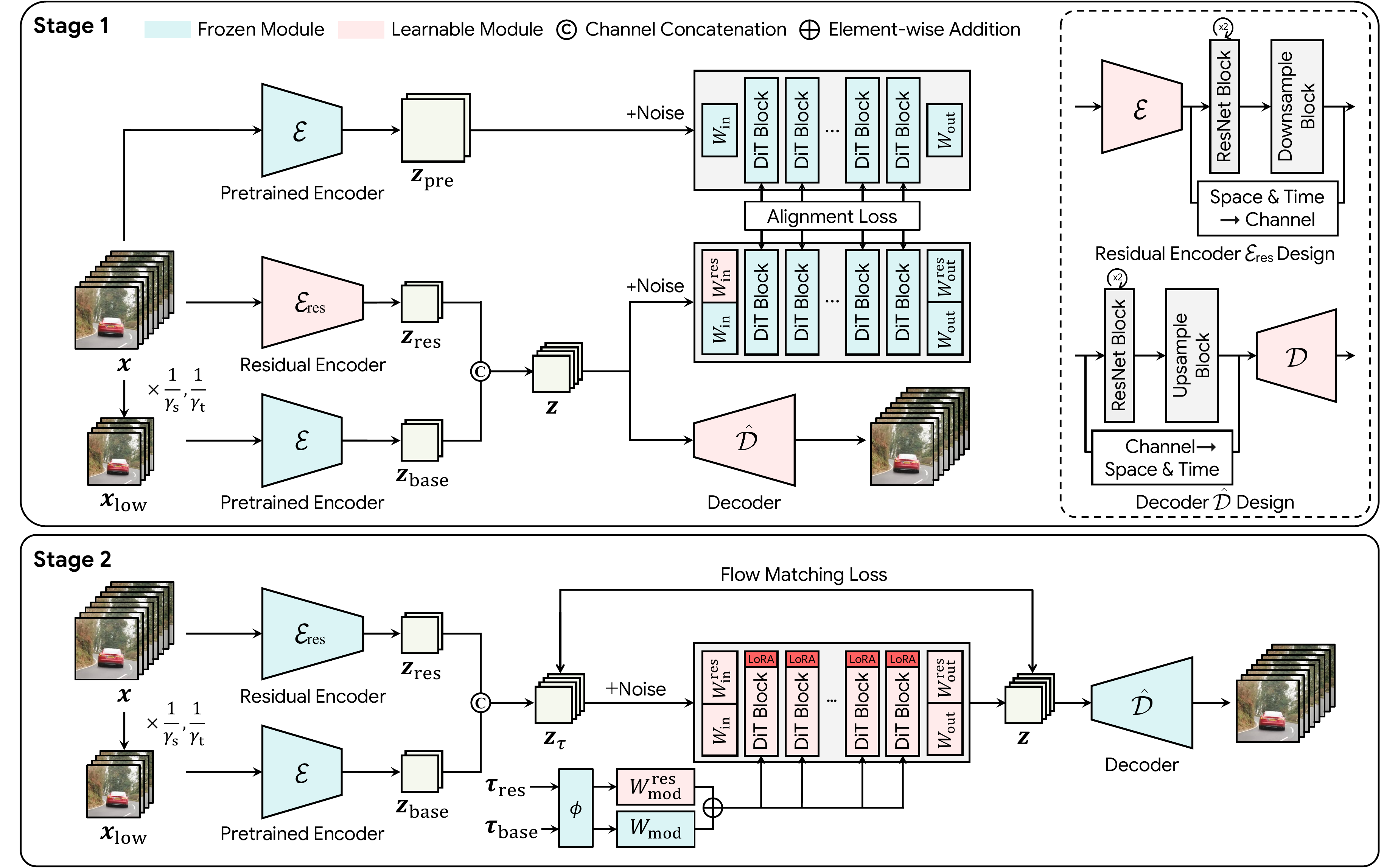}
\caption{\textbf{Overall architecture.} Stage 1 trains the autoencoder. The frozen encoder $\mathcal{E}$ maps the downsampled input to $\mathbf{z}_\text{base}$, the residual encoder $\mathcal{E}_\text{res}$ maps the full-resolution input to $\mathbf{z}_\text{res}$, and $\hat{\mathcal{D}}$ decodes both, while $\mathcal{L}_\text{align}$ matches the compressed latent to the pretrained latent inside the frozen DiT. The dashed box on the right of Stage 1 shows a simplified view of the blocks added to $\mathcal{E}_\text{res}$ and $\hat{\mathcal{D}}$. Stage 2 adapts the DiT to the compressed latent with LoRA, denoising $\mathbf{z}_\text{base}$ at a lower noise level than $\mathbf{z}_\text{res}$ at every step, so the base is denoised first.}
  \label{fig:main_architecture}
\end{figure*}

%% file: figures/11_latent_analysis_merged.tex
\begin{figure}[!t]
\centering
\includegraphics[width=\linewidth]{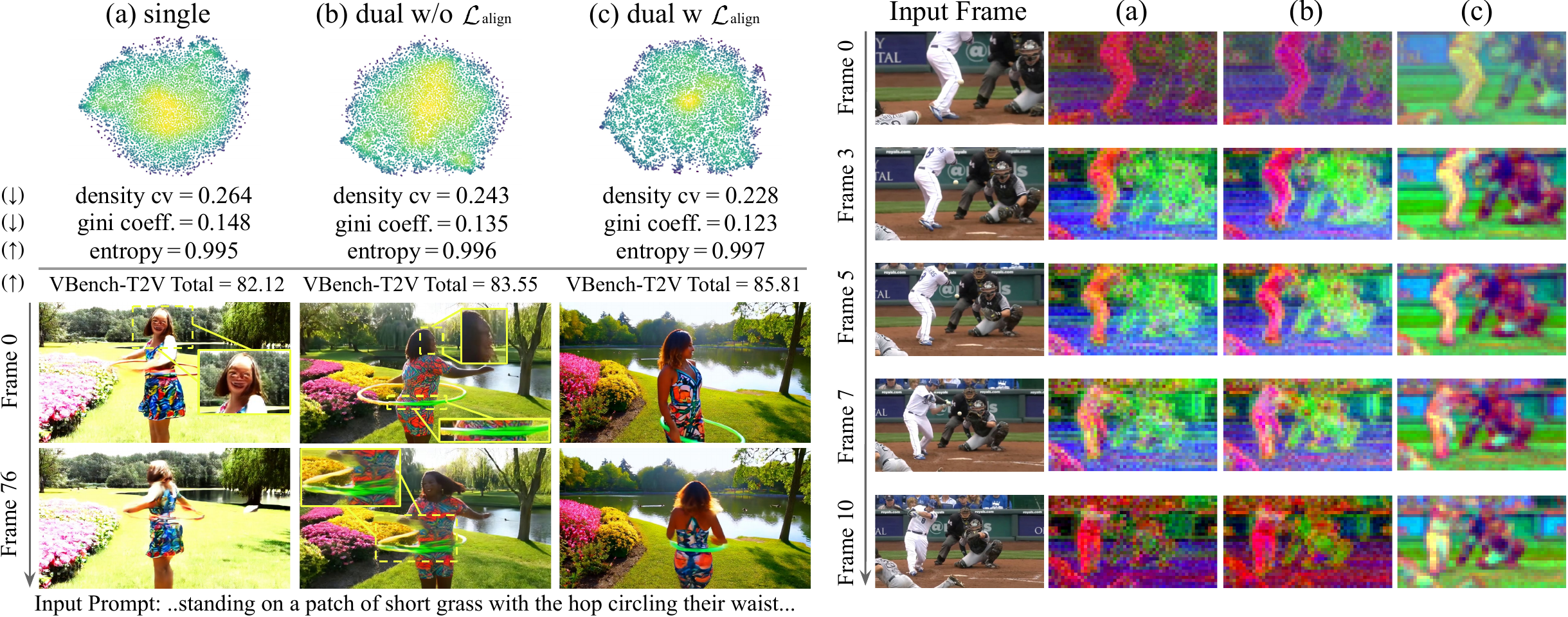}
\caption{\textbf{Effect of the dual latent and $\mathcal{L}_\text{align}$.} \textbf{(a)} single latent, \textbf{(b)} dual-latent representation without $\mathcal{L}_\text{align}$, and \textbf{(c)} ours, all generated at $480{\times}832{\times}81$. \textbf{(Left)} latent distribution, projected with t-SNE and colored by kernel density, with its uniformity metrics~\citep{yao2025reconstructionvsgenerationtaming} and the total VBench~\citep{huang2023vbenchcomprehensivebenchmarksuite, huang2024vbenchcomprehensiveversatilebenchmark} score below, followed by T2V samples generated after Stage 2. \textbf{(Right)} spatio-temporal structure of the latent, shown as its principal components mapped to RGB under the input frames. More samples are in Figs.~\ref{fig:pca_supp} and~\ref{fig:pca_supp_2}.}
\label{fig:latent_analysis}
\end{figure}


%% file: figures/7_offset_qual.tex
\begin{wrapfigure}{r}{0.5\linewidth}
  \centering
  \includegraphics[width=\linewidth]{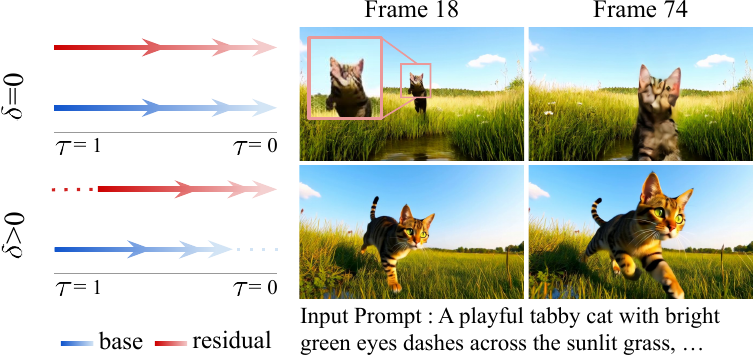}
  \caption{\textbf{Denoising order.} Each row shows the sampling steps of $\mathbf{z}_\text{base}$ and $\mathbf{z}_\text{res}$ (\textbf{Left}) and the generated frames (\textbf{Right}), with $\tau = 1$ pure noise and $\tau = 0$ clean. \textbf{(Top)} Both parts are denoised at the same noise level. \textbf{(Bottom)} $\mathbf{z}_\text{base}$ stays $\delta$ ahead of $\mathbf{z}_\text{res}$ in schedule time $u$ (Eq.~\ref{eq:offset}) at every step.}
\label{fig:denoising_order}
\end{wrapfigure}

%% file: tables/0_recon_tab.tex
\begin{table}[!t]
\centering
\small
\caption{\textbf{Video autoencoder comparison at $256{\times}256{\times}81$ reconstruction and $480{\times}832{\times}81$ generation.} We report the VBench-I2V~\citep{huang2023vbenchcomprehensivebenchmarksuite, huang2024vbenchcomprehensiveversatilebenchmark} total after adapting the same pretrained Wan2.1-I2V-14B~\citep{wan2025wanopenadvancedlargescale} to every latent under the same budget. \emph{Config} lists $f$, $t$, channel count $c$, and patch size $p$, which set the \emph{latent token} count. \emph{Single-latent Baseline} is our baseline without the dual latent or the alignment loss. The first row is the pretrained autoencoder before compression, and our method is \colorbox{gray!12}{shaded}. Bold marks the best value in each column among the last three rows (4.3k tokens). $^\dagger$Initialized by inflating its own 2D image autoencoder. $^\ddagger$Trained first at f8t4 and then extended to f16t8 with additional modules.}
\label{tab:recon}
\setlength{\tabcolsep}{4pt}
\resizebox{0.95\linewidth}{!}{%
\begin{tabular}{l c c c c c c c c}
\toprule
\multirow{2}{*}[-0.25\normalbaselineskip]{Autoencoder} & \multirow{2}{*}[-0.25\normalbaselineskip]{Config} & \multirow{2}{*}[-0.25\normalbaselineskip]{AE Training} & \multirow{2}{*}[-0.25\normalbaselineskip]{Latent Tokens} & \multicolumn{4}{c}{Reconstruction} & VBench-I2V \\
\cmidrule(lr){5-8} \cmidrule(lr){9-9}
 & & & & PSNR $\uparrow$ & SSIM $\uparrow$ & LPIPS $\downarrow$ & rFVD $\downarrow$ & Total $\uparrow$ \\
\midrule
Wan2.1-VAE~\citep{wan2025wanopenadvancedlargescale} & f8t4c16p2 & scratch$^\dagger$ & 32.8k & 35.15 & 0.958 & 0.016 & 1.13 & 87.92 \\
\arrayrulecolor{black!12}\midrule\arrayrulecolor{black}
Step-Video-VAE~\citep{ma2025stepvideot2vtechnicalreportpractice} & f16t8c64p1 & scratch$^\ddagger$ & 17.2k & 33.88 & 0.950 & 0.029 & 3.16 & 84.05 \\
Video DC-AE~\citep{zheng2026opensora20trainingcommerciallevel} & f32t4c128p1 & scratch & 8.2k & 34.61 & 0.956 & 0.024 & 3.70 & 84.94 \\
\arrayrulecolor{black!12}\midrule\arrayrulecolor{black}
LTX-VAE~\citep{hacohen2024ltxvideorealtimevideolatent} & f32t8c128p1 & scratch & 4.3k & 31.97 & 0.914 & 0.051 & 19.53 & 87.06 \\
Single-latent Baseline & f16t8c32p2 & fine-tuned & 4.3k & \textbf{33.76} & \textbf{0.956} & \textbf{0.031} & \textbf{13.11} & 86.44 \\
\rowcolor{gray!12}
\textbf{GRACE-VAE (Ours)} & f16t8c32p2 & fine-tuned & 4.3k & 32.63 & 0.930 & 0.032 & 13.53 & \textbf{87.90} \\
\bottomrule
\end{tabular}%
}
\end{table}

%% file: sec/5_Experiments.tex
\section{Experiments}
\label{sec:exp}

\subsection{Setup}
\label{sec:setup}
\noindent\textbf{Model configuration.}
We use Wan2.1~\citep{wan2025wanopenadvancedlargescale} as the pretrained pipeline before compression and compress its latent from f8t4p2 to f16t8p2 with $r_s{=}2$, $r_t{=}2$, and $C'{=}16$ residual channels. We refer to our autoencoder as GRACE-VAE, and to the full pipeline of the autoencoder and the adapted DiT as GRACE. We adapt Wan2.1-I2V-14B for image-to-video and Wan2.1-T2V-14B for text-to-video. Both stages are trained on Panda-70M~\citep{chen2024panda70mcaptioning70mvideos}. Full training and architecture details are in Appendices~\ref{app:ae_details} and~\ref{app:diffusion_details}.

\input{figures/5_gen_qual}
\noindent\textbf{Single-latent baseline.}
Under the same configuration above, our single-latent baseline only adds compression blocks to the pretrained autoencoder and fine-tunes it, without any further modification, matching our token and channel counts. The DiT is adapted following the training recipe of Wan2.1~\citep{wan2025wanopenadvancedlargescale}.

\input{figures/gen_main_quan_high_res}
\subsection{Video Reconstruction Results}
\label{sec:recon}

\noindent\textbf{Evaluation details.} We evaluate reconstruction on Panda-70M~\citep{chen2024panda70mcaptioning70mvideos} with PSNR, SSIM~\citep{1284395}, LPIPS~\citep{zhang2018unreasonableeffectivenessdeepfeatures}, and rFVD~\citep{unterthiner2019accurategenerativemodelsvideo}, and generation on VBench-I2V~\citep{huang2023vbenchcomprehensivebenchmarksuite,huang2024vbenchcomprehensiveversatilebenchmark}. We compare against publicly released video autoencoders that compress more aggressively than Wan2.1-VAE, with the pretrained pipeline before compression as the reference. The compared autoencoders are trained independently of Wan2.1-VAE, whereas our single-latent baseline and GRACE-VAE are fine-tuned from it. To measure generation, we adapt the pretrained Wan2.1-I2V-14B to every latent under the same total budget of 38.5 H200 GPU days. The compared latents spend the whole budget on DiT adaptation, while our pipeline spends 8.5 days on the autoencoder and 30 on the DiT. Our adaptation builds on the frozen Wan2.1 base latent, which is unavailable to the compared autoencoders, so we instead transfer them with the released implementation of DC-Gen~\citep{he2026dc,chen2025dcvideogenefficientvideogeneration}, which is designed to adapt a pretrained DiT to a new latent space (see Appendix~\ref{app:impl} for the adaptation details).

\noindent\textbf{Main results.} As shown in Tab.~\ref{tab:recon}, better reconstruction does not mean better generation. Step-Video-VAE reconstructs 1.91 dB above LTX-VAE but scores 3.01 lower on VBench-I2V, even at $4\times$ the tokens, and our single-latent baseline reconstructs 1.13 dB above GRACE-VAE yet scores 1.46 lower. GRACE-VAE reaches the highest generation quality among the compressed autoencoders at the smallest token count, within 0.02 of the pretrained pipeline, while Video DC-AE and Step-Video-VAE use $2\times$ and $4\times$ more tokens yet score 2.98 and 3.87 lower than the pretrained pipeline.

\subsection{Video Generation Results}
\label{sec:gen}
\noindent\textbf{Evaluation details.}
We evaluate on VBench~\citep{huang2023vbenchcomprehensivebenchmarksuite, huang2024vbenchcomprehensiveversatilebenchmark} for both I2V and T2V at $480{\times}832{\times}81$ with 50 sampling steps, generating one video per prompt over the full benchmark, and measure latency on a single A100 GPU. The models built on Wan2.1-14B share the same pretrained DiT, while LTX-Video and Open-Sora 2.0 use their own generators, which differ from Wan2.1-14B in model size and training data. Both also run a third guidance branch, which raises the forward passes per video to 150.

\noindent\textbf{Main results.}
As shown in Tab.~\ref{tab:gen}, GRACE stays within 0.02 of the pretrained pipeline before compression on I2V and exceeds it on T2V, while running $11.1\times$ faster. The T2V gain comes mostly from the semantic score, where GRACE leads the next best model by 5.28. The ablation in Tab.~\ref{tab:ablation} traces this gain to both stages: the dual latent alone keeps the semantic score at the pretrained level, and generation-aware alignment and the base-ahead offset raise it by 3.90 and 2.43. At $736{\times}1280{\times}81$ (Tab.~\ref{tab:gen_highres}), GRACE again reaches the highest T2V total at the lowest latency and stays within 0.04 of the pretrained pipeline on I2V while running $15.5\times$ faster.

\noindent\textbf{Quantitative comparison.}
In Tabs.~\ref{tab:gen} and~\ref{tab:gen_highres}, the most direct comparison is DC-Gen, which adapts the same pretrained DiT at twice our token count. GRACE runs faster than DC-Gen and scores higher on every VBench total, although DC-Gen is higher on the quality scores at $736{\times}1280{\times}81$, which we examine in the qualitative comparison below. GRACE also follows the conditioning image more closely, leading DC-Gen on the I2V score by 7.23 at $480{\times}832{\times}81$. LTX-Video reaches a higher I2V total at the same token count but runs slower, since a third guidance branch adds a DiT pass per step. On T2V, however, LTX-Video scores 6.64 below GRACE at $480{\times}832{\times}81$. Open-Sora 2.0 uses about twice our token count, yet runs slower and has the lowest VBench totals in both tasks at both resolutions. In text-to-video, GRACE leads on background consistency and temporal flickering at both resolutions, despite moving more than the pretrained pipeline. Appendix~\ref{app:user} reports a human evaluation against Wan2.1-14B and DC-Gen.

\noindent\textbf{Qualitative comparison.}
Figs.~\ref{fig:gen_qual} and~\ref{fig:gen_qual_highres} compare GRACE with the pretrained pipeline. GRACE preserves the scene layout, the subject, and the motion of the pretrained samples while using nearly $8\times$ fewer tokens. DC-Gen, by contrast, departs from the pretrained pipeline, producing samples that move more and are consistently more saturated (Figs.~\ref{fig:gen_i2v_qual_supp_1}--\ref{fig:gen_t2v_qual_supp_2}). These differences raise the quality scores of DC-Gen at $736{\times}1280{\times}81$, since the LAION aesthetic predictor used by VBench scores saturated images higher~\citep{schuhmann2022laion, Taylor_2026}. LTX-Video falls short in a different way, often missing the action described in the prompt. In the human action dimension of VBench-T2V, LTX-Video scores 86.00 at both resolutions, against 96.00--100.00 for every other model, including 99.00 and 100.00 for GRACE (Tabs.~\ref{tab:vbench_t2v} and~\ref{tab:vbench_t2v_highres}). At $736{\times}1280{\times}81$ in image-to-video, the motion of LTX-Video tends to come from a global zoom or a slow camera movement over the conditioning image, while the scene stays static (Figs.~\ref{fig:i2v_ltx_comp_0}--\ref{fig:i2v_ltx_comp_2}). GRACE instead stays close to the color and style of the pretrained pipeline.

\newpage
\subsection{Training Cost and Convergence}
\label{sec:cost}
\input{figures/9_train_efficiency}
Most of the training cost lies in adapting the DiT to the compressed latent.
Reusing the pretrained DiT reduces this cost substantially, since a DiT trained from scratch for the same number of steps falls far behind on both VBench-T2V and VBench-I2V (Tab.~\ref{tab:ablation}, rows II and III).
Reuse alone is not sufficient, however, as Open-Sora 2.0 reports blurry videos that do not fully converge even on 160 GPUs after adapting a pretrained DiT to Video DC-AE, an autoencoder trained from scratch~\citep{zheng2026opensora20trainingcommerciallevel}.
Under the same budget, both the Video DC-AE latent and our single-latent baseline fall short of the pretrained pipeline, although the latter is fine-tuned from the pretrained autoencoder with $\mathcal{L}_\text{recon}$ (Tab.~\ref{tab:recon}).
The adaptation cost thus appears to depend on how easily the pretrained DiT can learn the latent, which $\mathcal{L}_\text{recon}$ alone does not account for.
Adding $\mathcal{L}_\text{align}$ makes the residual converge faster to a lower flow matching loss (Fig.~\ref{fig:convergence}), while the base channels behave almost identically, since the DiT already models the base latent space.
As a result, GRACE matches the pretrained pipeline with 8.5 H200 GPU days for the autoencoder and 30 for the DiT.

\subsection{Ablations and Discussion}
\label{sec:ablation}

\noindent\textbf{Ablation on design components.}
Each component in Tab.~\ref{tab:ablation} improves generation on both tasks. The dual latent and the offset raise the T2V total by 1.43 and 1.13 but the I2V total by only 0.20 and 0.39, as both determine what the generation is anchored to, whereas I2V already provides the conditioning image as an anchor. The dual latent also makes the alignment possible, since $\mathcal{L}_\text{align}$ needs a pretrained latent to match against. Alignment behaves differently, raising the I2V score as well as both T2V scores, since the conditioning image anchors the content but does not resolve the mismatch between the compressed latent and the DiT's pretrained denoising space. Alignment therefore recovers most of the gap to the pretrained pipeline on I2V and goes further on T2V, where the total ends up above that pipeline, which may relate to the latent structure that the alignment loss induces (Fig.~\ref{fig:latent_analysis}).

\input{tables/5_ablation_tab}
\input{figures/6_align_conv}
\noindent\textbf{Alignment target.}
We align the compressed latent inside the pretrained DiT (Section~\ref{sec:alignment}), and Tab.~\ref{tab:align} assesses the impact of the alignment target while keeping everything else fixed. Prior work regularizes a high-dimensional latent with a pretrained vision foundation model so that it is easier for a diffusion model to learn~\citep{yao2025reconstructionvsgenerationtaming}. For that comparison, we use V-JEPA 2.1~\citep{murlabadia2026vjepa21unlockingdense}, a video foundation model whose dense features are spatially structured and temporally consistent. We keep the loss, the adaptive weighting, and the structure of $P_l$, and replace the target with V-JEPA 2.1 features of the clean video, applying $P_l$ directly to the compressed latent since the target is no longer in the DiT feature space (see Appendix~\ref{app:vjepa} for the setup). Aligning to V-JEPA 2.1 raises the semantic score from 78.65 to 80.04, while aligning inside the pretrained DiT raises it far higher, to 82.55, with the totals following the same order. These results suggest that vision foundation models provide useful semantic structure in general, but in our setting, the relevant structure is the one already represented by the pretrained DiT, since the compressed autoencoder is adapted to a fixed denoising model.

\input{figures/8_offset_tab}
\noindent\textbf{Denoising schedule design.}
Tab.~\ref{tab:offset} compares the denoising order, used in both DiT adaptation and sampling. Denoising the base first improves the total, while reversing it drops below the synchronous schedule, showing that the direction of the offset is what matters. The gain costs nothing at inference, since both parts are denoised in the same forward pass.

%% file: figures/5_gen_qual.tex
\begin{figure}[!t]
\centering
\small
\captionof{table}{\textbf{Video generation on VBench~\citep{huang2023vbenchcomprehensivebenchmarksuite, huang2024vbenchcomprehensiveversatilebenchmark} at $480{\times}832{\times}81$.} Notation follows Tab.~\ref{tab:recon}. Our method preserves the quality of Wan2.1-14B~\citep{wan2025wanopenadvancedlargescale} before compression on both tasks while using nearly $8\times$ fewer tokens and running $11.1\times$ faster. NFE counts the DiT forward passes per video over 50 sampling steps. Per-dimension scores are in Appendix~\ref{app:gen}.}
\label{tab:gen}
\setlength{\tabcolsep}{4pt}
\resizebox{\linewidth}{!}{%
\begin{tabular}{l l c c c c c c c c c c c c c}
\toprule
\multirow{2}{*}[-0.25\normalbaselineskip]{Model} & \multirow{2}{*}[-0.25\normalbaselineskip]{Autoencoder} & \multirow{2}{*}[-0.25\normalbaselineskip]{Config} & \multirow{2}{*}[-0.25\normalbaselineskip]{\makecell{Diffusion\\Model}} & \multirow{2}{*}[-0.25\normalbaselineskip]{\makecell{Latent\\Tokens}} & \multirow{2}{*}[-0.25\normalbaselineskip]{\makecell{DiT\\Params}} & \multirow{2}{*}[-0.25\normalbaselineskip]{NFE} & \multicolumn{2}{c}{Latency (s) $\downarrow$} & \multicolumn{3}{c}{VBench-T2V} & \multicolumn{3}{c}{VBench-I2V} \\
\cmidrule(lr){8-9} \cmidrule(lr){10-12} \cmidrule(lr){13-15}
 & & & & & & & T2V & I2V & Quality $\uparrow$ & Semantic $\uparrow$ & Total $\uparrow$ & I2V $\uparrow$ & Quality $\uparrow$ & Total $\uparrow$ \\
\midrule
Wan2.1-14B & Wan2.1-VAE & f8t4c16p2 & Wan2.1-14B & 32.8k & 14B & 100 & 851.5 & 863.2 & 85.24 & 78.70 & 83.93 & \textbf{95.82} & 80.01 & 87.92 \\

\arrayrulecolor{black!12}\midrule\arrayrulecolor{black}
Open-Sora 2.0 & Video DC-AE & f32t4c128p1 & Open-Sora & 8.2k & 11B & 150 & 138.6 & 138.4 & 78.80 & 72.35 & 77.51 & 91.17 & 76.96 & 84.07 \\
DC-Gen & DC-AE-V & f32t4c32p1 & Wan2.1-14B & 8.2k & 14B & 100 & 157.1 & 165.0 & 85.85 & 79.70 & 84.62 & 88.25 & 80.01 & 84.13 \\
\arrayrulecolor{black!12}\midrule\arrayrulecolor{black}
LTX-Video 0.9.7 & LTX-VAE & f32t8c128p1 & LTX-Video & 4.3k & 13B & 150 & 99.6 & 104.1 & 82.88 & 64.31 & 79.17 & 95.62 & \textbf{80.32} & \textbf{87.97} \\
Single-latent Baseline & Single-latent Baseline & f16t8c32p2 & Wan2.1-14B & 4.3k & 14B & 100 & 76.1 & 77.8 & 83.21 & 77.75 & 82.12 & 93.67 & 79.21 & 86.44 \\
\rowcolor{gray!12}
\textbf{GRACE (Ours)} & \textbf{GRACE-VAE} & f16t8c32p2 & Wan2.1-14B & 4.3k & 14B & 100 & \textbf{75.8} & \textbf{77.7} & \textbf{86.02} & \textbf{84.98} & \textbf{85.81} & 95.48 & 80.31 & 87.90 \\
\bottomrule
\end{tabular}%
}

\vspace{5pt}
\includegraphics[width=\linewidth]{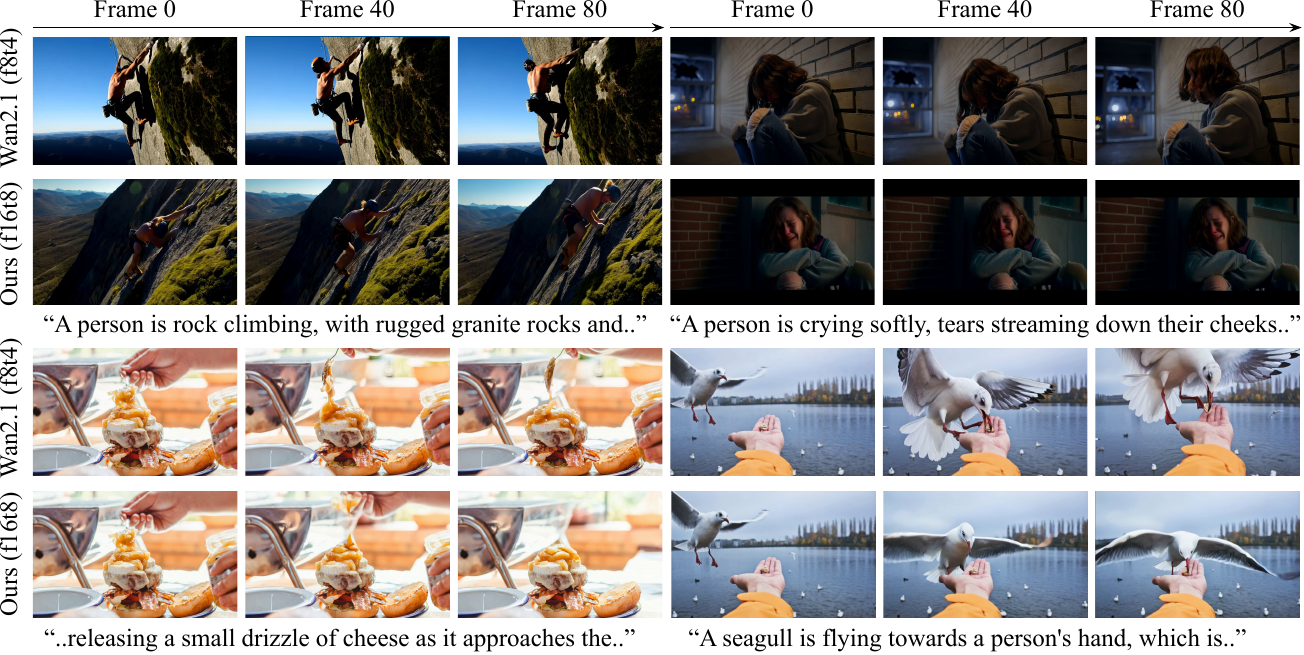}
\captionof{figure}{\textbf{Qualitative comparison at $480{\times}832{\times}81$.} VBench~\citep{huang2023vbenchcomprehensivebenchmarksuite, huang2024vbenchcomprehensiveversatilebenchmark} samples from Wan2.1-14B~\citep{wan2025wanopenadvancedlargescale} before compression and GRACE (Ours), generated from the same prompt, for text-to-video (\textbf{top}) and image-to-video (\textbf{bottom}). Best viewed when zoomed in.}
\label{fig:gen_qual}
\vspace{-8pt}
\end{figure}

%% file: figures/gen_main_quan_high_res.tex
\begin{figure}[!t]
\centering
\small
\captionof{table}{\textbf{Video generation on VBench~\citep{huang2023vbenchcomprehensivebenchmarksuite, huang2024vbenchcomprehensiveversatilebenchmark} at $736{\times}1280{\times}81$.} Notation follows Tab.~\ref{tab:gen}. Per-dimension scores are in Appendix~\ref{app:gen}. $^\ddagger$Measured with spatial tiling in the VAE encoder to avoid running out of memory when encoding the conditioning image at this resolution.}
\label{tab:gen_highres}
\setlength{\tabcolsep}{4pt}
\resizebox{\linewidth}{!}{%
\begin{tabular}{l l c c c c c c c c c c c c c}
\toprule
\multirow{2}{*}[-0.25\normalbaselineskip]{Model} & \multirow{2}{*}[-0.25\normalbaselineskip]{Autoencoder} & \multirow{2}{*}[-0.25\normalbaselineskip]{Config} & \multirow{2}{*}[-0.25\normalbaselineskip]{\makecell{Diffusion\\Model}} & \multirow{2}{*}[-0.25\normalbaselineskip]{\makecell{Latent\\Tokens}} & \multirow{2}{*}[-0.25\normalbaselineskip]{\makecell{DiT\\Params}} & \multirow{2}{*}[-0.25\normalbaselineskip]{NFE} & \multicolumn{2}{c}{Latency (s) $\downarrow$} & \multicolumn{3}{c}{VBench-T2V} & \multicolumn{3}{c}{VBench-I2V} \\
\cmidrule(lr){8-9} \cmidrule(lr){10-12} \cmidrule(lr){13-15}
 & & & & & & & T2V & I2V & Quality $\uparrow$ & Semantic $\uparrow$ & Total $\uparrow$ & I2V $\uparrow$ & Quality $\uparrow$ & Total $\uparrow$ \\
\midrule
Wan2.1-14B & Wan2.1-VAE & f8t4c16p2 & Wan2.1-14B & 77.3k & 14B & 100 & 3361.3 & 3396.8 & 84.69 & 76.01 & 82.96 & 95.56 & 80.20 & 87.88 \\
\arrayrulecolor{black!12}\midrule\arrayrulecolor{black}
Open-Sora 2.0 & Video DC-AE & f32t4c128p1 & Open-Sora & 19.3k & 11B & 150 & 425.4 & 425.4 & 80.73 & 78.16 & 80.22 & 93.72 & 77.82 & 85.77 \\
DC-Gen & DC-AE-V & f32t4c32p1 & Wan2.1-14B & 19.3k & 14B & 100 & 456.4 & 550.7$^\ddagger$ & \textbf{86.08} & 78.80 & 84.62 & 92.04 & 80.73 & 86.39 \\
\arrayrulecolor{black!12}\midrule\arrayrulecolor{black}
LTX-Video 0.9.7 & LTX-VAE & f32t8c128p1 & LTX-Video & 10.1k & 13B & 150 & 264.2 & 274.6 & 84.46 & 63.58 & 80.29 & \textbf{95.71} & \textbf{81.79} & \textbf{88.75} \\
\rowcolor{gray!12}
\textbf{GRACE (Ours)} & \textbf{GRACE-VAE} & f16t8c32p2 & Wan2.1-14B & 10.1k & 14B & 100 & \textbf{215.6} & \textbf{218.8} & 85.42 & \textbf{82.04} & \textbf{84.74} & 95.54 & 80.15 & 87.84 \\
\bottomrule
\end{tabular}%
}

\vspace{10pt}
\includegraphics[width=\linewidth]{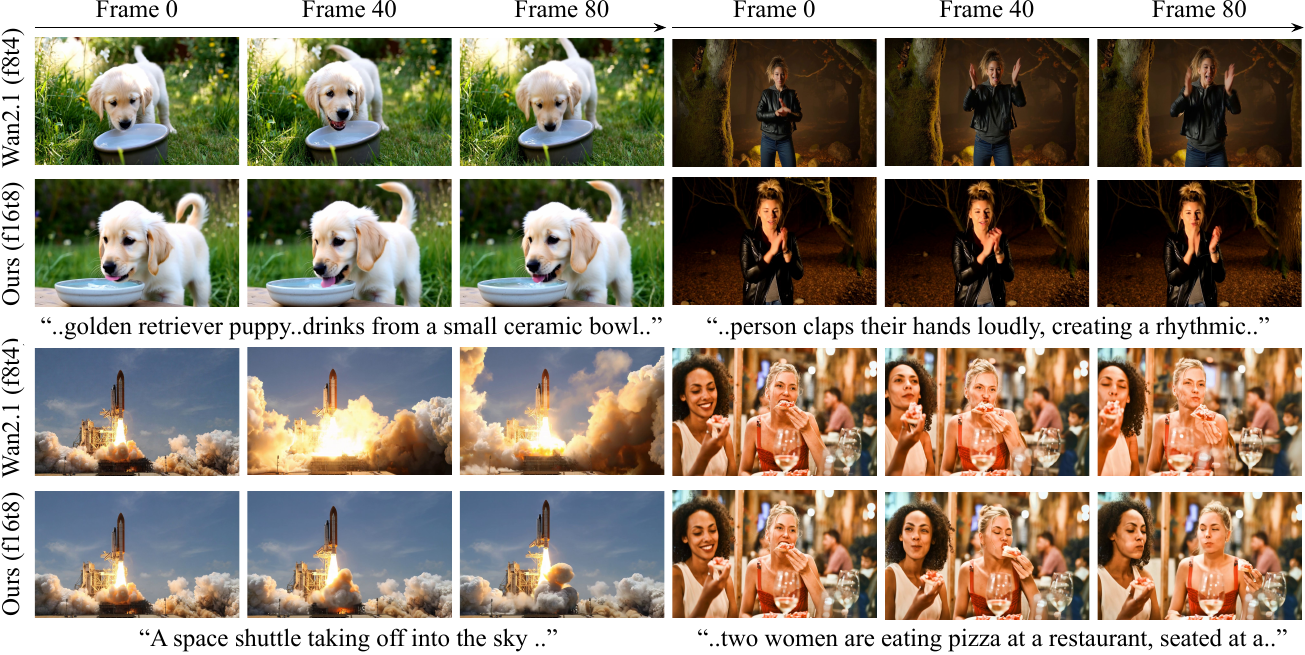}
\captionof{figure}{\textbf{Qualitative comparison at $736{\times}1280{\times}81$.} VBench~\citep{huang2023vbenchcomprehensivebenchmarksuite, huang2024vbenchcomprehensiveversatilebenchmark} samples from Wan2.1-14B~\citep{wan2025wanopenadvancedlargescale} before compression and GRACE (Ours), generated from the same prompt, for text-to-video (\textbf{top}) and image-to-video (\textbf{bottom}). Best viewed when zoomed in.}
\label{fig:gen_qual_highres}
\vspace{-5pt}
\end{figure}

%% file: figures/9_train_efficiency.tex
\begin{wrapfigure}{r}{0.25\linewidth}
\vspace{-10pt}
\centering
\includegraphics[width=\linewidth]{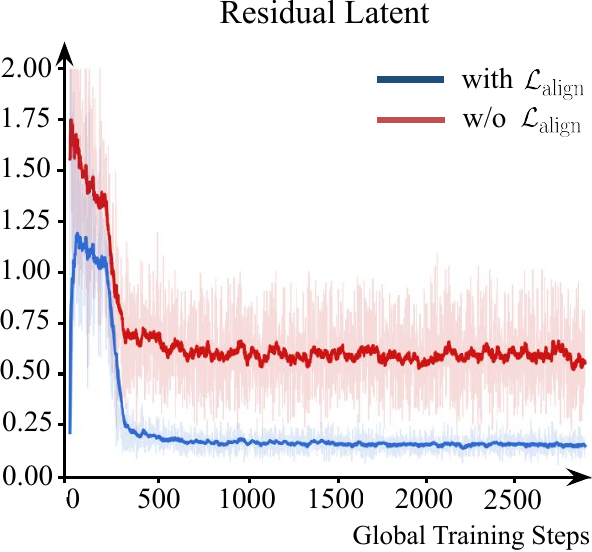}
\vspace{-20pt}
\captionof{figure}{\textbf{Convergence during DiT adaptation.} With and without $\mathcal{L}_\text{align}$; solid lines show EMA.}
\vspace{-8pt}
\label{fig:convergence}
\end{wrapfigure}


%% file: tables/5_ablation_tab.tex
\providecommand{\cmark}{\textcolor{green!60!black}{\ensuremath{\checkmark}}}
\providecommand{\xmark}{\textcolor{red}{\ensuremath{\times}}}
\begin{table}[!t]
\centering
\small
\caption{\textbf{Ablation on design components.} Row (III) keeps the autoencoder of (II) but trains the DiT from scratch for the same number of steps. Bold marks the best value in each column among rows (II) and (IV)--(VI).}
\label{tab:ablation}
\setlength{\tabcolsep}{6pt}
\setlength{\aboverulesep}{0pt}
\setlength{\belowrulesep}{0pt}
\renewcommand{\arraystretch}{1.15}
\resizebox{\linewidth}{!}{%
\begin{tabular}{l|l @{\hspace{10pt}} c c c @{\hspace{10pt}} c c c @{\hspace{10pt}} c c c @{\hspace{10pt}} c c c}
\toprule
\multicolumn{2}{c}{\multirow{2}{*}[-0.25\normalbaselineskip]{Method}} & \multicolumn{3}{c}{Components} & \multicolumn{3}{c}{Reconstruction} & \multicolumn{3}{c}{VBench-T2V} & \multicolumn{3}{c}{VBench-I2V} \\
\cmidrule(lr){3-5} \cmidrule(lr){6-8} \cmidrule(lr){9-11} \cmidrule(lr){12-14}
\multicolumn{2}{c}{} & $[\mathbf{z}_\text{base}; \mathbf{z}_\text{res}]$ & $\mathcal{L}_\text{align}$ & $\delta > 0$ & PSNR $\uparrow$ & LPIPS $\downarrow$ & rFVD $\downarrow$ & Quality $\uparrow$ & Semantic $\uparrow$ & Total $\uparrow$ & I2V $\uparrow$ & Quality $\uparrow$ & Total $\uparrow$ \\
\midrule
(I) & Wan2.1-14B (pretrained) & -- & -- & -- & 35.15 & 0.016 & 1.13 & 85.24 & 78.70 & 83.93 & 95.82 & 80.01 & 87.92 \\
\midrule
(II) & Single-latent Baseline & \xmark & \xmark & \xmark & \textbf{33.76} & \textbf{0.031} & \textbf{13.11} & 83.21 & 77.75 & 82.12 & 93.67 & 79.21 & 86.44 \\
(III) & (II) + DiT from scratch & \xmark & \xmark & \xmark & 33.76 & 0.031 & 13.11 & 74.20 & 58.40 & 71.04 & 53.06 & 73.94 & 63.50 \\
\midrule
(IV) & (II) + dual-latent & \cmark & \xmark & \xmark & 33.10 & \textbf{0.031} & 13.89 & 84.77 & 78.65 & 83.55 & 93.96 & 79.32 & 86.64 \\
(V) & (IV) + $\mathcal{L}_\text{align}$ & \cmark & \cmark & \xmark & 32.63 & 0.032 & 13.53 & 85.22 & 82.55 & 84.68 & 95.23 & 79.78 & 87.51 \\
\rowcolor{gray!12}
(VI) & (V) + $\delta > 0$ (\textbf{Ours}) & \cmark & \cmark & \cmark & 32.63 & 0.032 & 13.53 & \textbf{86.02} & \textbf{84.98} & \textbf{85.81} & \textbf{95.48} & \textbf{80.31} & \textbf{87.90} \\
\bottomrule
\end{tabular}%
}
\vspace{-10pt}
\end{table}

%% file: figures/6_align_conv.tex
\begin{wraptable}{r}{0.5\linewidth}
\centering
\vspace{-10pt}
\small
\captionof{table}{\textbf{Alignment target.} Asymmetric denoising is disabled in all rows.}
\vspace{-5pt}
\label{tab:align}
\setlength{\tabcolsep}{4pt}
\resizebox{\linewidth}{!}{%
\begin{tabular}{l c c c c c}
\toprule
\multirow{2}{*}[-0.25\normalbaselineskip]{Alignment Target} & \multicolumn{2}{c}{Reconstruction} & \multicolumn{3}{c}{VBench-T2V} \\
\cmidrule(lr){2-3} \cmidrule(lr){4-6}
 & PSNR $\uparrow$ & rFVD $\downarrow$ & Quality $\uparrow$ & Semantic $\uparrow$ & Total $\uparrow$ \\
\midrule
None & \textbf{33.10} & 13.89 & 84.77 & 78.65 & 83.55 \\
V-JEPA 2.1 features & 32.61 & 13.93 & 84.98 & 80.04 & 83.99 \\
Pretrained DiT features (\textbf{Ours}) & 32.63 & \textbf{13.53} & \textbf{85.22} & \textbf{82.55} & \textbf{84.68} \\
\bottomrule
\end{tabular}%
\vspace{-20pt}
}
\end{wraptable}

%% file: figures/8_offset_tab.tex
\begin{wraptable}{r}{0.4\linewidth}
\centering
\small
\captionof{table}{\textbf{Ablation on denoising order.}}
\vspace{-5pt}
\label{tab:offset}
\setlength{\tabcolsep}{4pt}
\resizebox{0.95\linewidth}{!}{%
\begin{tabular}{l c c c c}
\toprule
\multirow{2}{*}[-0.25\normalbaselineskip]{Denoising Order} & \multirow{2}{*}[-0.25\normalbaselineskip]{Offset} & \multicolumn{3}{c}{VBench-I2V} \\
\cmidrule(lr){3-5}
 & & I2V $\uparrow$ & Quality $\uparrow$ & Total $\uparrow$ \\
\midrule
$\tau_\text{base} = \tau_\text{res}$        & $\delta = 0$ & 95.23 & 79.78 & 87.51 \\
$\tau_\text{base} > \tau_\text{res}$        & $\delta < 0$ & 94.61 & 79.72 & 87.17 \\
$\tau_\text{base} < \tau_\text{res}$ \textbf{(Ours)} & $\delta > 0$ & \textbf{95.48} & \textbf{80.31} & \textbf{87.90} \\
\bottomrule
\end{tabular}%
}
\end{wraptable}

%% file: sec/6_Conclusion.tex
\section{Conclusion}
\label{sec:con}
We presented GRACE, a framework that compresses the latent of a pretrained video diffusion pipeline and adapts the DiT to the compressed latent. GRACE keeps a frozen base latent and learns a residual latent for the information lost under stronger compression, aligns the compressed latent with the pretrained latent in the feature space of the frozen DiT, and denoises the base ahead of the residual so that the residual builds on a settled base. GRACE matches the VBench generation quality of the pretrained pipeline before compression on Wan2.1-I2V-14B while running $11.1\times$ faster at $480{\times}832{\times}81$, and the speedup grows with resolution, reaching $15.5\times$ at $736{\times}1280{\times}81$. Across the compared autoencoders, better reconstruction does not mean better generation. Optimizing the autoencoder for reconstruction alone moves the latent away from the distribution the DiT has learned, which is why GRACE supervises compression in the space the DiT operates in.

%% file: sec/X_suppl.tex
\setcounter{section}{0}
\setcounter{table}{0}
\setcounter{figure}{0}
\setcounter{equation}{20}
\renewcommand{\thesection}{\Alph{section}}          
\renewcommand{\theHsection}{appendix.\Alph{section}} 
\renewcommand{\thefigure}{\thesection.\arabic{figure}}
\makeatletter
\@addtoreset{table}{section}
\@addtoreset{figure}{section}
\makeatother
\renewcommand{\thetable}{\thesection.\arabic{table}}

\clearpage
\section*{\LARGE Appendix}
\vspace{-10pt}
\noindent This appendix provides supplementary material to support the main paper.
\begin{itemize}[leftmargin=*,itemsep=2pt,topsep=4pt,parsep=0pt]
\item Appendix~\ref{app:ae_details} describes the architecture and the training setup of our autoencoder, and Appendix~\ref{app:diffusion_details} provides the corresponding details for the DiT, including how it is extended to the compressed latent.
\item Appendix~\ref{app:eval_details} describes how the compared autoencoders are adapted to the pretrained DiT and how every model is evaluated.
\item Appendix~\ref{app:eval} reports per-dimension VBench scores, results at a higher resolution, and a human evaluation.
\item Appendix~\ref{app:ablation} ablates the autoencoder design and the alignment design.
\item Appendix~\ref{app:analysis} analyzes the compressed latent and the convergence of each of its parts, and Appendix~\ref{app:cost} reports the training and inference cost.
\item Appendix~\ref{app:related} extends the discussion of related work, Appendix~\ref{app:add_qual} collects the qualitative samples referenced throughout the paper, and Appendix~\ref{app:limitations} states the limitations of our method.
\end{itemize}
\vspace{-10pt}

\bigskip
\noindent{\Large\textbf{Contents}}\\[2pt]
\noindent\rule{\linewidth}{0.5pt}
\vspace{3pt}

\vspace{-10pt}
\noindent\hbox to \linewidth{\textbf{A\quad GRACE Autoencoder Details}\dotfill\textbf{\pageref{app:ae_details}}}\\[1pt]
\noindent\hbox to \linewidth{\hspace*{1.5em}A.1\quad Architectural Details\dotfill\pageref{app:ae_architecture}}\\[1pt]
\noindent\hbox to \linewidth{\hspace*{1.5em}A.2\quad Implementation Details\dotfill\pageref{app:ae_implementation}}\\[1pt]

\vspace{-5pt}
\noindent\hbox to \linewidth{\textbf{B\quad GRACE Diffusion Model Details}\dotfill\textbf{\pageref{app:diffusion_details}}}\\[1pt]
\noindent\hbox to \linewidth{\hspace*{1.5em}B.1\quad Architectural Details\dotfill\pageref{app:diffusion_architecture}}\\[1pt]
\noindent\hbox to \linewidth{\hspace*{1.5em}B.2\quad Implementation Details\dotfill\pageref{app:diffusion_implementation}}\\[1pt]

\vspace{-5pt}
\noindent\hbox to \linewidth{\textbf{C\quad More Details on Evaluation}\dotfill\textbf{\pageref{app:eval_details}}}\\[1pt]
\noindent\hbox to \linewidth{\hspace*{1.5em}C.1\quad Adaptation of Other Autoencoders\dotfill\pageref{app:impl}}\\[1pt]
\noindent\hbox to \linewidth{\hspace*{1.5em}C.2\quad Evaluation Protocol\dotfill\pageref{app:protocol}}\\[1pt]
\noindent\hbox to \linewidth{\hspace*{1.5em}C.3\quad Alignment to V-JEPA 2.1\dotfill\pageref{app:vjepa}}\\[1pt]

\vspace{-5pt}
\noindent\hbox to \linewidth{\textbf{D\quad Additional Evaluation}\dotfill\textbf{\pageref{app:eval}}}\\[1pt]
\noindent\hbox to \linewidth{\hspace*{1.5em}D.1\quad Video Generation\dotfill\pageref{app:gen}}\\[1pt]
\noindent\hbox to \linewidth{\hspace*{1.5em}D.2\quad Human Evaluation\dotfill\pageref{app:user}}\\[1pt]

\vspace{-5pt}
\noindent\hbox to \linewidth{\textbf{E\quad Additional Ablations}\dotfill\textbf{\pageref{app:ablation}}}\\[1pt]
\noindent\hbox to \linewidth{\hspace*{1.5em}E.1\quad Autoencoder Design\dotfill\pageref{app:ablation_ae}}\\[1pt]
\noindent\hbox to \linewidth{\hspace*{1.5em}E.2\quad Alignment Design\dotfill\pageref{app:align}}\\[1pt]

\vspace{-5pt}
\noindent\hbox to \linewidth{\textbf{F\quad Additional Analysis}\dotfill\textbf{\pageref{app:analysis}}}\\[1pt]
\noindent\hbox to \linewidth{\hspace*{1.5em}F.1\quad Latent Analysis\dotfill\pageref{app:latent}}\\[1pt]
\noindent\hbox to \linewidth{\hspace*{1.5em}F.2\quad Convergence Behavior\dotfill\pageref{app:loss_split}}\\[1pt]

\vspace{-5pt}
\noindent\hbox to \linewidth{\textbf{G\quad Computational Cost}\dotfill\textbf{\pageref{app:cost}}}\\[1pt]

\vspace{-5pt}
\noindent\hbox to \linewidth{\textbf{H\quad More Discussion on Related Work}\dotfill\textbf{\pageref{app:related}}}\\[1pt]

\vspace{-5pt}
\noindent\hbox to \linewidth{\textbf{I\quad Additional Qualitative Results}\dotfill\textbf{\pageref{app:add_qual}}}\\[1pt]

\vspace{-5pt}
\noindent\hbox to \linewidth{\textbf{J\quad Limitations}\dotfill\textbf{\pageref{app:limitations}}}\\[1pt]
\vspace{3pt}
\noindent\rule{\linewidth}{0.5pt}
\clearpage

\section{GRACE Autoencoder Details}
\label{app:ae_details}

\subsection{Architectural Details}
\label{app:ae_architecture}
\input{supple/figures/0_ae_arch}

Fig.~\ref{fig:ae_arch} shows the architecture of our autoencoder in detail. We reuse the encoder and decoder backbones of Wan2.1 and describe our additions below.

\noindent\textbf{Base encoding.}
We obtain $\mathbf{z}_\text{base}$ from the frozen encoder $\mathcal{E}$. Since $\mathcal{E}$ compresses at fixed factors, we reduce the video first. We downsample the frames by $r_s$ with bilinear interpolation, and subsample them by $r_t$ in time. We experiment with average pooling, strided sampling, and bilinear interpolation as reduction strategies, and the setting above reconstructs best.

\noindent\textbf{Residual encoding.}
The residual encoder $\mathcal{E}_\text{res}$ has the same architecture as $\mathcal{E}$, but takes the full-resolution video, so it needs an additional compression step to match the target resolution. In the pretrained encoder, four downsampling stages are followed by middle blocks that refine the feature at the final resolution through residual and attention layers, and then by a head that projects it to the latent. We add one downsample block between the middle blocks and the head. The block reuses the design of the pretrained downsample blocks, which pass the feature through two residual blocks before a strided convolution over space and a strided causal convolution over time. We attach a non-parametric shortcut to this block, following the residual autoencoding of DC-AE~\citep{chen2025deepcompressionautoencoderefficient}. Since the block compresses time as well as space, the shortcut moves both axes into the channel axis, then averages channel groups to match the channel number of the block output. Its result is added to the block output, which leaves the block to learn the residual.

\noindent\textbf{Decoding.}
The decoder $\hat{\mathcal{D}}$ extends the input convolution of $\mathcal{D}$ from $C$ to $C{+}C'$ channels for the concatenated latent, and adds one upsample block symmetric to the downsample block of $\mathcal{E}_\text{res}$. The weights for the residual channels are zero-initialized, so training starts at the pretrained reconstruction quality. The non-parametric shortcut of this block runs in the opposite direction, duplicating channels and then applying a channel-to-space-and-time operation. In the image-to-video setting, the first frame is available at inference as well as during training, and we pass it to $\hat{\mathcal{D}}$ to restore detail the compressed latent cannot carry on its own. We encode the first frame with $\mathcal{E}_\text{res}$ and inject its intermediate features into $\hat{\mathcal{D}}$ through gated cross-attention~\citep{tian2025reduciogenerating1kvideo}. We take the feature before each downsampling stage of the encoder and attend to it at the matching stage of $\hat{\mathcal{D}}$. A learned gate on each block controls how much of the first frame reaches $\hat{\mathcal{D}}$.

\noindent\textbf{Single-latent baseline.}
The baseline replaces the base and residual pair with a single encoder, so there is no frozen $\mathcal{E}$ and no reduction of the input. The compression block keeps the same design and placement, and both the non-parametric shortcut and the first-frame cross-attention are unchanged.

\clearpage
\subsection{Implementation Details}
\label{app:ae_implementation}

We train the autoencoder in two phases. In the first phase, we update $\mathcal{E}_\text{res}$ and $\hat{\mathcal{D}}$ at $256{\times}256{\times}81$, along with $W^\text{res}_\text{in}$ and the per-layer projections $P_l$. All other parameters of $\mathbf{v}_\theta$ and $\mathbf{v}_{\theta,+}$ remain frozen. In the second phase, we raise the resolution to $512{\times}512{\times}81$ and train $\hat{\mathcal{D}}$ alone, following the decoupled high-resolution adaptation of DC-AE~\citep{chen2025deepcompressionautoencoderefficient}, which freezes $\mathcal{E}_\text{res}$ as well so that the latent space stays fixed while the decoder adapts. $\mathcal{E}$ stays frozen in both phases. We train a separate autoencoder for each task, with $\mathcal{L}_\text{align}$ supervised by the pretrained DiT of that task, and report reconstruction with the image-to-video autoencoder.

\input{supple/tables/ae_training_details}
\noindent\textbf{Training setup.}
We train on Panda-70M~\citep{chen2024panda70mcaptioning70mvideos} with AdamW, using the hyperparameters in Tab.~\ref{tab:hp_ae}. The clip length stays at $81$ frames, since reconstruction degrades quickly on shorter clips under high temporal compression. The modules we add on the DiT side, $W^\text{res}_\text{in}$ and $P_l$, are trained with a separate learning rate.

\noindent\textbf{Initialization.}
We initialize $\mathcal{E}_\text{res}$ and $\hat{\mathcal{D}}$ from the pretrained weights. The added downsample and upsample blocks are trained from scratch, and the input convolution of $\hat{\mathcal{D}}$ is zero-initialized on the residual channels. The gate of each first-frame cross-attention block is also zero, so the first frame has no effect at the start of training.

\noindent\textbf{Training objective.}
We follow the loss configuration of the pretrained autoencoder, with the weights listed in Tab.~\ref{tab:hp_ae}. We apply $\mathcal{L}_\text{align}$ only in the first phase, since the second phase leaves the latent unchanged. Each $P_l$ is a two-layer projection with a bottleneck of $16$ channels, zero-initialized on the output so that the alignment starts from no contribution. We sample $\tau$ with the scheduler of the pretrained pipeline.

\noindent\textbf{Single-latent baseline.}
The baseline trains with $\mathcal{L}_\text{recon}$ alone, since there is no residual to shape and $\mathcal{L}_\text{align}$ never applies. The two phases and every hyperparameter in Tab.~\ref{tab:hp_ae} are the same.

\clearpage

\section{GRACE Diffusion Model Details}
\label{app:diffusion_details}

\subsection{Architectural Details}
\label{app:diffusion_architecture}

We reuse the Wan2.1 DiT and modify only what the compressed latent requires.

\noindent\textbf{Input and output projections.}
Wan2.1-14B patchifies a latent of $C$ channels, and for image-to-video it also takes a mask and the latent of the conditioning frame. We extend all three along the channel axis. The mask width is tied to the temporal compression factor, so it doubles with the latent, and we copy the pretrained weights into the channels it had before and zero-initialize the rest. The output projection is widened the same way, with the added channels zero-initialized, so the model starts from its pretrained behavior.

\noindent\textbf{Position encoding.}
Compression places tokens farther apart in the video than the DiT was trained to expect. We therefore scale the RoPE~\citep{su2023roformerenhancedtransformerrotary} indices by $r_s$ in space and $r_t$ in time. This restores the token spacing the DiT was pretrained with.

\noindent\textbf{Timestep conditioning.}
The base and the residual latents are denoised at different noise levels, so the DiT receives two timesteps. For the transformer blocks, we add a second modulation projection $W_\text{mod}^\text{res}$ for the residual timestep, and the two modulations are summed. The two output heads are conditioned separately, each on its own timestep.

\noindent\textbf{Single-latent baseline.}
The baseline extends the input and output projections in the same way, but the latent is not split, so one timestep suffices and $W_\text{mod}^\text{res}$ is not added.

\subsection{Implementation Details}
\label{app:diffusion_implementation}

We adapt the pretrained DiT to the compressed latent while the autoencoder stays frozen. Before the latent enters the DiT, we standardize each part with statistics measured on the training set, following the pretrained pipeline. We fully fine-tune the input projection, the output heads, and $W_\text{mod}^\text{res}$, together with the modulation and normalization parameters of each block. LoRA~\citep{hu2021loralowrankadaptationlarge} is applied to all linear layers in the attention and feed-forward blocks, as well as to the key and value projections of the image cross-attention for image-to-video.

\input{supple/tables/dit_training_details}

\noindent\textbf{Training setup.}
We train on Panda-70M~\citep{chen2024panda70mcaptioning70mvideos} at $480{\times}832{\times}81$ with AdamW, using the hyperparameters in Tab.~\ref{tab:hp_dit}. We apply the standard video preprocessing, filtering out clips with scene cuts or excessive motion, and rewrite the captions with GPT-4o~\citep{openai2024gpt4technicalreport} so that they match the length and detail of the prompts Wan2.1-14B was trained on. Text conditioning is dropped with probability $0.1$ for classifier-free guidance.

\noindent\textbf{Initialization.}
We keep the pretrained weights for the base channels, and zero-initialize the weights for the residual channels along with $W_\text{mod}^\text{res}$.

\noindent\textbf{Denoising schedule.}
We sample the offset per video rather than fixing it, so a single model covers a range of offsets at inference. The shift $s$ of Eq.~\ref{eq:shift} is sampled the same way, and we find that a lower shift, which spends more sampling steps near the clean end, recovers detail better on the compressed latent. The values are listed in Tab.~\ref{tab:hp_dit}. At inference, we run the same 50 sampling steps as the synchronous schedule over $\hat{u} \in [0, 1+\delta]$, and $\delta = 0$ recovers the synchronous schedule exactly. In the first few steps, only the base is denoised while the residual stays at pure noise, and in the last few, only the residual is denoised while the base is kept fixed until the final step.

\noindent\textbf{Single-latent baseline.}
The baseline follows the training recipe of Wan2.1, where a single timestep leaves no offset or shift to sample. The weights added to the projections are randomly initialized, following DC-Gen~\citep{he2026dc}.

\section{More Details on Evaluation}
\label{app:eval_details}

\subsection{Adaptation of Other Autoencoders}
\label{app:impl}
This section details how the pretrained DiT is adapted to each compared autoencoder in Tab.~\ref{tab:recon}. Our own autoencoder is adapted as described in Section~\ref{sec:dit}, which extends the pretrained input and output projections and denoises the base ahead of the residual. Both steps build on the base latent from the frozen Wan2.1 encoder, which is unavailable to the compared autoencoders, as their latent spaces are entirely different from the one the DiT was trained on. For each of them, we therefore adapt Wan2.1-I2V-14B with the released implementation of DC-Gen~\citep{he2026dc}, a recent method that adapts a pretrained DiT to a new latent space in three stages: first the patch embedding, then the input and output layers, and finally the whole transformer with LoRA~\citep{hu2021loralowrankadaptationlarge}. Each autoencoder keeps the patch size of 1 used in its original generation model. This adaptation uses the same total budget of 38.5 H200 GPU days as our full pipeline, and all other settings are shared.

\subsection{Evaluation Protocol}
\label{app:protocol}
\noindent\textbf{Autoencoder comparison.}
In Tab.~\ref{tab:recon}, we use Step-Video-VAE v2~\citep{ma2025stepvideot2vtechnicalreportpractice}, the Video DC-AE of Open-Sora 2.0~\citep{zheng2026opensora20trainingcommerciallevel}, and LTX-VAE 0.9.7~\citep{hacohen2024ltxvideorealtimevideolatent}, whose 13B release is closest in size to the generation backbone we compare against. Video DC-AE is evaluated with the spatial and temporal tiling of its released implementation, which uses spatial tiles of 256 pixels and temporal tiles of 32 frames with an overlap factor of 0.25, and reconstructs better than the untiled variant.

\noindent\textbf{Generation comparison.}
Every model adapted from Wan2.1, including ours, the single-latent baseline, and the latents in Tab.~\ref{tab:recon}, is sampled with a guidance scale of 5.0 and a flow shift of 3.0 on the full prompt and conditioning image set of VBench-I2V~\citep{huang2023vbenchcomprehensivebenchmarksuite, huang2024vbenchcomprehensiveversatilebenchmark}. The other generators in Tab.~\ref{tab:gen} are run from their released checkpoints, the 13B development release of LTX-Video 0.9.7, the high-compression Video DC-AE release of Open-Sora 2.0, and the released DC-Gen model~\citep{he2026dc}. We run LTX-Video as a single 50-step pass at the target resolution, without its multiscale pipeline and spatial upscaler, so that every model denoises the same number of steps at the same resolution. For text-to-video, each model is run with the prompt pipeline recommended by its authors: Wan2.1-14B, GRACE, and DC-Gen each use GPT-enhanced prompts from the VBench release, LTX-Video uses its built-in prompt enhancer, and Open-Sora 2.0 follows its default text-to-image-to-video path without prompt refinement. For image-to-video, VBench-I2V provides only short prompts. We expand them once with GPT-4o~\citep{openai2024gpt4technicalreport} and use the same expanded set for Wan2.1-14B, GRACE, and Open-Sora 2.0, which do not release prompts for this setting. DC-Gen uses the extended prompts released with it, and LTX-Video uses its built-in prompt enhancer. Every model takes 50 sampling steps, but the number of forward passes through the DiT differs. LTX-Video combines classifier-free guidance with spatio-temporal guidance, and Open-Sora 2.0 guides on text and image separately, so both evaluate the DiT three times per step rather than twice. Latency covers the conditioning encode, the denoising loop, and the decode of the output video, measured on a single NVIDIA A100 SXM4-80GB in PyTorch with bfloat16 and no inference-time compilation. It excludes model loading, text encoding, and writing the decoded frames to an MP4 file, none of which depend on the latent resolution.

\noindent\textbf{Latent uniformity.}
Following VA-VAE~\citep{yao2025reconstructionvsgenerationtaming}, we measure how evenly the latent tokens are distributed. We encode $10{,}000$ clips at $256{\times}256{\times}81$, each from a distinct source video in the Panda-70M~\citep{chen2024panda70mcaptioning70mvideos} training split, and keep one token per clip, sampled at a random spatio-temporal position. Every autoencoder is probed at the same clips and positions. The tokens are standardized per channel and embedded with t-SNE~\citep{JMLR:v9:vandermaaten08a} at a perplexity of $30$, and we report the coefficient of variation, Gini coefficient, and normalized entropy of a Gaussian kernel density estimate on the embedding, averaged over two t-SNE seeds.

\subsection{Alignment to V-JEPA 2.1}
\label{app:vjepa}
For Tab.~\ref{tab:align}, we use the frozen V-JEPA 2.1~\citep{murlabadia2026vjepa21unlockingdense} ViT-L encoder at $256{\times}256$, whose $16{\times}16$ patch grid already matches the spatial grid of our f16 latent, so no spatial interpolation is needed. Since V-JEPA 2.1 is trained on clips of up to 64 frames, we apply the alignment to the first 65 frames of each 81-frame training clip. Our causal latent encodes the first frame on its own, so we encode it with V-JEPA 2.1 separately and align it with the first latent frame. Each remaining latent frame covers 8 frames and corresponds to four V-JEPA 2.1 tokens of 2 frames each, so a learnable transposed 3D convolution upsamples these latent frames $4\times$ in time to match. We align to the features of block 23, the final block of the encoder, whose output V-JEPA 2.1 uses directly for dense downstream tasks~\citep{murlabadia2026vjepa21unlockingdense}. The alignment uses the same per-token cosine loss and adaptive weighting as $\mathcal{L}_\text{align}$.

\raggedbottom 
\section{Additional Evaluation}
\label{app:eval}

\subsection{Video Generation}
\label{app:gen}
\input{supple/tables/gen_add_Eval_i2v_t2v_merged}
\input{supple/tables/gen_add_eval_high_res}
We report the VBench scores for each dimension in Tab.~\ref{tab:vbench_t2v} and Tab.~\ref{tab:vbench_i2v}. Several dimensions rank differently from the totals or across settings, and we discuss them below.

\noindent\textbf{Aesthetic quality.}
DC-Gen scores above GRACE on this dimension in T2V settings, while scoring below on every total. We observe that its samples are consistently more saturated than those of the pretrained pipeline (Figs.~\ref{fig:gen_i2v_qual_supp_1}--\ref{fig:gen_t2v_qual_supp_2}), and the LAION aesthetic predictor behind this dimension has been shown to track photographic taste rather than fidelity to the prompt or to the conditioning image~\citep{Taylor_2026}. The dimension therefore rewards a shift in appearance that the other dimensions penalize.

\noindent\textbf{Background consistency, motion smoothness, and temporal flickering.}
In text-to-video, GRACE leads on background consistency and temporal flickering at both resolutions while moving more than the pretrained pipeline, with a dynamic degree of 62.50 against 56.94 at $480{\times}832$. In image-to-video at $480{\times}832$, LTX-Video leads on all three dimensions. It runs an extra guidance branch for temporal consistency, and in the image-to-video setting it also moves the least of all compared models, with a dynamic degree of 24.80 against 38.62 for GRACE. Videos that move less tend to score higher on these dimensions~\citep{huang2023vbenchcomprehensivebenchmarksuite, huang2024vbenchcomprehensiveversatilebenchmark}, and the same holds at $736{\times}1280$, where GRACE moves less than LTX-Video (30.89 against 49.59) and leads on background consistency and temporal flickering instead.

\noindent\textbf{Preserving the pretrained appearance.}
Given the same prompt, GRACE stays close to the color and style of the pretrained pipeline in text-to-video, while DC-Gen often departs from it (Figs.~\ref{fig:gen_t2v_qual_supp_1} and \ref{fig:gen_t2v_qual_supp_2}). The effect is weaker in image-to-video, where the conditioning image fixes the appearance for every model.

At $736{\times}1280{\times}81$ (Tab.~\ref{tab:vbench_t2v_highres} and Tab.~\ref{tab:vbench_i2v_highres}), the other per-dimension scores follow the same pattern. DC-Gen again leads on aesthetic quality in text-to-video while trailing on both totals, and LTX-Video keeps the highest subject consistency in image-to-video.

\clearpage
\newlength{\wanw}
\newlength{\dcw}

\subsection{Human Evaluation}
\label{app:user}
\input{supple/tables/user_study}
We conduct a blind user study comparing our method with Wan2.1-14B~\citep{wan2025wanopenadvancedlargescale} and the adaptation-based DC-Gen~\citep{he2026dc} in both the T2V and I2V settings, using 40 prompts randomly selected per setting from VBench~\citep{huang2023vbenchcomprehensivebenchmarksuite,huang2024vbenchcomprehensiveversatilebenchmark} (with their conditioning images for I2V).
Each model uses its own prompt extension; participants see the original VBench prompt.
Participants view two videos generated from the same prompt, ours and one baseline, in randomized A/B order, and judge which is better in
visual quality (fewer unnatural colors or shapes and a better overall appearance),
temporal consistency (no flicker, stutter, or objects changing over time), and
text alignment (the subjects, actions, details, and style described in the prompt), with an ``about equal'' option for each.
39 human raters each rated 16 pairs, eight per setting, so each of the 160 comparison pairs is rated by at least 3 different participants, giving 156 votes per setting, baseline, and criterion.
Tab.~\ref{tab:human_eval} reports the win, tie, and loss rates of our method.

Against DC-Gen, which also operates in a compressed latent space, participants prefer our model by a wide margin in both settings, choosing it in 60.3--71.2\% of the votes on visual quality and temporal consistency against 14.1--24.4\% for DC-Gen. Against Wan2.1-14B before compression, our model is preferred in T2V on all three criteria, with 46.8--52.6\% of the votes against 25.6--35.3\%, whereas Wan2.1-14B is preferred in I2V with 34.6--44.9\% against 22.4--29.5\%. In I2V, many votes also rate the two as about equal, up to 42.9\% for text alignment.
Figs.~\ref{fig:user_study_T2V} and \ref{fig:user_study_I2V} show the interface used in the study.

\input{supple/figures/14_user_study_interface_T2V}
\input{supple/figures/15_user_study_interface_I2V}

\clearpage

\section{Additional Ablations}
\label{app:ablation}

\subsection{Autoencoder Design}
\label{app:ablation_ae}
We examine the design choices behind the autoencoder that are not covered in Appendix~\ref{app:ae_details}.

\input{supple/tables/first_frame}
\noindent\textbf{First-frame conditioning.}
For image-to-video, $\hat{\mathcal{D}}$ receives the first frame through gated cross-attention. We drop it with probability $0.5$ during training, so that the decoder learns to reconstruct without it and relies on the compressed latent rather than copying from the first frame. Even without the first frame at inference, reconstruction stays close to that with it (Tab.~\ref{tab:firstframe}).

\subsection{Alignment Design}
\label{app:align}
We examine the design choices behind $\mathcal{L}_\text{align}$, the alignment applied in Stage 1.

\input{supple/tables/align_depth}
\noindent\textbf{Alignment depth.}
We compare supervising the first 10 blocks of the DiT with supervising all 40. As Tab.~\ref{tab:depth} shows, the two settings generate at the same level, within 0.04 on the I2V total, but supervising all 40 loses 0.42 dB in PSNR, raises LPIPS from 0.032 to 0.038, and converges more slowly. The later blocks likely conflict with $\mathcal{L}_\text{recon}$ more strongly, so the encoder spends its capacity on matching them rather than on reconstruction. We therefore supervise the first 10.

\clearpage

\section{Additional Analysis}
\label{app:analysis}

\subsection{Latent Analysis}
\label{app:latent}
In this section, we look into what the base and the residual latent each contain, first by perturbing one part while the other stays clean and then through their principal components.

\input{supple/figures/2_latent_analysis}

\noindent\textbf{Per-latent analysis on reconstruction.}
We analyze how $\mathbf{z}_\text{base}$ and $\mathbf{z}_\text{res}$ each affect the reconstruction performance by injecting a controlled amount of noise into one of them while keeping the other clean. Fig.~\ref{fig:latent_perturbation} sweeps $\tau$ from $0$ to $1$ in the flow matching interpolation of Eq.~\ref{eq:add_noise}, so that $\tau = 0$ leaves the latent unchanged and $\tau = 1$ replaces it with pure noise. The noise is scaled by the per-channel variance of the latent it is added to. We report PSNR and LPIPS on Panda-70M~\citep{chen2024panda70mcaptioning70mvideos} at $480{\times}832{\times}81$ for each level.

As plotted in Fig.~\ref{fig:latent_perturbation}~(a), both PSNR and LPIPS degrade steadily for either latent under noise, and faster for $\mathbf{z}_\text{base}$ across the whole range. Fig.~\ref{fig:latent_perturbation}~(b) also shows that both $\mathbf{z}_\text{base}$ and $\mathbf{z}_\text{res}$ fail in different ways, with $\mathbf{z}_\text{base}$ breaking the spatial layout of the scene while $\mathbf{z}_\text{res}$ keeps the layout and instead blurs each frame and makes the motion appear at a lower frame rate. This is expected from the design, since the reduction in space and time leaves the overall scene in $\mathbf{z}_\text{base}$ while $\mathbf{z}_\text{res}$ carries what the reduction removes and the motion between the sampled frames. The larger drop for $\mathbf{z}_\text{base}$ follows as well, and it is what we intend: $\mathbf{z}_\text{base}$ stays in the space the DiT was pretrained on, so the latent that carries more of the reconstruction is also the one that needs the least adaptation. The burden of Stage 2 therefore falls on $\mathbf{z}_\text{res}$, which lies outside the space the pretrained DiT was trained on. A video built from $\mathbf{z}_\text{base}$ alone would show blurred frames and motion that steps between the sampled frames rather than flowing through them, precisely what $\mathbf{z}_\text{res}$ was trained to restore.

\input{supple/figures/3_latent_pca}
\noindent\textbf{PCA visualization.}
Fig.~\ref{fig:pca_split} visualizes the Principal Component Analysis (PCA) of $\mathbf{z}_\text{base}$ and $\mathbf{z}_\text{res}$ across video frames, encoded from a $480{\times}832{\times}81$ video. The three leading components are computed per clip and mapped to RGB, so colors are not comparable across panels. The components of $\mathbf{z}_\text{base}$ are semantically organized, following the objects in the frame. Without $\mathcal{L}_\text{align}$, noise dominates the components of $\mathbf{z}_\text{res}$, which show little of the scene. $\mathcal{L}_\text{align}$ suppresses much of that noise and brings out the objects, and $\mathbf{z}_\text{res}$ occasionally resolves them at an even coarser scale than $\mathbf{z}_\text{base}$. Note that $\mathbf{z}_\text{base}$ is identical in both settings, since $\mathcal{E}$ is frozen.

\subsection{Convergence Behavior}
\label{app:loss_split}
\input{supple/figures/13_convergence_behavior}
Fig.~\ref{fig:convergence} tracks the flow matching loss on the residual channels during DiT adaptation. Fig.~\ref{fig:loss_split} adds the base channels of the same runs, trained under identical settings except for $\mathcal{L}_\text{align}$, with each part standardized by its own statistics. The base channels behave almost identically with and without alignment, since the DiT already models that space and has little to adapt. The two runs separate only on the residual, which converges faster and reaches a lower loss with alignment, so the gain in Fig.~\ref{fig:convergence} does not come at the cost of the base.

\clearpage
\section{Computational Cost}
\label{app:cost}
Tab.~\ref{tab:cost} breaks down the training cost of each stage and the inference latency of the pretrained pipeline before compression and ours.
\input{supple/tables/train_cost}

\vspace{10pt}
\section{More Discussion on Related Work}
\label{app:related}

\noindent\textbf{Improving the latent space for generation.}
Beyond compression, several works improve the latent space itself so that the diffusion model learns it more easily~\citep{zheng2025diffusiontransformersrepresentationautoencoders, chen2025dcae15acceleratingdiffusion, wu2024improvedvideovaelatent}. These works mostly train the generation model from scratch on the resulting latent, whereas we shape the latent so that an already trained model can be reused.

\noindent\textbf{Efficient video autoencoder architectures.}
Compared to images, video carries an additional temporal dimension, which makes the autoencoder both harder to design and more expensive to run. Wavelet-based designs reduce this cost by replacing part of the convolutional stack with a fixed multi-resolution transform~\citep{li2025wfvaeenhancingvideovae, cheng2025leanvaeultraefficientreconstructionvae, nvidia2025cosmosworldfoundationmodel}. These reduce the cost of the autoencoder itself rather than the token count the diffusion model processes.

\section{Additional Qualitative Results}
\label{app:add_qual}
This section collects the qualitative samples referenced throughout the paper: image-to-video and text-to-video generation at $480{\times}832$ (Figs.~\ref{fig:gen_i2v_qual_supp_1}--\ref{fig:gen_t2v_qual_supp_2}), text-to-video generation at $736{\times}1280$ (Figs.~\ref{fig:highres_qual} and~\ref{fig:highres_qual_2}), image-to-video generation at $736{\times}1280$ against LTX-Video~\citep{hacohen2024ltxvideorealtimevideolatent} (Figs.~\ref{fig:i2v_ltx_comp_0}--\ref{fig:i2v_ltx_comp_2}), generation in various styles (Fig.~\ref{fig:style_qual}), and further latent visualizations (Figs.~\ref{fig:pca_supp} and~\ref{fig:pca_supp_2}).

\input{supple/figures/5_gen_i2v_qual}
\input{supple/figures/17_gen_i2v_qual_2}
\input{supple/figures/6_gen_t2v_qual}
\input{supple/figures/18_gen_t2v_qual_2}
\input{supple/figures/7_high_res_qual}
\input{supple/figures/19_high_res_qual_2}
\input{supple/figures/20_i2v_ltx_comp_0}
\input{supple/figures/21_i2v_ltx_comp_1}
\input{supple/figures/22_i2v_ltx_comp_2}
\input{supple/figures/16_style_qual}
\label{app:add_pca_qual}
\input{supple/figures/9_pca_qual}
\input{supple/figures/9_pca_qual_1}

\section{Limitations}
\label{app:limitations}
Compressing the latent leaves fewer tokens for each frame, and small objects suffer the most. Distant faces, text on signs, and thin structures are often lost or broken, while the overall scene and its motion stay intact. The effect shows in Figs.~\ref{fig:gen_t2v_qual_supp_1} and \ref{fig:gen_t2v_qual_supp_2}, and it is weaker at $736{\times}1280$, where the same compression ratio leaves more tokens per frame. GRACE-VAE is also optimized for the DiT rather than for reconstruction alone, and trades some reconstruction fidelity for generation quality. At the same token count, its PSNR is 1.13 dB below our single-latent baseline (Tab.~\ref{tab:recon}). Adding more residual channels could recover more of this detail, but adapting a pretrained DiT to a higher-dimensional latent is itself difficult~\citep{zheng2026opensora20trainingcommerciallevel}, and we leave this to future work.

Applying GRACE to a new pretrained pipeline also requires training a new autoencoder, since the base latent comes from that pipeline's frozen encoder and $\mathcal{L}_\text{align}$ is supervised by its DiT. Each new pipeline therefore costs one more Stage 1 run, 8.5 H200 GPU days in our setting.

%% file: supple/figures/0_ae_arch.tex
\begin{figure}[H]
  \centering
  \vspace{-10pt}
  \includegraphics[width=0.7\linewidth]{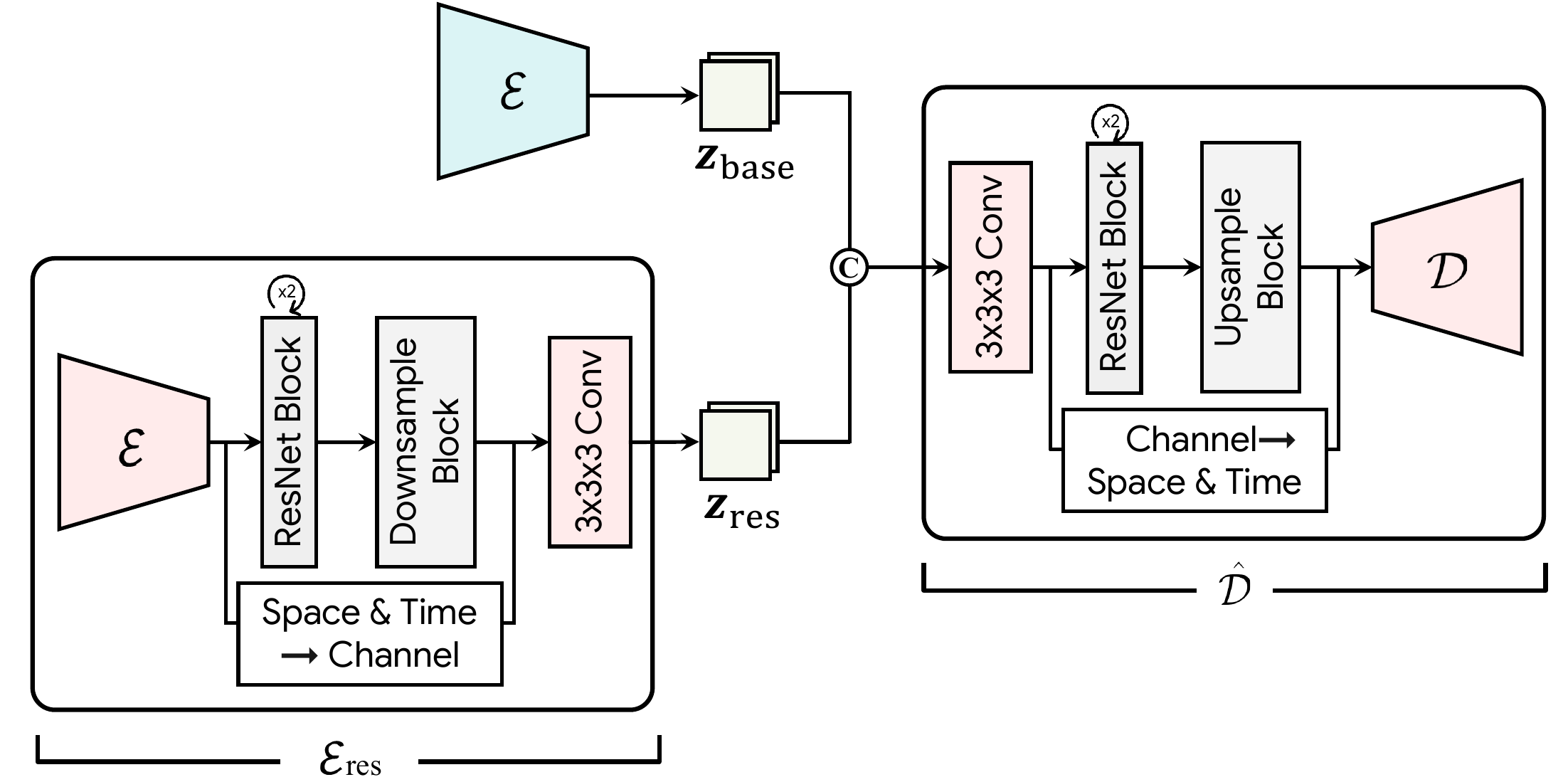}
  \caption{\textbf{Detailed autoencoder architecture.} $\mathcal{E}_\text{res}$ adds a downsampling stage between its middle blocks and head, consisting of residual blocks and a strided causal convolution, paired with a parameter-free shortcut that folds space and time into channels.}
  \label{fig:ae_arch}
  \vspace{-15pt}
\end{figure}

%% file: supple/tables/ae_training_details.tex
\begin{wraptable}{r}{0.55\linewidth}
\vspace{-12pt}
\centering
\small
\captionof{table}{\textbf{Stage 1 hyperparameters.} All settings follow the training recipe of the pretrained autoencoder.}
\label{tab:hp_ae}
\setlength{\tabcolsep}{4pt}
\resizebox{\linewidth}{!}{%
\begin{tabular}{c l c c}
\toprule
& Hyperparameter & Phase 1 & Phase 2 \\
\midrule
\multirow{3}{*}{Architecture}
& pretrained autoencoder & Wan2.1-VAE & Wan2.1-VAE \\
& $(C, C')$              & $(16, 16)$ & $(16, 16)$ \\
& $(r_s, r_t)$           & $(2, 2)$   & $(2, 2)$ \\
\midrule
\multirow{9}{*}{Training setup}
& input shape          & $256{\times}256{\times}81$ & $512{\times}512{\times}81$ \\
& trained modules      & $\mathcal{E}_\text{res}$, $\hat{\mathcal{D}}$ & $\hat{\mathcal{D}}$ \\
& optimizer            & AdamW & AdamW \\
& learning rate        & 8e-5 & 4e-5 \\
& betas                & $(0.9, 0.999)$ & $(0.9, 0.999)$ \\
& weight decay         & 1e-4 & 1e-4 \\
& scheduler            & constant & constant \\
& precision            & bf16 & bf16 \\
& effective batch size & 32 & 64 \\
\midrule
\multirow{3}{*}{$\mathcal{L}_\text{recon}$}
& $\lambda_\text{L1}$    & 1.0  & 1.0 \\
& $\lambda_\text{LPIPS}$ & 3.0  & 3.0 \\
& $\lambda_\text{KL}$    & 3e-6 & 3e-6 \\
\midrule
\multirow{4}{*}{$\mathcal{L}_\text{align}$}
& reference DiT              & Wan2.1-14B & -- \\
& alignment depth            & 10   & -- \\
& $P_l$ bottleneck           & 16   & -- \\
& added-module learning rate & 1e-4 & -- \\
\bottomrule
\end{tabular}%
}
\end{wraptable}

%% file: supple/tables/dit_training_details.tex
\begin{wraptable}{r}{0.55\linewidth}
\vspace{-12pt}
\centering
\small
\captionof{table}{\textbf{Stage 2 hyperparameters.} Offset and shift are sampled from a range in training, fixed at inference.}
\label{tab:hp_dit}
\setlength{\tabcolsep}{4pt}
\resizebox{\linewidth}{!}{%
\begin{tabular}{c l c c}
\toprule
& Hyperparameter & I2V & T2V \\
\midrule
\multirow{7}{*}{Architecture}
& pretrained DiT & Wan2.1-I2V-14B & Wan2.1-T2V-14B \\
& input dim      & 72   & 32 \\
& hidden dim     & 5120 & 5120 \\
& blocks         & 40   & 40 \\
& num. heads     & 40   & 40 \\
& patch size     & 2    & 2 \\
& $(C, C')$      & $(16, 16)$ & $(16, 16)$ \\
\midrule
\multirow{6}{*}{Training setup}
& input shape          & $480{\times}832{\times}81$ & $480{\times}832{\times}81$ \\
& optimizer            & AdamW & AdamW \\
& learning rate        & 1e-4  & 1e-4  \\
& effective batch size & 128   & 128 \\
& offset $\delta$      & $[0, 0.6]$ & $[0, 0.6]$ \\
& shift $s$            & $[2, 5]$   & $[2, 5]$ \\
\midrule
\multirow{3}{*}{LoRA}
& rank                  & 512 & 512 \\
& $\alpha$              & 512 & 512 \\
& image cross-attention & \cmark & \xmark \\
\midrule
\multirow{4}{*}{Sampling}
& steps           & 50 & 50 \\
& guidance scale  & 5.0 & 5.0 \\
& offset $\delta$ & 0.15 & 0.15 \\
& shift $s$       & 3 & 3 \\
\bottomrule
\end{tabular}%
}
\vspace{-20pt}
\end{wraptable}

%% file: supple/tables/gen_add_Eval_i2v_t2v_merged.tex
\begin{table}[p]
\vspace*{\fill}
\centering
\small
\setlength{\tabcolsep}{3pt}
\caption{\textbf{Per-dimension VBench~\citep{huang2023vbenchcomprehensivebenchmarksuite, huang2024vbenchcomprehensiveversatilebenchmark} scores} at $480{\times}832{\times}81$. \emph{Single-latent Baseline} is our baseline without the dual latent or the alignment loss. The best score in each row is in bold. $^\dagger$Reported for reference only, as the official VBench-I2V quality score does not include this dimension.}
\label{tab:vbench_dims}
\begin{minipage}{\linewidth}
\centering
\subcaption{VBench-T2V}
\label{tab:vbench_t2v}
\resizebox{0.95\linewidth}{!}{%
\begin{tabular}{l c c c c c c}
\toprule
\multirow{2}{*}[-0.72\normalbaselineskip]{Dimension} & \multirow{2}{*}[-0.72\normalbaselineskip]{Wan2.1-14B} & \multirow{2}{*}[-0.72\normalbaselineskip]{Open-Sora 2.0} & \multirow{2}{*}[-0.72\normalbaselineskip]{LTX-Video 0.9.7} & \multicolumn{3}{c}{Wan2.1-14B +} \\
\cmidrule(lr){5-7}
 & & & & DC-Gen & \makecell{Single-latent\\Baseline} & \textbf{GRACE (Ours)} \\
\midrule
Latent tokens & 32.8k & 8.2k & 4.3k & 8.2k & 4.3k & 4.3k \\
Latency (s) $\downarrow$ & 851.5 & 138.6 & 99.6 & 157.1 & 76.1 & \textbf{75.8} \\
\midrule
\multicolumn{7}{l}{\emph{Quality}} \\
\quad Subject consistency & 96.48 & 94.55 & 96.74 & 96.42 & 94.91 & \textbf{96.76} \\
\quad Background consistency & 98.14 & 96.98 & 96.20 & 97.88 & 98.04 & \textbf{98.70} \\
\quad Temporal flickering & 99.20 & 98.98 & 99.35 & 99.25 & 99.22 & \textbf{99.42} \\
\quad Motion smoothness & 98.71 & 99.28 & \textbf{99.45} & 97.74 & 99.20 & 99.36 \\
\quad Dynamic degree & 56.94 & 58.33 & 51.39 & \textbf{70.83} & 47.22 & 62.50 \\
\quad Aesthetic quality & 70.41 & 56.09 & 60.62 & \textbf{71.06} & 64.77 & 68.66 \\
\quad Imaging quality & 67.55 & 41.81 & 64.18 & 67.50 & 65.10 & \textbf{67.65} \\
\multicolumn{7}{l}{\emph{Semantic}} \\
\quad Object class & 92.09 & 89.08 & 81.33 & 92.25 & 88.77 & \textbf{94.15} \\
\quad Multiple objects & 79.50 & 61.28 & 40.62 & 84.68 & 72.94 & \textbf{85.67} \\
\quad Human action & 97.00 & 98.00 & 86.00 & 97.00 & 98.00 & \textbf{99.00} \\
\quad Color & 89.68 & 69.01 & 72.10 & 88.36 & 86.41 & \textbf{97.29} \\
\quad Spatial relationship & 78.94 & 57.87 & 58.16 & 79.18 & 79.23 & \textbf{91.16} \\
\quad Scene & 47.31 & 45.35 & 39.97 & 52.91 & 51.60 & \textbf{57.49} \\
\quad Appearance style & 22.57 & 24.12 & 20.19 & 22.84 & 23.12 & \textbf{24.79} \\
\quad Temporal style & 23.34 & 24.01 & 21.10 & 22.49 & 23.10 & \textbf{25.52} \\
\quad Overall consistency & 25.63 & 25.60 & 23.08 & 25.39 & 24.48 & \textbf{25.76} \\
\midrule
Quality score & 85.24 & 78.80 & 82.88 & 85.85 & 83.21 & \textbf{86.02} \\
Semantic score & 78.70 & 72.35 & 64.31 & 79.70 & 77.75 & \textbf{84.98} \\
\textbf{Total score} & 83.93 & 77.51 & 79.17 & 84.62 & 82.12 & \textbf{85.81} \\
\bottomrule
\end{tabular}%
}
\end{minipage}

\vspace{12pt}

\begin{minipage}{\linewidth}
\centering
\subcaption{VBench-I2V}
\label{tab:vbench_i2v}
\resizebox{0.95\linewidth}{!}{%
\begin{tabular}{l c c c c c c}
\toprule
\multirow{2}{*}[-0.72\normalbaselineskip]{Dimension} & \multirow{2}{*}[-0.72\normalbaselineskip]{Wan2.1-14B} & \multirow{2}{*}[-0.72\normalbaselineskip]{Open-Sora 2.0} & \multirow{2}{*}[-0.72\normalbaselineskip]{LTX-Video 0.9.7} & \multicolumn{3}{c}{Wan2.1-14B +} \\
\cmidrule(lr){5-7}
 & & & & DC-Gen & \makecell{Single-latent\\Baseline} & \textbf{GRACE (Ours)} \\
\midrule
Latent tokens & 32.8k & 8.2k & 4.3k & 8.2k & 4.3k & 4.3k \\
Latency (s) $\downarrow$ & 863.2 & 138.4 & 104.1 & 165.0 & 77.8 & \textbf{77.7} \\
\midrule
\multicolumn{7}{l}{\emph{I2V}} \\
\quad Video-image subject fidelity & 98.92 & 94.58 & \textbf{98.97} & 90.97 & 96.72 & 98.37 \\
\quad Video-image background fidelity & \textbf{99.51} & 94.03 & 99.20 & 93.37 & 97.94 & 99.28 \\
\quad Camera motion & 31.59 & \textbf{58.85} & 30.93 & 48.89 & 33.29 & 33.94 \\
\multicolumn{7}{l}{\emph{Quality}} \\
\quad Subject consistency & 96.74 & 93.70 & \textbf{97.87} & 93.30 & 94.79 & 95.17 \\
\quad Background consistency & 97.79 & 95.48 & \textbf{98.39} & 96.94 & 97.49 & 97.58 \\
\quad Motion smoothness & 98.73 & 98.75 & \textbf{99.52} & 97.05 & 99.11 & 99.11 \\
\quad Dynamic degree & 25.20 & 42.68 & 24.80 & \textbf{59.76} & 32.52 & 38.62 \\
\quad Aesthetic quality & \textbf{66.69} & 57.17 & 64.59 & 61.60 & 62.34 & 64.73 \\
\quad Imaging quality & \textbf{71.10} & 61.72 & 70.24 & 69.85 & 68.79 & 68.82 \\
\quad Temporal flickering$^\dagger$ & 98.04 & 97.72 & \textbf{99.18} & 95.75 & 98.43 & 97.91 \\
\midrule
I2V score & \textbf{95.82} & 91.17 & 95.62 & 88.25 & 93.67 & 95.48 \\
Quality score & 80.01 & 76.96 & \textbf{80.32} & 80.01 & 79.21 & 80.31 \\
\textbf{Total score} & 87.92 & 84.07 & \textbf{87.97} & 84.13 & 86.44 & 87.90 \\
\bottomrule
\end{tabular}%
}
\end{minipage}
\vspace*{\fill}
\end{table}

%% file: supple/tables/gen_add_eval_high_res.tex
\begin{table}[p]
\vspace*{\fill}
\centering
\small
\setlength{\tabcolsep}{3pt}
\caption{\textbf{Per-dimension VBench~\citep{huang2023vbenchcomprehensivebenchmarksuite, huang2024vbenchcomprehensiveversatilebenchmark} scores} at $736{\times}1280{\times}81$, following Tab.~\ref{tab:vbench_dims}. The best score in each row is in bold. $^\ddagger$Measured with spatial tiling in the VAE encoder to avoid running out of memory when encoding the conditioning image at this resolution.}
\label{tab:vbench_dims_highres}
\begin{minipage}{\linewidth}
\centering
\subcaption{VBench-T2V}
\label{tab:vbench_t2v_highres}
\resizebox{0.95\linewidth}{!}{%
\begin{tabular}{l c c c c c}
\toprule
\multirow{2}{*}[-0.25\normalbaselineskip]{Dimension} & \multirow{2}{*}[-0.25\normalbaselineskip]{Wan2.1-14B} & \multirow{2}{*}[-0.25\normalbaselineskip]{Open-Sora 2.0} & \multirow{2}{*}[-0.25\normalbaselineskip]{LTX-Video 0.9.7} & \multicolumn{2}{c}{Wan2.1-14B +} \\
\cmidrule(lr){5-6}
 & & & & DC-Gen & \textbf{GRACE (Ours)} \\
\midrule
Latent tokens & 77.3k & 19.3k & 10.1k & 19.3k & 10.1k \\
Latency (s) $\downarrow$ & 3361.3 & 425.4 & 264.2 & 456.4 & \textbf{215.6} \\
\midrule
\multicolumn{6}{l}{\emph{Quality}} \\
\quad Subject consistency & 94.78 & 94.17 & 92.96 & \textbf{96.72} & 95.10 \\
\quad Background consistency & 97.98 & 97.54 & 95.46 & 97.92 & \textbf{98.48} \\
\quad Temporal flickering & 98.98 & 98.81 & 98.24 & 99.12 & \textbf{99.43} \\
\quad Motion smoothness & 98.70 & \textbf{99.18} & 98.92 & 97.74 & \textbf{99.18} \\
\quad Dynamic degree & 58.33 & 52.78 & \textbf{88.89} & 72.22 & 59.72 \\
\quad Aesthetic quality & 68.16 & 58.52 & 59.89 & \textbf{70.55} & 67.85 \\
\quad Imaging quality & 68.39 & 55.18 & 66.67 & 68.80 & \textbf{68.81} \\
\multicolumn{6}{l}{\emph{Semantic}} \\
\quad Object class & 88.21 & 87.90 & 76.03 & 87.58 & \textbf{90.27} \\
\quad Multiple objects & 77.29 & 69.97 & 34.53 & \textbf{82.32} & 78.96 \\
\quad Human action & 97.00 & \textbf{100.00} & 86.00 & 96.00 & \textbf{100.00} \\
\quad Color & 86.60 & 81.18 & 77.60 & 89.96 & \textbf{90.29} \\
\quad Spatial relationship & 67.01 & 73.88 & 53.98 & 81.17 & \textbf{81.51} \\
\quad Scene & 46.29 & 52.25 & 41.42 & 49.27 & \textbf{55.67} \\
\quad Appearance style & 22.20 & 24.73 & 19.87 & 22.52 & \textbf{24.74} \\
\quad Temporal style & 23.04 & 24.76 & 21.48 & 22.78 & \textbf{25.71} \\
\quad Overall consistency & 25.73 & \textbf{26.35} & 23.72 & 25.78 & \textbf{26.35} \\
\midrule
Quality score & 84.69 & 80.73 & 84.46 & \textbf{86.08} & 85.42 \\
Semantic score & 76.01 & 78.16 & 63.58 & 78.80 & \textbf{82.04} \\
\textbf{Total score} & 82.96 & 80.22 & 80.29 & 84.62 & \textbf{84.74} \\
\bottomrule
\end{tabular}%
}
\end{minipage}

\vspace{12pt}

\begin{minipage}{\linewidth}
\centering
\subcaption{VBench-I2V}
\label{tab:vbench_i2v_highres}
\resizebox{0.95\linewidth}{!}{%
\begin{tabular}{l c c c c c}
\toprule
\multirow{2}{*}[-0.25\normalbaselineskip]{Dimension} & \multirow{2}{*}[-0.25\normalbaselineskip]{Wan2.1-14B} & \multirow{2}{*}[-0.25\normalbaselineskip]{Open-Sora 2.0} & \multirow{2}{*}[-0.25\normalbaselineskip]{LTX-Video 0.9.7} & \multicolumn{2}{c}{Wan2.1-14B +} \\
\cmidrule(lr){5-6}
 & & & & DC-Gen & \textbf{GRACE (Ours)} \\
\midrule
Latent tokens & 77.3k & 19.3k & 10.1k & 19.3k & 10.1k \\
Latency (s) $\downarrow$ & 3396.8 & 425.4 & 274.6 & 550.7$^\ddagger$ & \textbf{218.8} \\
\midrule
\multicolumn{6}{l}{\emph{I2V}} \\
\quad Video-image subject fidelity & 98.55 & 96.65 & \textbf{98.88} & 94.62 & 98.34 \\
\quad Video-image background fidelity & 99.22 & 97.09 & 98.95 & 96.50 & \textbf{99.42} \\
\quad Camera motion & 34.34 & \textbf{46.66} & 37.35 & 43.25 & 33.55 \\
\multicolumn{6}{l}{\emph{Quality}} \\
\quad Subject consistency & 95.03 & 95.17 & \textbf{96.76} & 94.07 & 95.95 \\
\quad Background consistency & 96.69 & 96.71 & 96.88 & 97.62 & \textbf{98.04} \\
\quad Motion smoothness & 98.22 & 98.96 & \textbf{99.23} & 97.83 & 99.17 \\
\quad Dynamic degree & 41.87 & 26.42 & 49.59 & \textbf{57.72} & 30.89 \\
\quad Aesthetic quality & \textbf{65.30} & 60.03 & 64.07 & 61.67 & 64.90 \\
\quad Imaging quality & 70.44 & 67.61 & \textbf{70.78} & 70.26 & 69.83 \\
\quad Temporal flickering$^\dagger$ & 96.92 & 98.04 & 97.90 & 96.70 & \textbf{98.50} \\
\midrule
I2V score & 95.56 & 93.72 & \textbf{95.71} & 92.04 & 95.54 \\
Quality score & 80.20 & 77.82 & \textbf{81.79} & 80.73 & 80.15 \\
\textbf{Total score} & 87.88 & 85.77 & \textbf{88.75} & 86.39 & 87.84 \\
\bottomrule
\end{tabular}%
}
\end{minipage}
\vspace*{\fill}
\end{table}

%% file: supple/tables/user_study.tex
\begin{table}[H]
\centering
\small
\caption{\textbf{Human evaluation.}
Participants compared each pair of videos without knowing which model produced which.
Each cell gives the percentage of votes preferring our model (Ours), rating both videos about equal (Equal), or preferring the baseline named in the column.}
\label{tab:human_eval}
\setlength{\tabcolsep}{5pt}
\newcommand{\wanhead}{vs.\ Wan2.1-14B~\citep{wan2025wanopenadvancedlargescale} (\%)}
\newcommand{\dchead}{vs.\ DC-Gen~\citep{he2026dc} (\%)}
\settowidth{\wanw}{\wanhead}
\settowidth{\dcw}{\dchead}
\setlength{\wanw}{\dimexpr(\wanw-4\tabcolsep)/3\relax}
\setlength{\dcw}{\dimexpr(\dcw-4\tabcolsep)/3\relax}
\settowidth{\dimen0}{Wan2.1-14B}\ifdim\dimen0>\wanw\setlength{\wanw}{\dimen0}\fi 
\resizebox{\linewidth}{!}{%
\begin{tabular}{l l *{3}{>{\centering\arraybackslash}p{\wanw}} *{3}{>{\centering\arraybackslash}p{\dcw}}}
\toprule
\multirow{2}{*}[-0.25\normalbaselineskip]{Task} & \multirow{2}{*}[-0.25\normalbaselineskip]{Aspect} & \multicolumn{3}{c}{\wanhead} & \multicolumn{3}{c}{\dchead} \\
\cmidrule(lr){3-5} \cmidrule(lr){6-8}
 & & Ours & Equal & \mbox{Wan2.1-14B} & Ours & Equal & DC-Gen \\
\midrule
\multirow{3}{*}{T2V} & Visual quality       & 49.4 & 15.4 & 35.3 & 71.2 & 11.5 & 17.3 \\
                     & Temporal consistency & 46.8 & 21.2 & 32.1 & 64.7 & 21.2 & 14.1 \\
                     & Text alignment       & 52.6 & 21.8 & 25.6 & 62.2 & 23.1 & 14.7 \\
\midrule
\multirow{3}{*}{I2V} & Visual quality       & 23.7 & 32.7 & 43.6 & 60.3 & 16.7 & 23.1 \\
                     & Temporal consistency & 29.5 & 25.6 & 44.9 & 64.1 & 11.5 & 24.4 \\
                     & Text alignment       & 22.4 & 42.9 & 34.6 & 46.8 & 29.5 & 23.7 \\
\bottomrule
\end{tabular}%
}
\end{table}
\vspace{-10pt}
    

%% file: supple/figures/14_user_study_interface_T2V.tex
\begin{figure}[H]
  \centering
  \includegraphics[width=\linewidth]{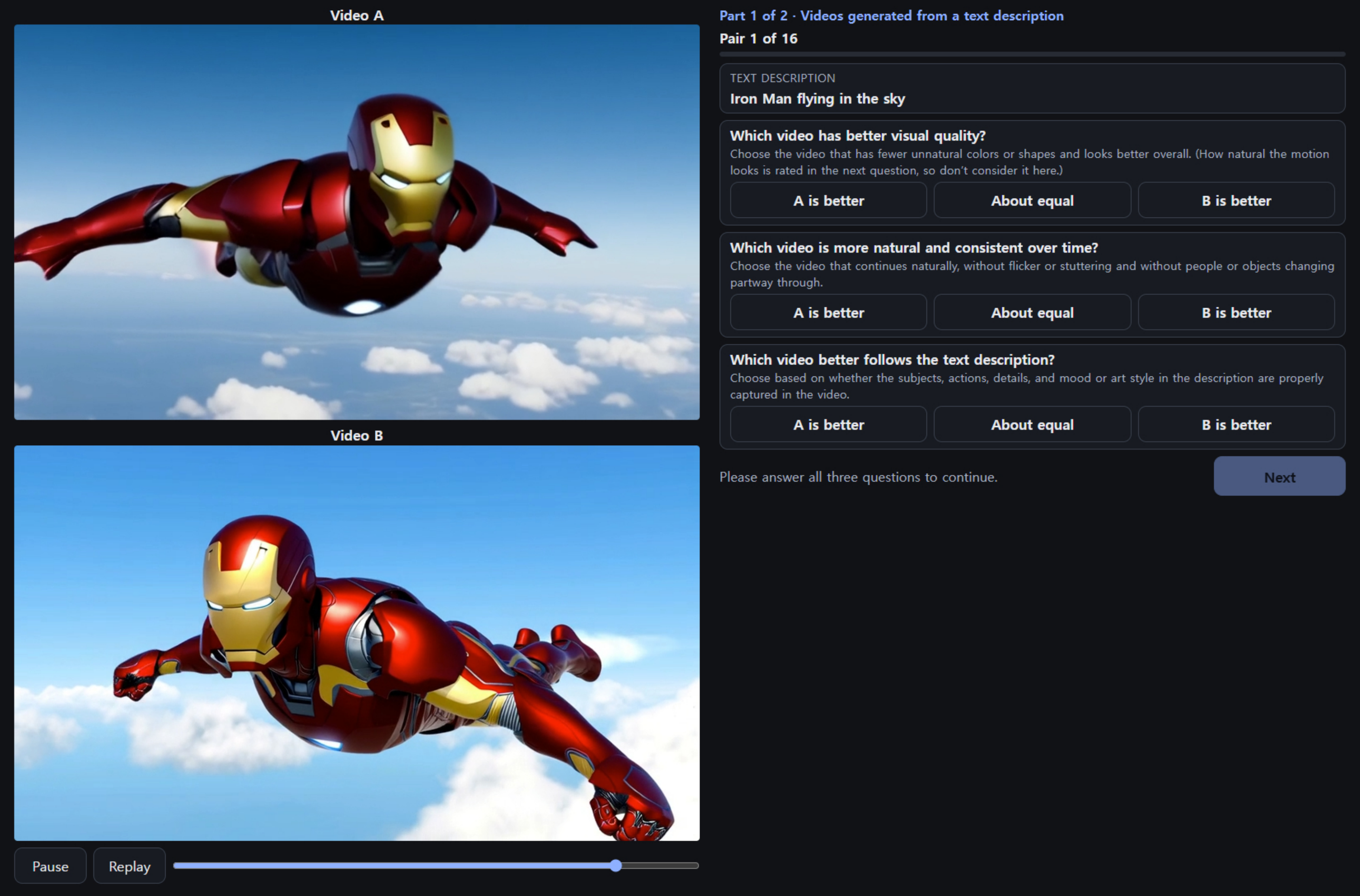}
  \caption{\textbf{User study interface for T2V samples.} }
\label{fig:user_study_T2V}
\end{figure}

%% file: supple/figures/15_user_study_interface_I2V.tex
\begin{figure}[H]
  \centering
  \includegraphics[width=\linewidth]{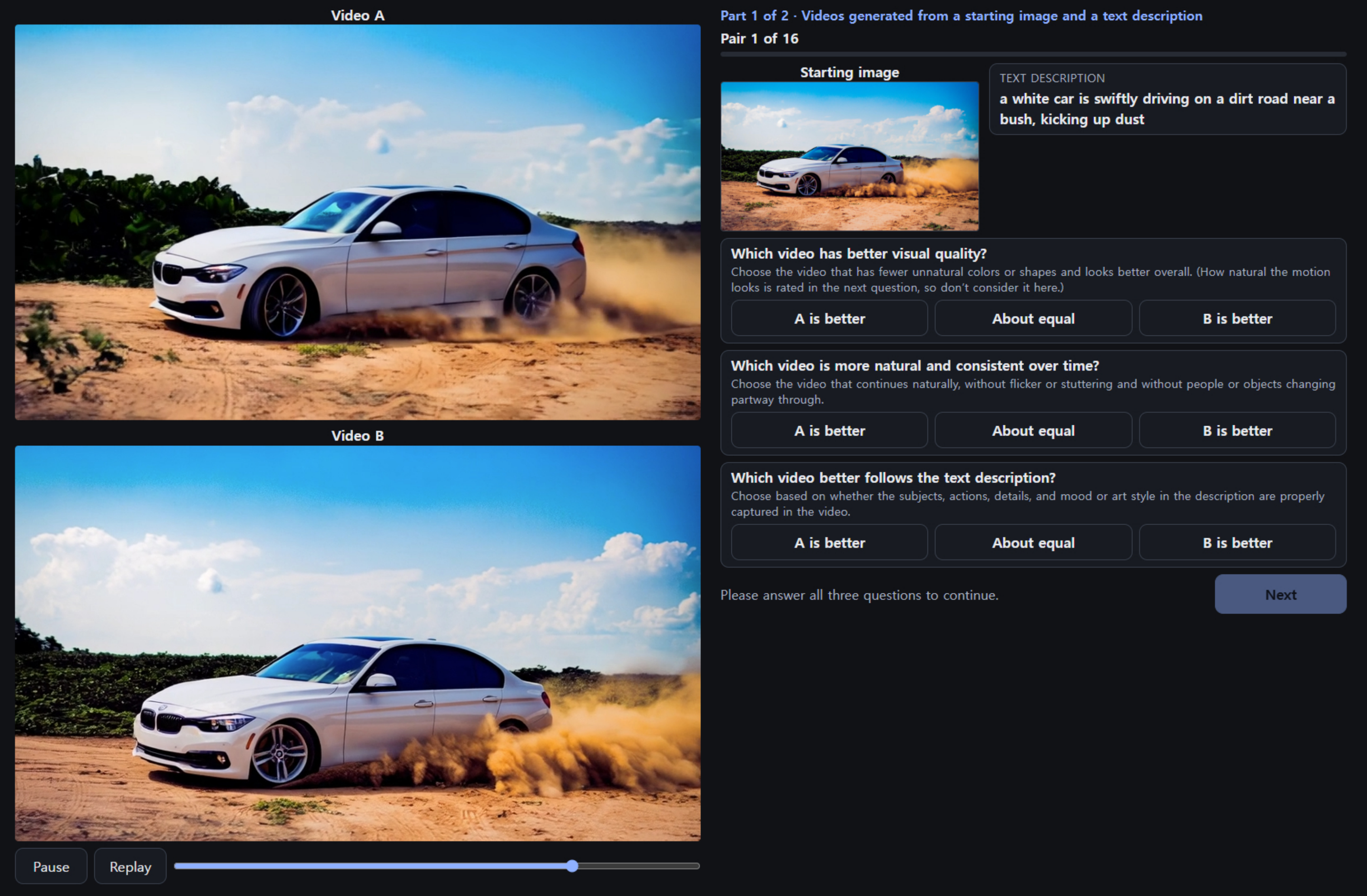}
  \caption{\textbf{User study interface for I2V samples.} }
\label{fig:user_study_I2V}
\end{figure}

%% file: supple/tables/first_frame.tex
\begin{wraptable}{r}{0.43\linewidth}
\vspace{-12pt}
\centering
\small
\captionof{table}{\textbf{Ablation on first-frame conditioning.}}
\label{tab:firstframe}
\setlength{\tabcolsep}{5pt}
\begin{tabular}{c c c c}
\toprule
First frame & PSNR $\uparrow$ & SSIM $\uparrow$ & LPIPS $\downarrow$ \\
\midrule
\xmark &  31.97 & 0.928 & 0.038 \\
\cmark & 32.63 & 0.930 & 0.032 \\
\bottomrule
\end{tabular}
\end{wraptable}

%% file: supple/tables/align_depth.tex
\begin{wraptable}{r}{0.55\linewidth}
\vspace{-10pt}
\centering
\small
\captionof{table}{\textbf{Ablation on alignment depth.}}
\label{tab:depth}
\setlength{\tabcolsep}{4pt}
\resizebox{\linewidth}{!}{%
\begin{tabular}{c c c c c c c}
\toprule
\multirow{2}{*}[-0.25\normalbaselineskip]{Depth} & \multicolumn{3}{c}{Reconstruction} & \multicolumn{3}{c}{VBench-I2V} \\
\cmidrule(lr){2-4} \cmidrule(lr){5-7}
 & PSNR $\uparrow$ & SSIM $\uparrow$ & LPIPS $\downarrow$ & I2V $\uparrow$ & Quality $\uparrow$ & Total $\uparrow$ \\
\midrule
10 & 32.63 & 0.930 & 0.032 & 95.48 & 80.31 & 87.90 \\
40 & 32.21 & 0.929 & 0.038 & 95.70 & 80.19 & 87.94 \\
\bottomrule
\end{tabular}%
}
\end{wraptable}

%% file: supple/figures/2_latent_analysis.tex
\begin{figure}[H]
  \centering
  \includegraphics[width=0.9\linewidth]{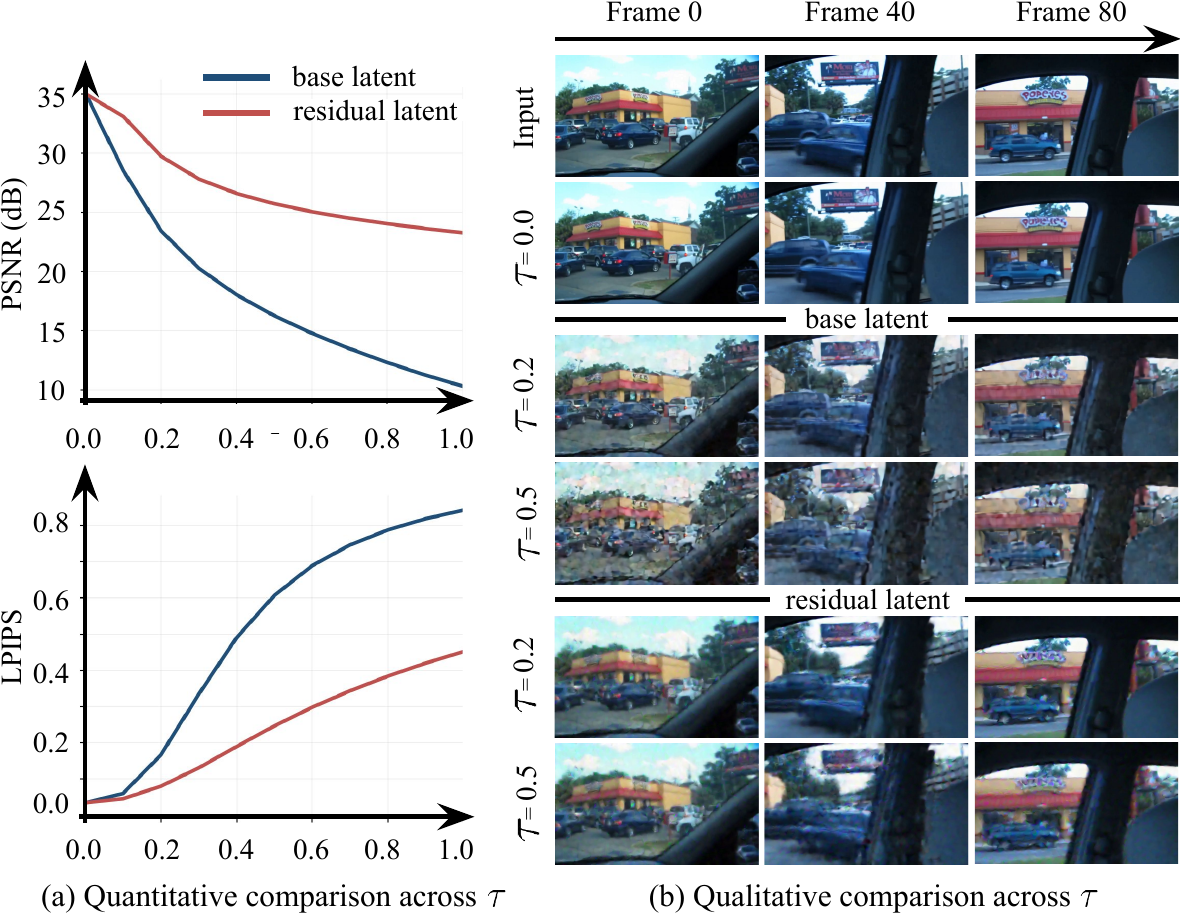}
  %
  %
  %
  %
  %
  %
  %
  %
  %
  %
  %
  \caption{\textbf{Reconstruction under latent noise.} We noise $\mathbf{z}_\text{base}$ or $\mathbf{z}_\text{res}$ at level $\tau$, leave the other unchanged, and reconstruct. \textbf{(a)} Reconstruction quality on Panda-70M~\citep{chen2024panda70mcaptioning70mvideos} at $480{\times}832{\times}81$ as $\tau$ grows, measured in PSNR and LPIPS. \textbf{(b)} Reconstructed frames at two levels of $\tau$, shown below the input and the noise-free reconstruction, with the first block noising $\mathbf{z}_\text{base}$ and the second noising $\mathbf{z}_\text{res}$.}
\label{fig:latent_perturbation}
  \vspace{-10pt}
\end{figure}

%% file: supple/figures/3_latent_pca.tex
\begin{figure}[H]
  \centering
  \includegraphics[width=\linewidth]{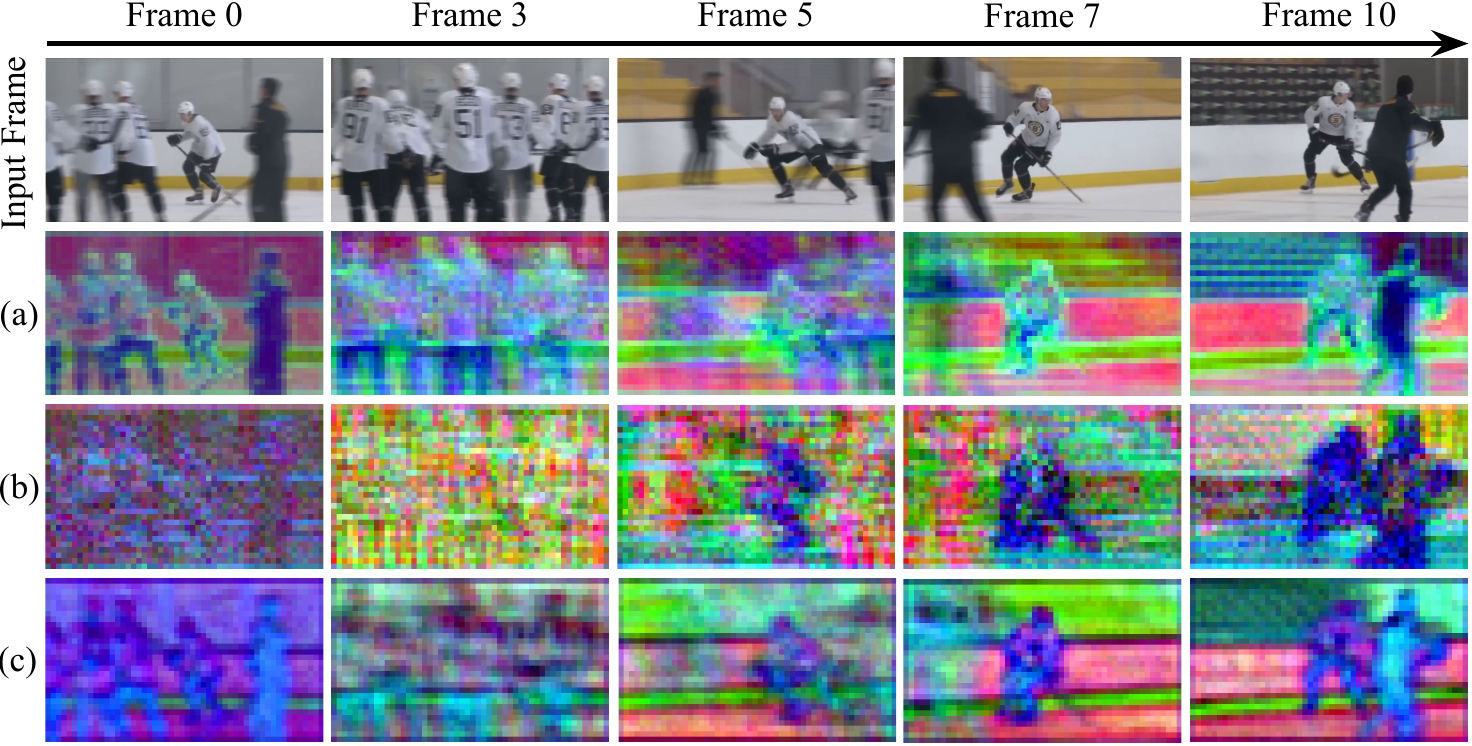}
  \captionof{figure}{\textbf{PCA visualization of the base and residual latents.} Principal Component Analysis (PCA) of each part at f16t8p2, encoded from a $480{\times}832{\times}81$ video, for \textbf{(a)} $\mathbf{z}_\text{base}$, \textbf{(b)} $\mathbf{z}_\text{res}$ without $\mathcal{L}_\text{align}$, and \textbf{(c)} $\mathbf{z}_\text{res}$ with $\mathcal{L}_\text{align}$. Since $\mathcal{E}$ is frozen, $\mathbf{z}_\text{base}$ is the same in both settings.}
\label{fig:pca_split}
  \vspace{-10pt}
\end{figure}

%% file: supple/figures/13_convergence_behavior.tex
\begin{figure}[H]
\centering
\includegraphics[width=0.8\linewidth]{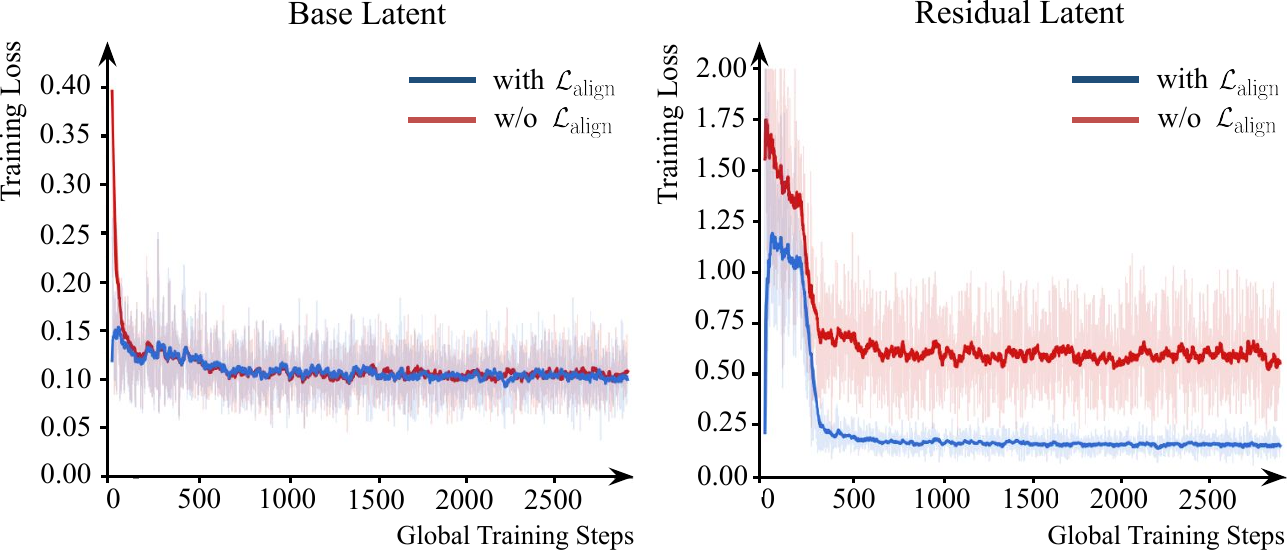}
\vspace{-5pt}
\caption{\textbf{Convergence of the base and the residual.} Flow matching loss on the base channels (left) and the residual channels (right) during DiT adaptation, for the dual latent with and without $\mathcal{L}_\text{align}$. Each part is standardized with its own statistics. Faint curves show the raw loss and solid curves its EMA.}
\label{fig:loss_split}
\vspace{-10pt}
\end{figure}

%% file: supple/tables/train_cost.tex
\begin{table}[H]
\centering
\small
\caption{\textbf{Computational cost.} Training in H200 GPU days, and inference at $480{\times}832{\times}81$ on a single A100 with 50 sampling steps.}
\label{tab:cost}
\setlength{\tabcolsep}{5pt}
\begin{tabular}{l l c c c c}
\toprule
& & \multicolumn{2}{c}{T2V} & \multicolumn{2}{c}{I2V} \\
\cmidrule(lr){3-4} \cmidrule(lr){5-6}
& & Wan2.1-14B & Ours & Wan2.1-14B & Ours \\
\midrule
\multirow{4}{*}{Training}
& Stage 1, full autoencoder     & -- & 6.9 & -- & 6.9 \\
& Stage 1, decoder-only (+EMA)  & -- & 1.6 & -- & 1.6 \\
& Stage 2                       & -- & 30  & -- & 30  \\
& Total (H200 GPU days)         & -- & 38.5 & -- & 38.5 \\
\midrule
\multicolumn{2}{l}{Inference latency (s) $\downarrow$} & 851.5 & \textbf{75.8} & 863.2 & \textbf{77.7} \\
\bottomrule
\end{tabular}
\end{table}

%% file: supple/figures/5_gen_i2v_qual.tex
\begin{figure}[H]
  \centering
  \includegraphics[width=\linewidth]{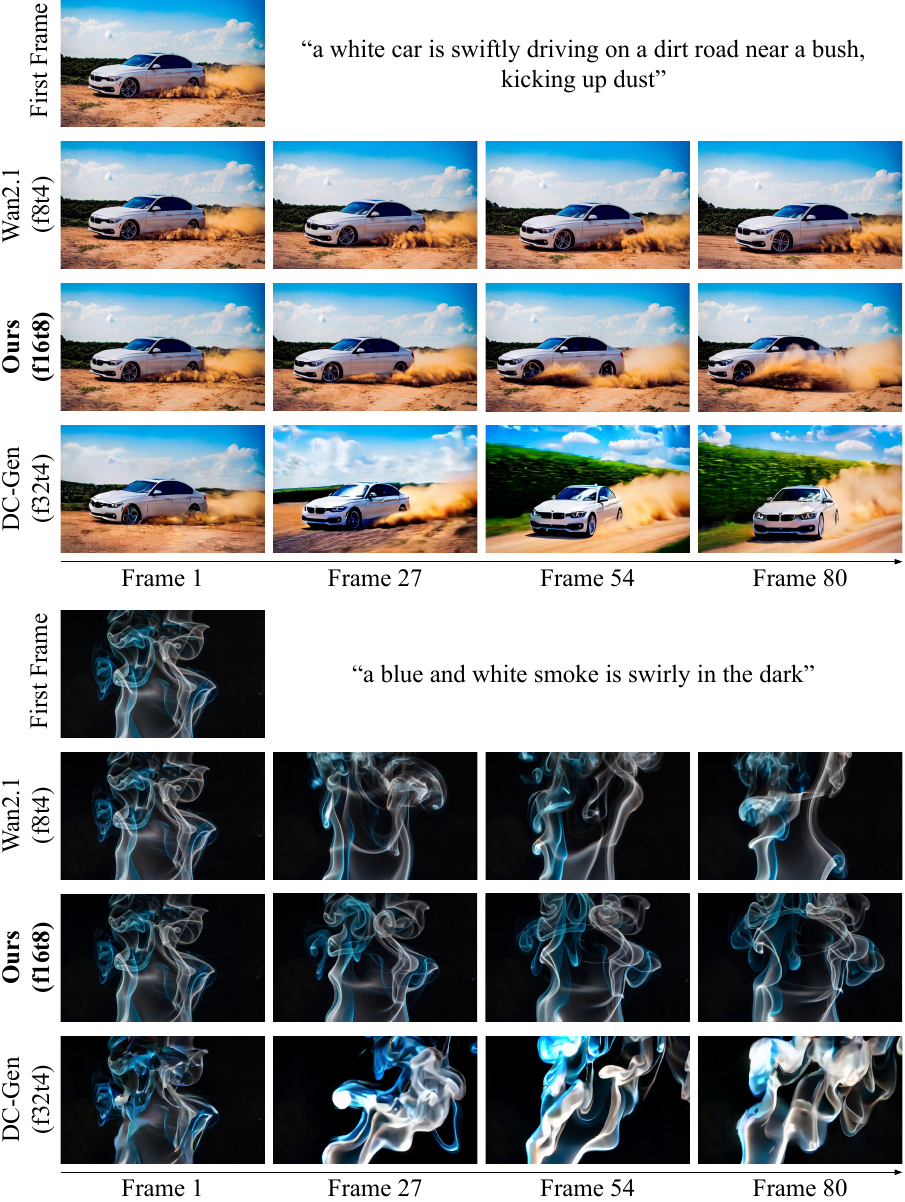}
  \caption{\textbf{Additional image-to-video results.}
    Samples from VBench-I2V~\citep{huang2023vbenchcomprehensivebenchmarksuite, huang2024vbenchcomprehensiveversatilebenchmark} at $480{\times}832{\times}81$, generated by the pretrained Wan2.1-14B~\citep{wan2025wanopenadvancedlargescale} at f8t4p2, ours at f16t8p2, and DC-Gen~\citep{he2026dc} at f32t4p1, from the same conditioning frame and prompt.}
  \label{fig:gen_i2v_qual_supp_1}
\end{figure}

%% file: supple/figures/17_gen_i2v_qual_2.tex
\clearpage
\vspace*{\fill}
\begin{figure}[H]
  \centering
  \includegraphics[width=\linewidth]{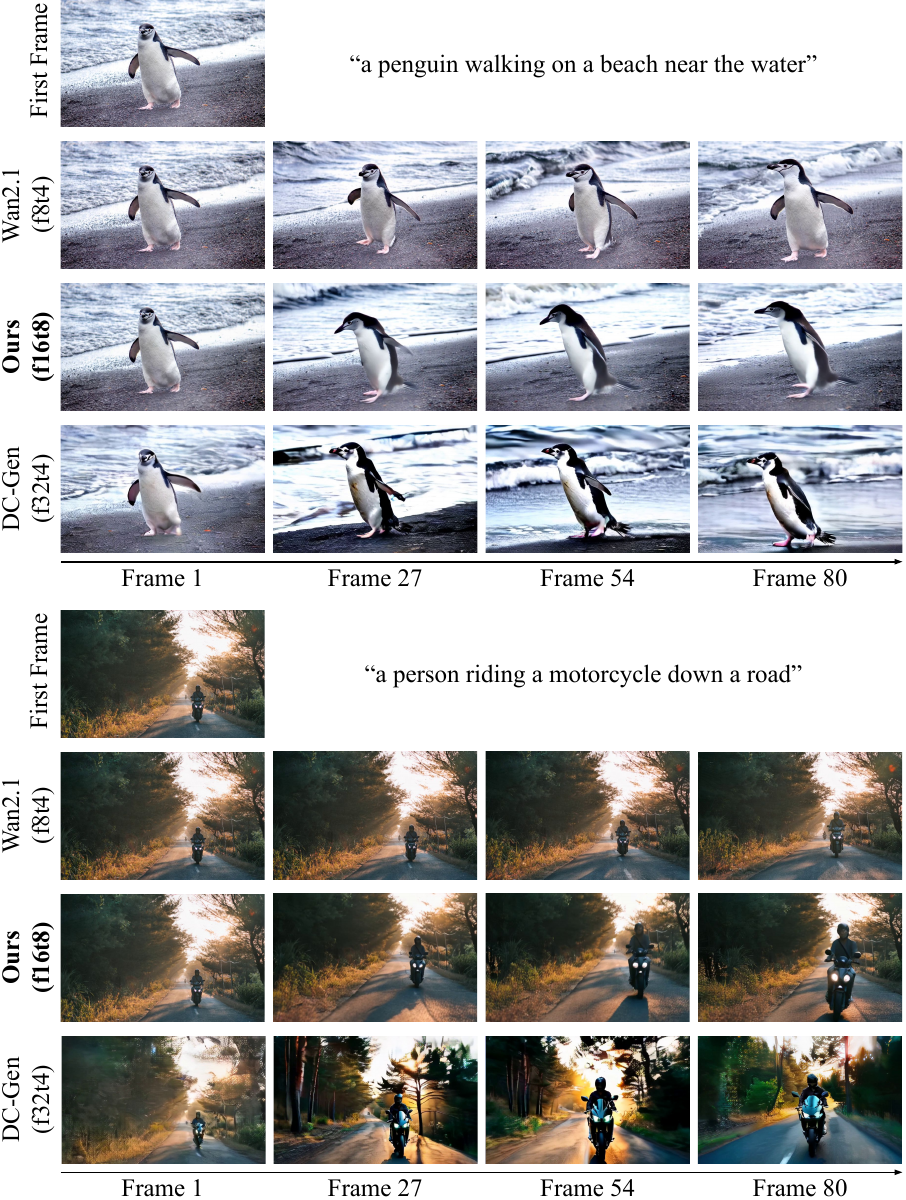}
  \caption{\textbf{Additional image-to-video results (continued).}
    Samples from VBench-I2V~\citep{huang2023vbenchcomprehensivebenchmarksuite, huang2024vbenchcomprehensiveversatilebenchmark} at $480{\times}832{\times}81$, generated by the pretrained Wan2.1-14B~\citep{wan2025wanopenadvancedlargescale} at f8t4p2, ours at f16t8p2, and DC-Gen~\citep{he2026dc} at f32t4p1, from the same conditioning frame and prompt.}
  \label{fig:gen_i2v_qual_supp_2}
\end{figure}
\vspace*{\fill}
\clearpage

%% file: supple/figures/6_gen_t2v_qual.tex
\clearpage
\vspace*{\fill}
\begin{figure}[H]
  \centering
  \includegraphics[width=\linewidth]{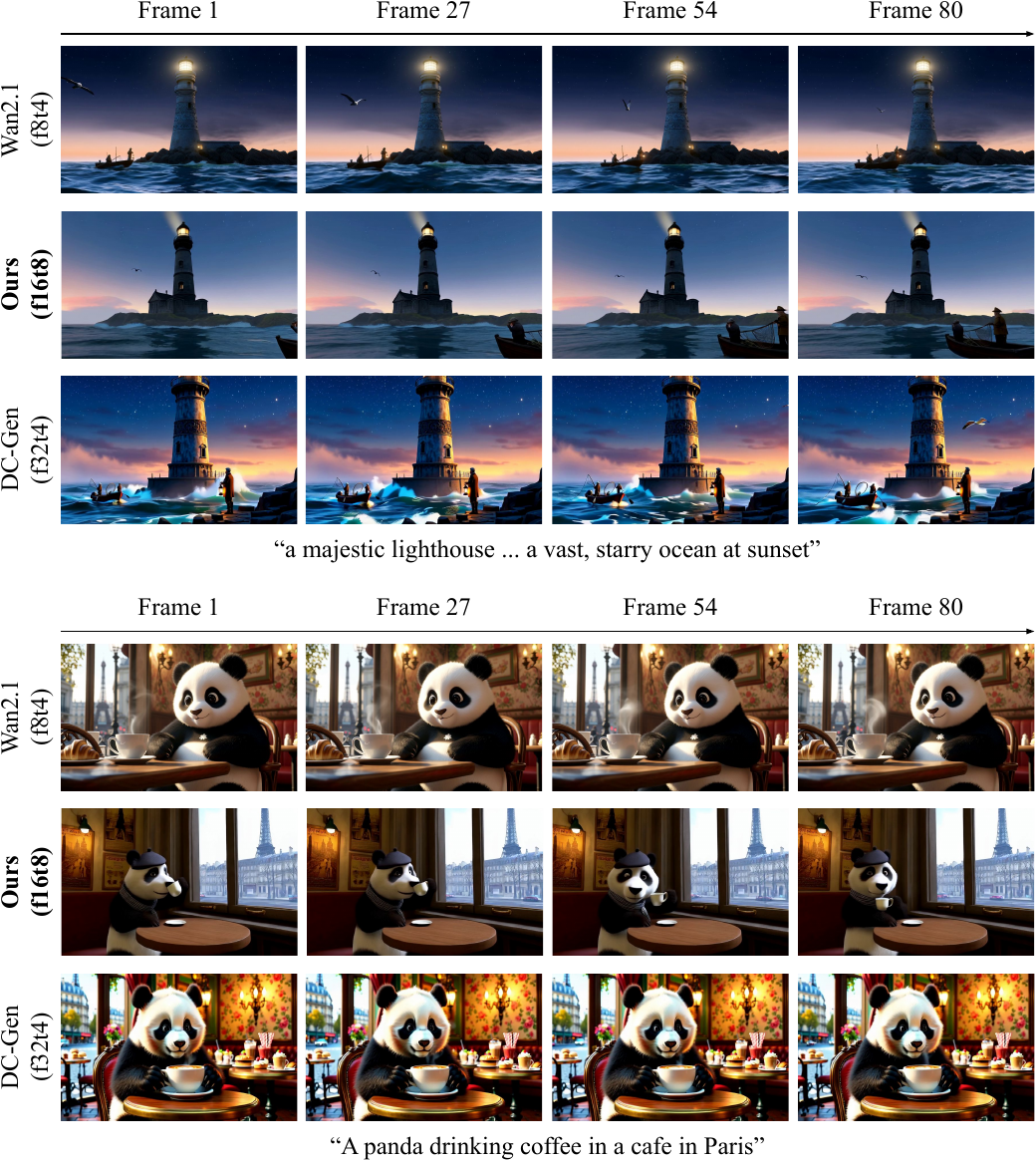}
  \caption{\textbf{Additional text-to-video results.}
    Samples from VBench-T2V~\citep{huang2023vbenchcomprehensivebenchmarksuite, huang2024vbenchcomprehensiveversatilebenchmark} at $480{\times}832{\times}81$, generated by the pretrained Wan2.1-14B~\citep{wan2025wanopenadvancedlargescale} at f8t4p2, ours at f16t8p2, and DC-Gen~\citep{he2026dc} at f32t4p1, from the same prompt.}
  \label{fig:gen_t2v_qual_supp_1}
\end{figure}
\vspace*{\fill}
\clearpage

%% file: supple/figures/18_gen_t2v_qual_2.tex
\clearpage
\vspace*{\fill}
\begin{figure}[H]
  \centering
  \includegraphics[width=\linewidth]{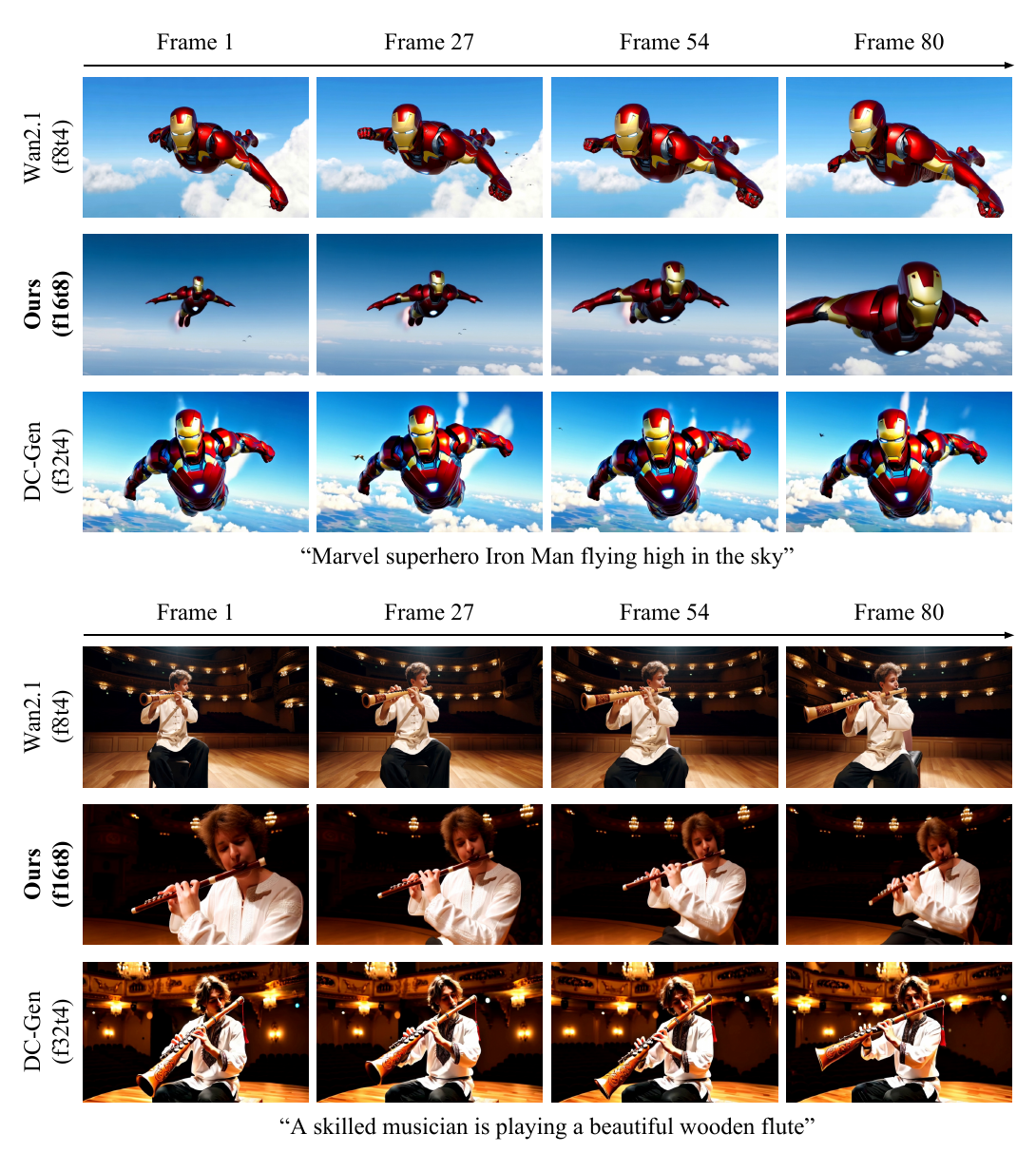}
  \caption{\textbf{Additional text-to-video results (continued).}
    Samples from VBench-T2V~\citep{huang2023vbenchcomprehensivebenchmarksuite, huang2024vbenchcomprehensiveversatilebenchmark} at $480{\times}832{\times}81$, generated by the pretrained Wan2.1-14B~\citep{wan2025wanopenadvancedlargescale} at f8t4p2, ours at f16t8p2, and DC-Gen~\citep{he2026dc} at f32t4p1, from the same prompt.}
  \label{fig:gen_t2v_qual_supp_2}
\end{figure}
\vspace*{\fill}
\clearpage

%% file: supple/figures/7_high_res_qual.tex
\clearpage
\vspace*{\fill}
\begin{figure}[H]
  \centering
  \includegraphics[width=\linewidth]{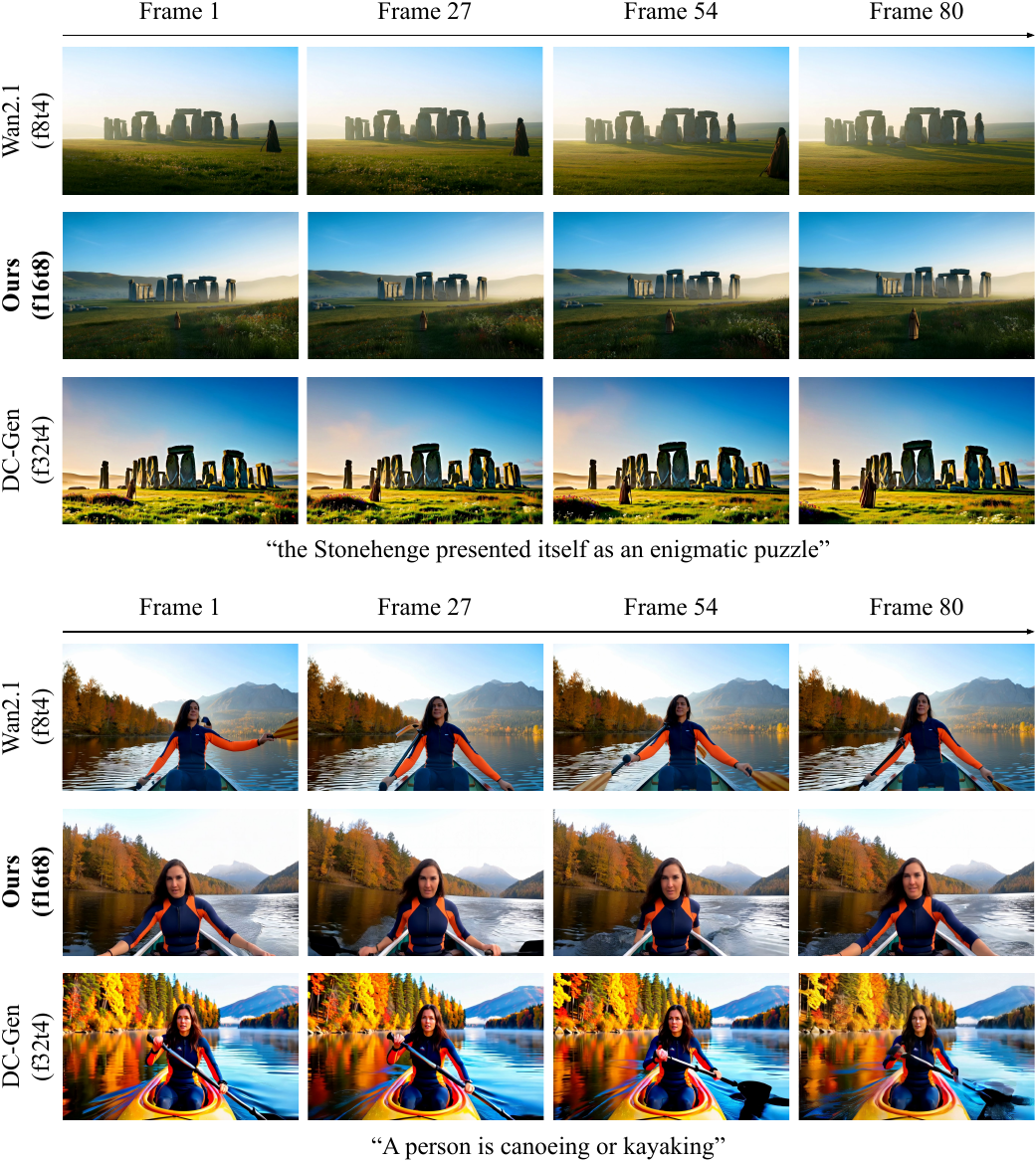}
  \caption{\textbf{Additional high-resolution text-to-video results.}
    Samples at $736{\times}1280{\times}81$, generated by the pretrained Wan2.1-14B~\citep{wan2025wanopenadvancedlargescale} at f8t4p2, ours at f16t8p2, and DC-Gen~\citep{he2026dc} at f32t4p1, from the same prompt.}
  \label{fig:highres_qual}
\end{figure}
\vspace*{\fill}
\clearpage

%% file: supple/figures/19_high_res_qual_2.tex
\clearpage
\vspace*{\fill}
\begin{figure}[H]
  \centering
  \includegraphics[width=\linewidth]{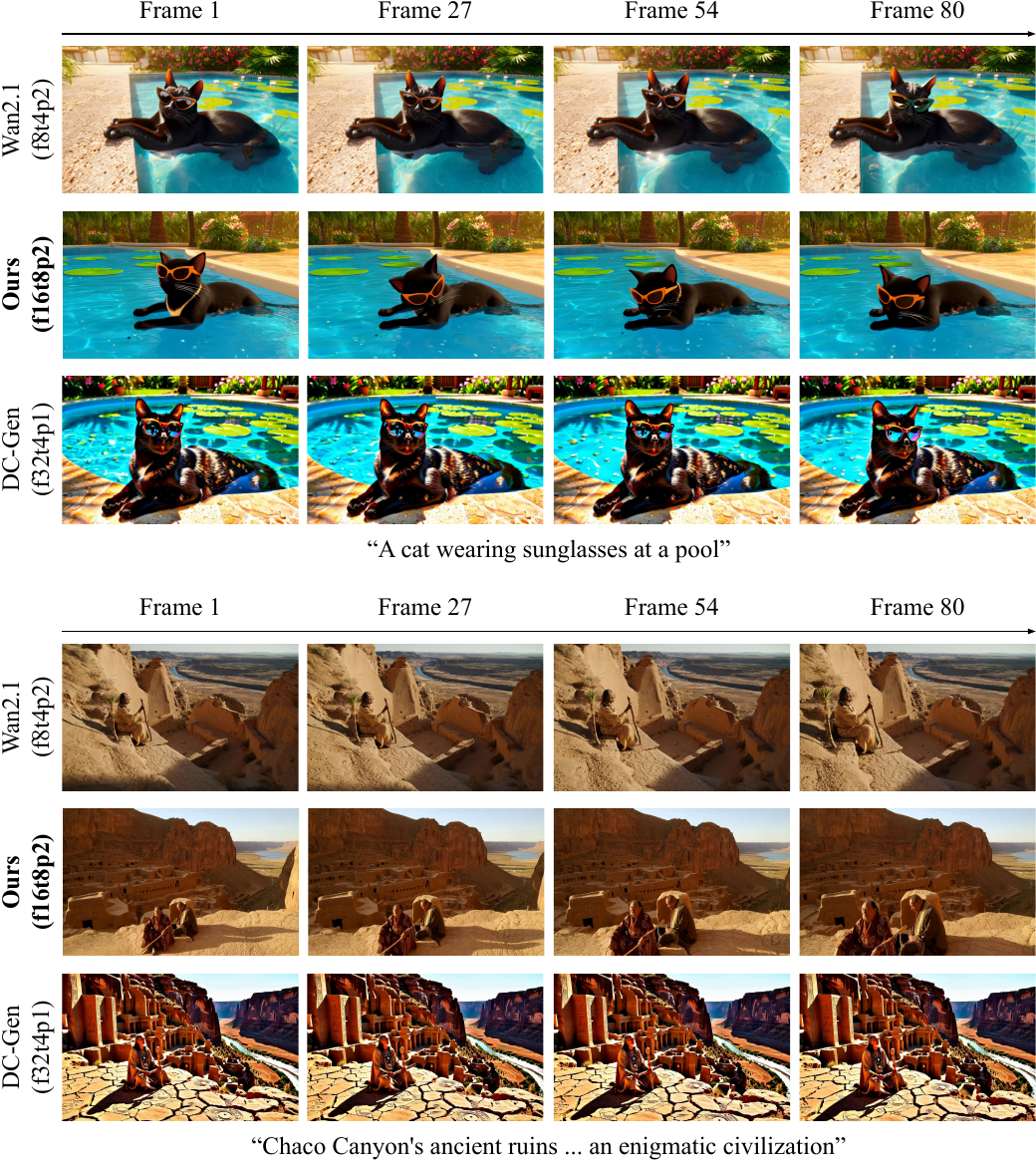}
  \caption{\textbf{Additional high-resolution text-to-video results (continued).}
    Samples at $736{\times}1280{\times}81$, generated by the pretrained Wan2.1-14B~\citep{wan2025wanopenadvancedlargescale} at f8t4p2, ours at f16t8p2, and DC-Gen~\citep{he2026dc} at f32t4p1, from the same prompt.}
  \label{fig:highres_qual_2}
\end{figure}
\vspace*{\fill}
\clearpage

%% file: supple/figures/20_i2v_ltx_comp_0.tex
\clearpage
\vspace*{\fill}
\begin{figure}[H]
  \centering
  \includegraphics[width=\linewidth]{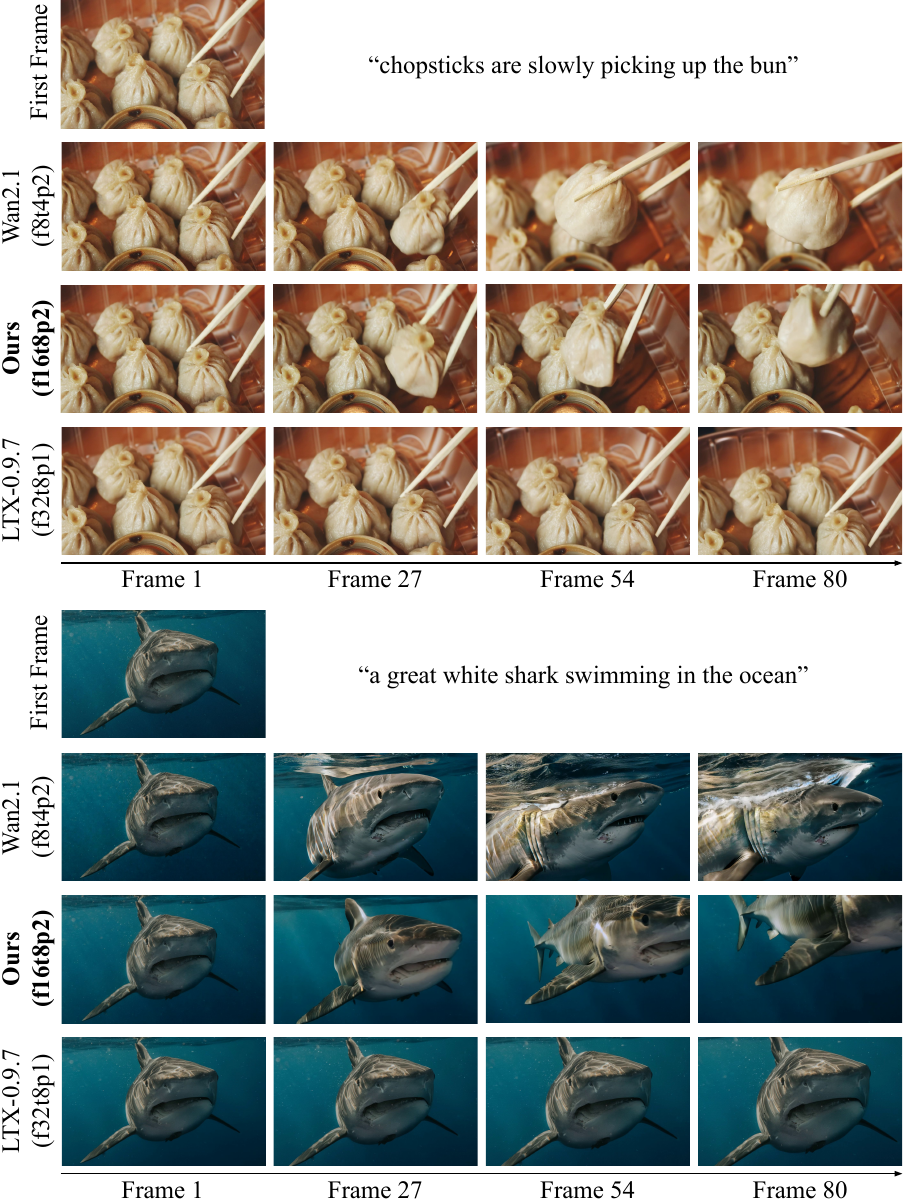}
  \caption{\textbf{Additional high-resolution image-to-video results.}
    Samples at $736{\times}1280{\times}81$, generated by the pretrained Wan2.1-14B~\citep{wan2025wanopenadvancedlargescale} at f8t4p2, ours at f16t8p2, and LTX-Video~\citep{hacohen2024ltxvideorealtimevideolatent} 0.9.7 at f32t8p1, from the same conditioning frame and prompt. The motion of LTX-Video tends to come from a global zoom or a slow camera movement over the conditioning frame, while the scene stays static.}
  \label{fig:i2v_ltx_comp_0}
\end{figure}
\vspace*{\fill}
\clearpage

%% file: supple/figures/21_i2v_ltx_comp_1.tex
\clearpage
\vspace*{\fill}
\begin{figure}[H]
  \centering
  \includegraphics[width=\linewidth]{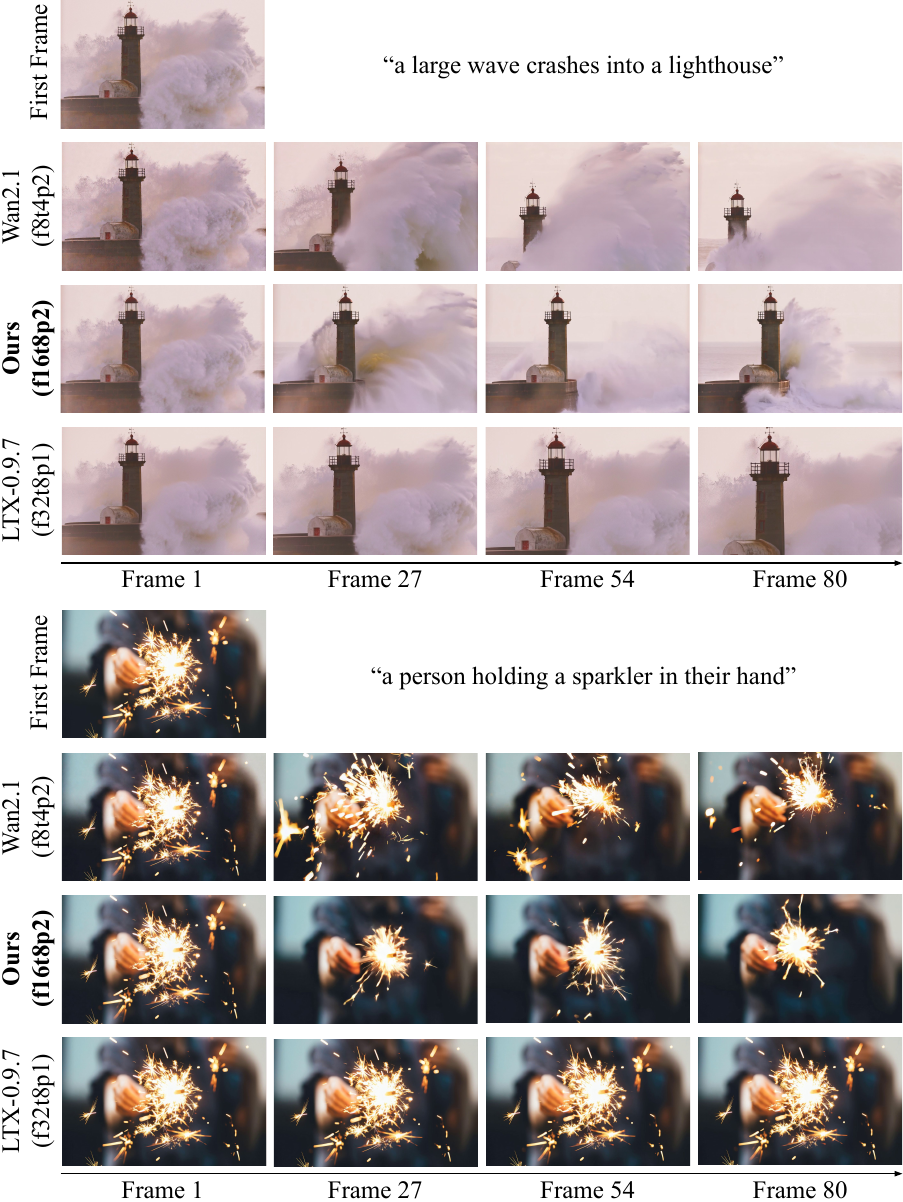}
  \caption{\textbf{Additional high-resolution image-to-video results (continued).}
    Samples at $736{\times}1280{\times}81$, generated by the pretrained Wan2.1-14B~\citep{wan2025wanopenadvancedlargescale} at f8t4p2, ours at f16t8p2, and LTX-Video~\citep{hacohen2024ltxvideorealtimevideolatent} 0.9.7 at f32t8p1, from the same conditioning frame and prompt. The motion of LTX-Video tends to come from a global zoom or a slow camera movement over the conditioning frame, while the scene stays static.}
  \label{fig:i2v_ltx_comp_1}
\end{figure}
\vspace*{\fill}
\clearpage

%% file: supple/figures/22_i2v_ltx_comp_2.tex
\clearpage
\vspace*{\fill}
\begin{figure}[H]
  \centering
  \includegraphics[width=\linewidth]{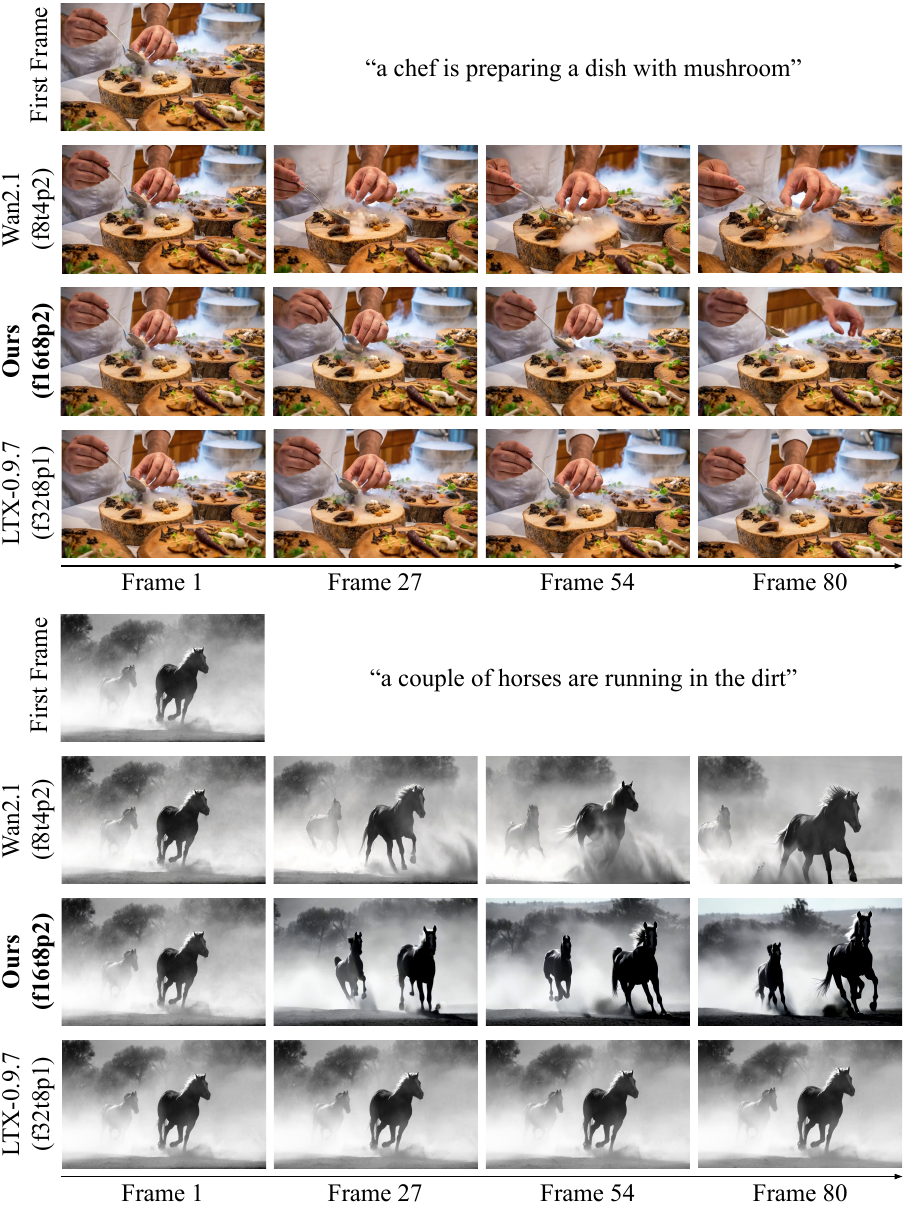}
  \caption{\textbf{Additional high-resolution image-to-video results (continued).}
    Samples at $736{\times}1280{\times}81$, generated by the pretrained Wan2.1-14B~\citep{wan2025wanopenadvancedlargescale} at f8t4p2, ours at f16t8p2, and LTX-Video~\citep{hacohen2024ltxvideorealtimevideolatent} 0.9.7 at f32t8p1, from the same conditioning frame and prompt. The motion of LTX-Video tends to come from a global zoom or a slow camera movement over the conditioning frame, while the scene stays static.}
  \label{fig:i2v_ltx_comp_2}
\end{figure}
\vspace*{\fill}
\clearpage

%% file: supple/figures/16_style_qual.tex
\clearpage
\vspace*{\fill}
\begin{figure}[H]
  \centering
  \includegraphics[width=\linewidth]{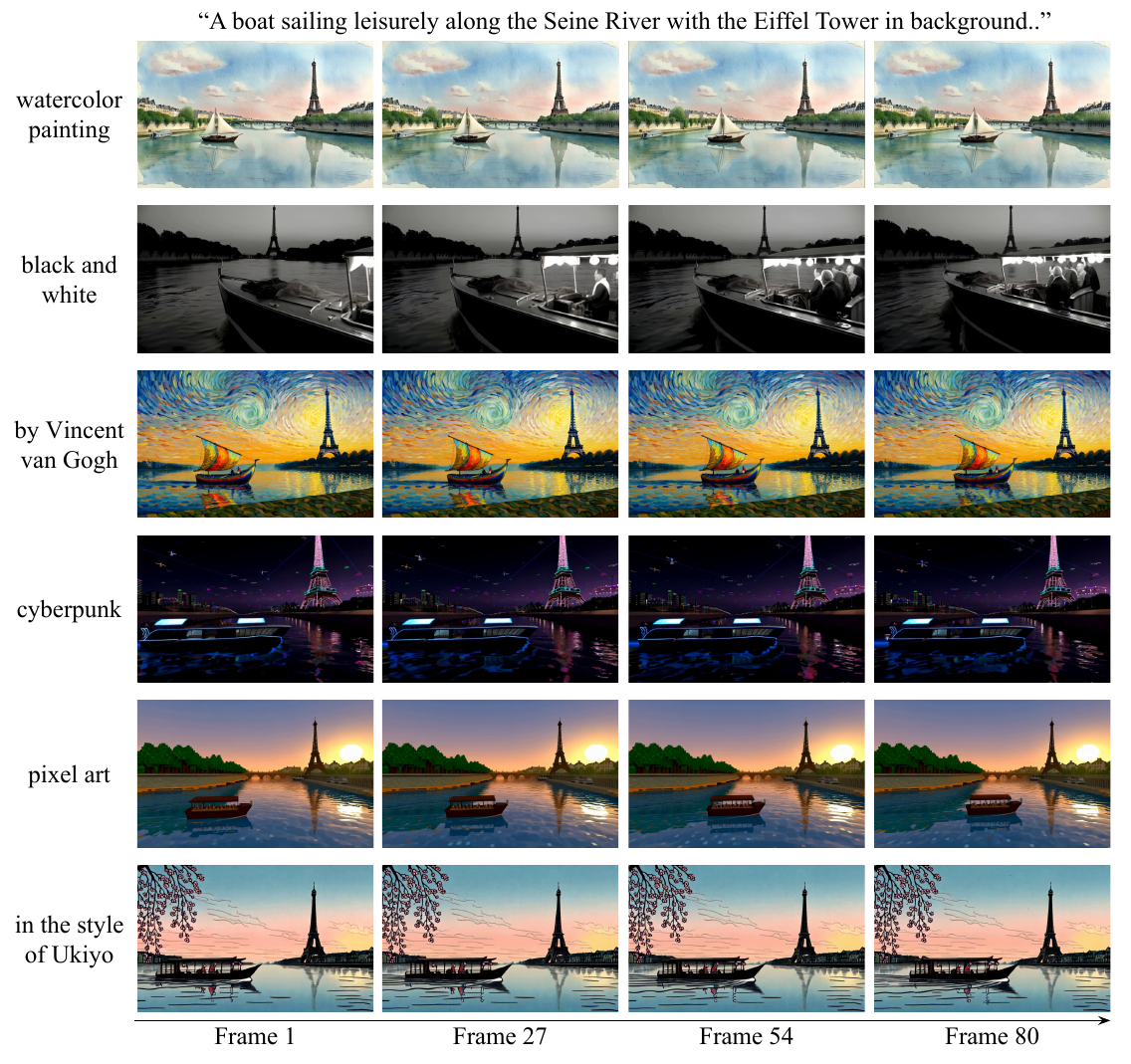}
  \caption{\textbf{Generation in various styles.}
    Text-to-video samples from ours at $736{\times}1280{\times}81$, generated from a single prompt with a different style suffix in each row.}
  \label{fig:style_qual}
\end{figure}
\vspace*{\fill}

%% file: supple/figures/9_pca_qual.tex
\begin{figure}[H]
\centering
\begin{minipage}{\linewidth}
  \centering
  \includegraphics[width=\linewidth]{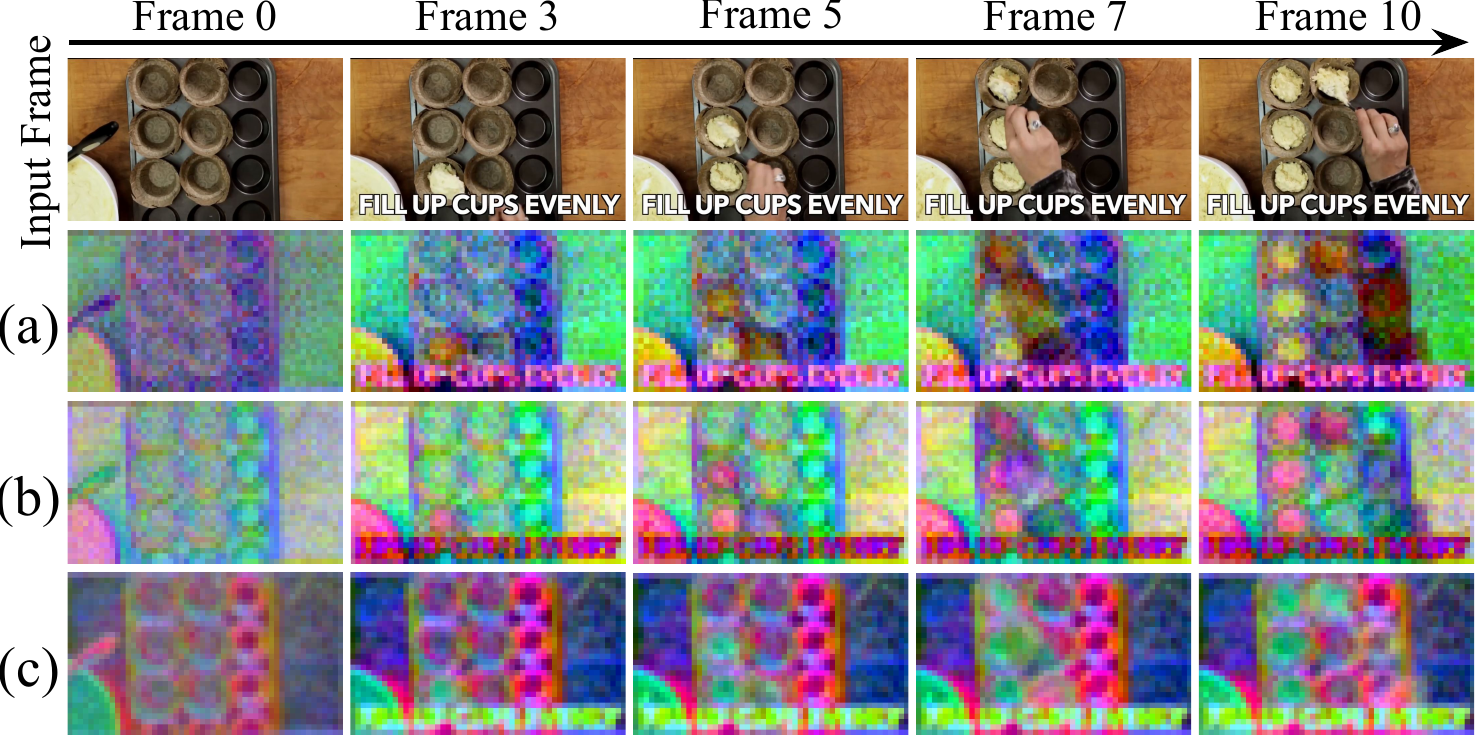}
\end{minipage}
 
\vspace{6pt}
 
\begin{minipage}{\linewidth}
  \centering
  \includegraphics[width=\linewidth]{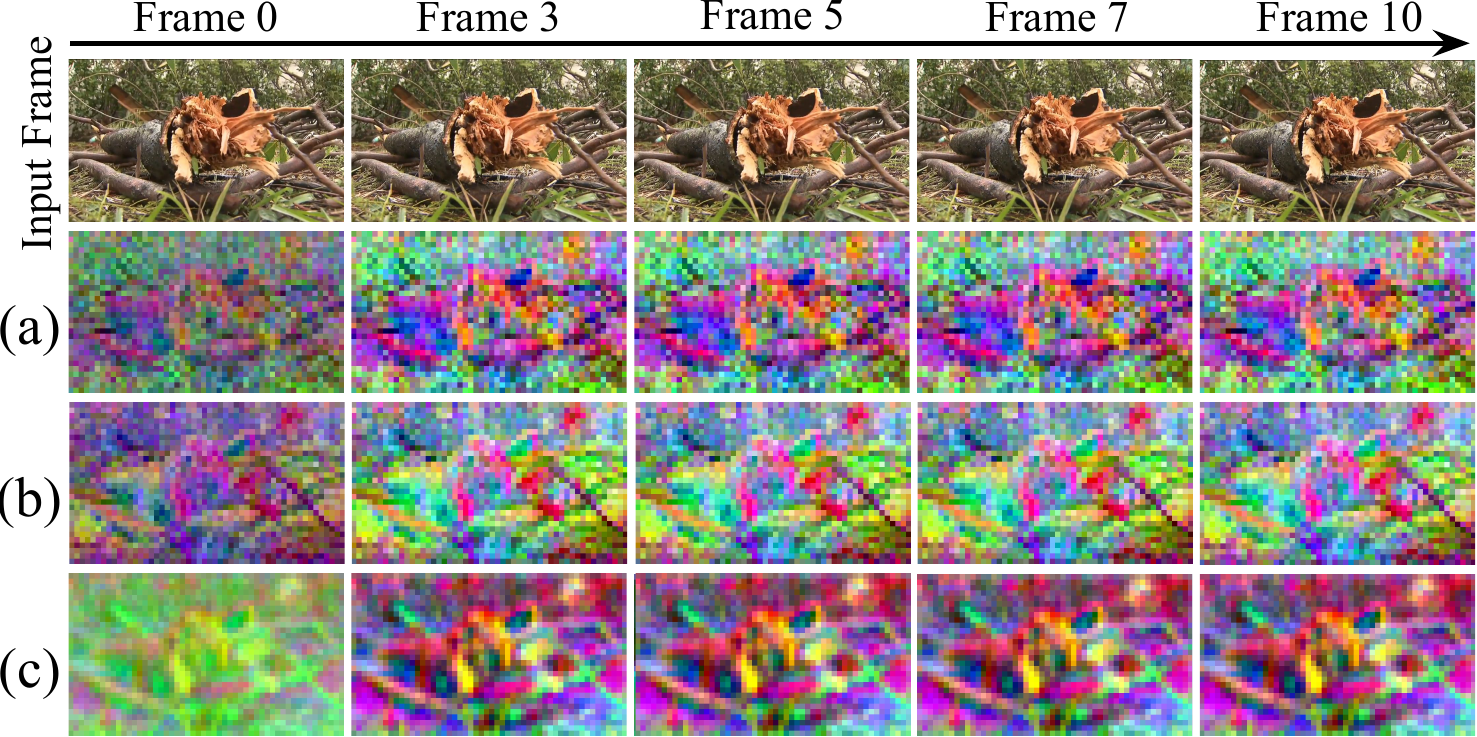}
\end{minipage}
 
\vspace{6pt}
 
\begin{minipage}{\linewidth}
  \centering
  \includegraphics[width=\linewidth]{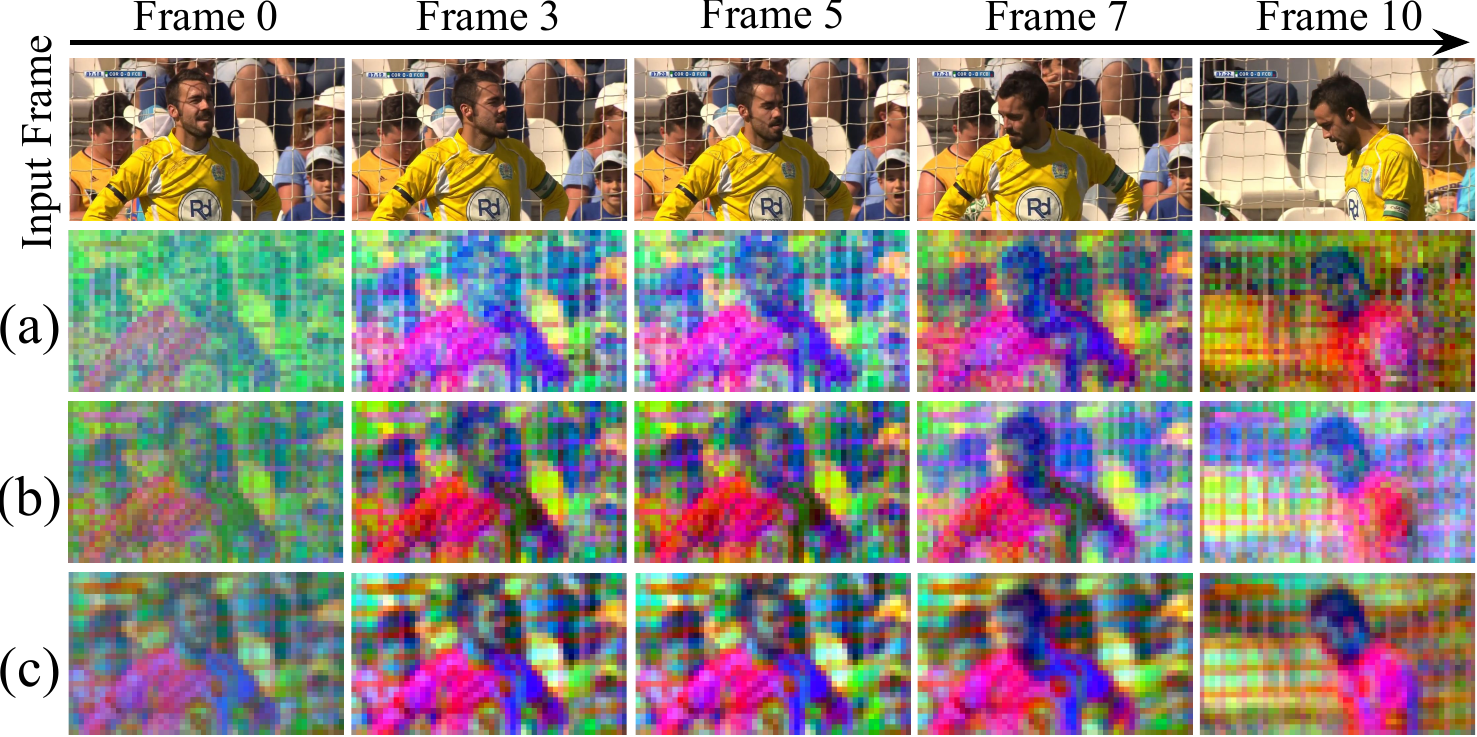}
\end{minipage}
 
\caption{\textbf{Additional latent PCA visualizations.}
    Principal components of the full $C{+}C'$ latent at f16t8p2, encoded from $480{\times}832{\times}81$ videos, for \textbf{(a)} a single encoder, \textbf{(b)} ours without $\mathcal{L}_\text{align}$, and \textbf{(c)} ours, with the input frames on top. The three leading components are computed per clip and mapped to RGB, so colors are not comparable across panels.}
\label{fig:pca_supp}
\end{figure}

%% file: supple/figures/9_pca_qual_1.tex
\begin{figure}[H]
\centering
\begin{minipage}{\linewidth}
  \centering
  \includegraphics[width=\linewidth]{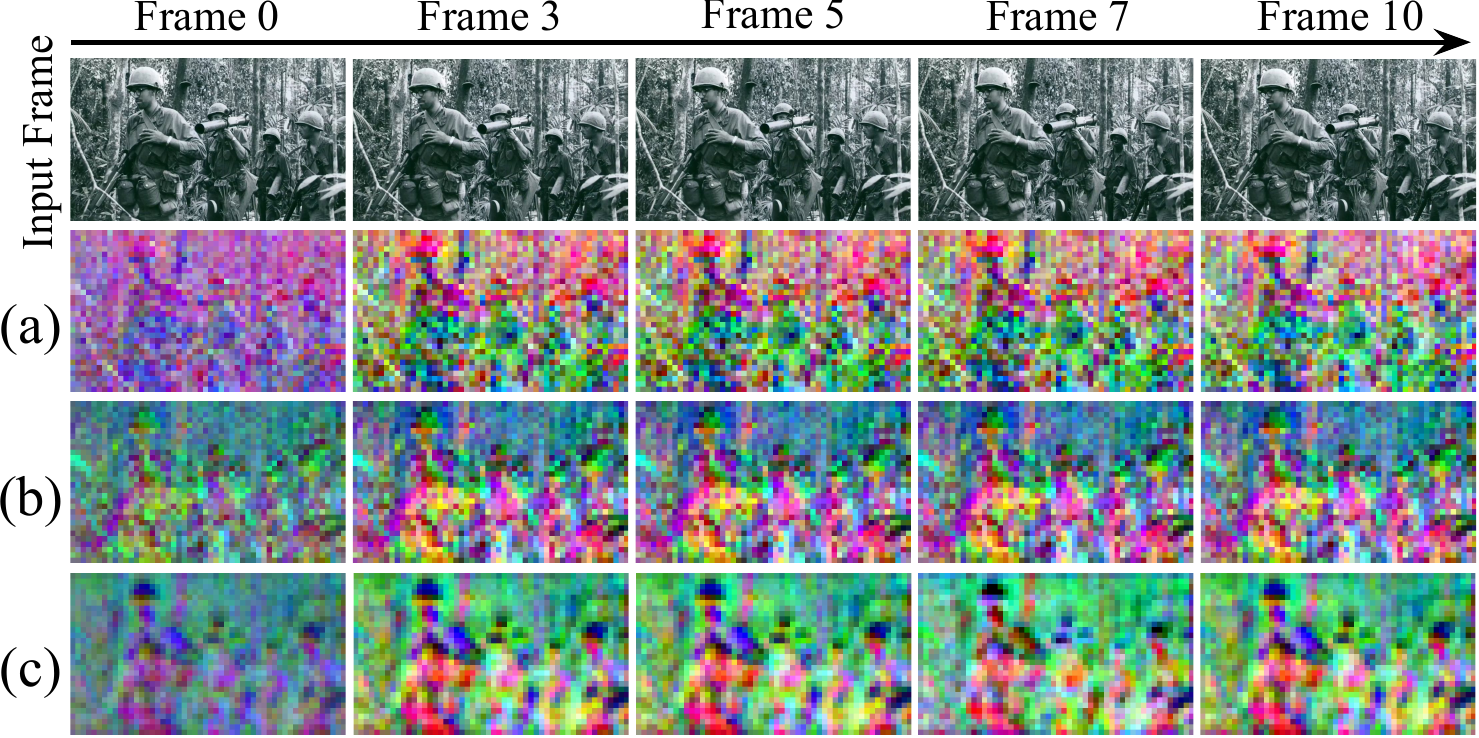}
\end{minipage}
 
\vspace{6pt}
 
\begin{minipage}{\linewidth}
  \centering
  \includegraphics[width=\linewidth]{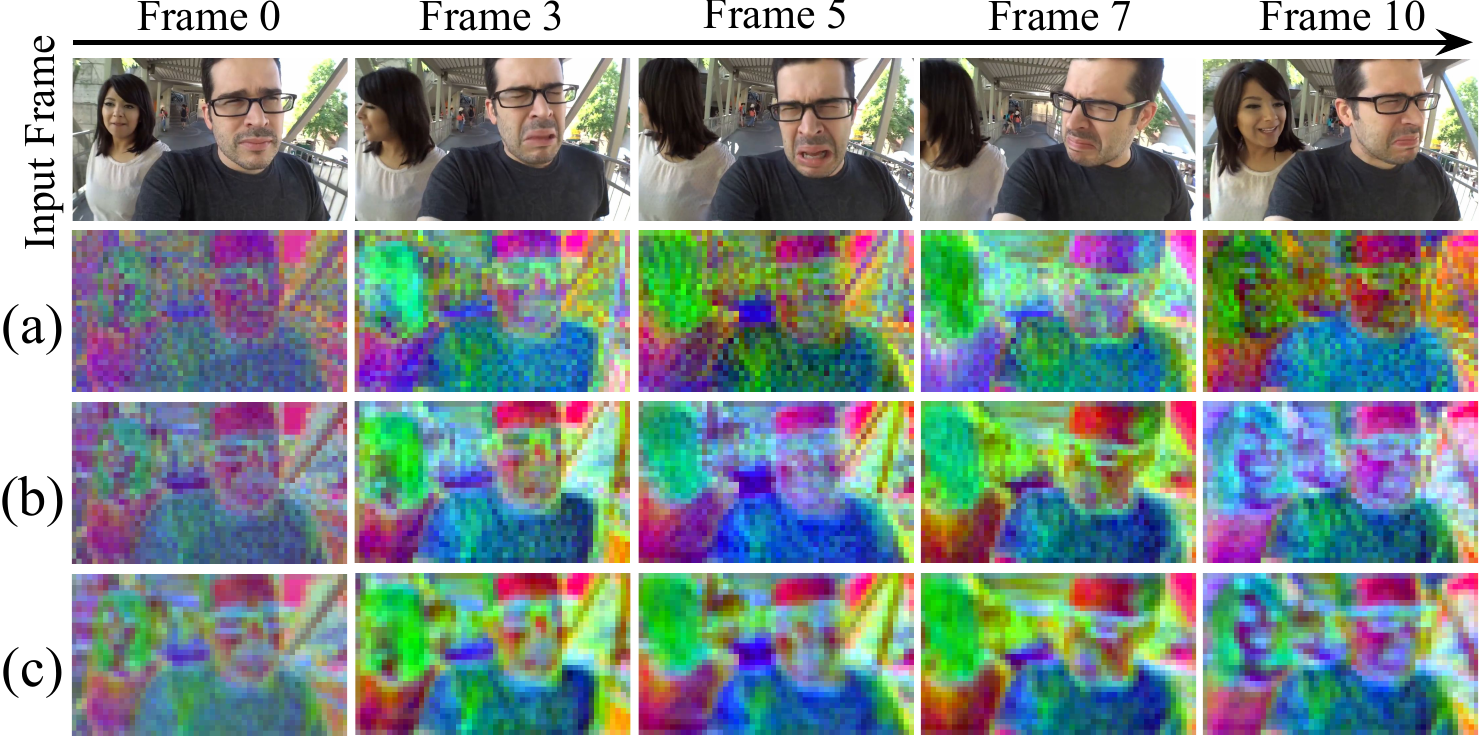}
\end{minipage}
 
\vspace{6pt}
 
\begin{minipage}{\linewidth}
  \centering
  \includegraphics[width=\linewidth]{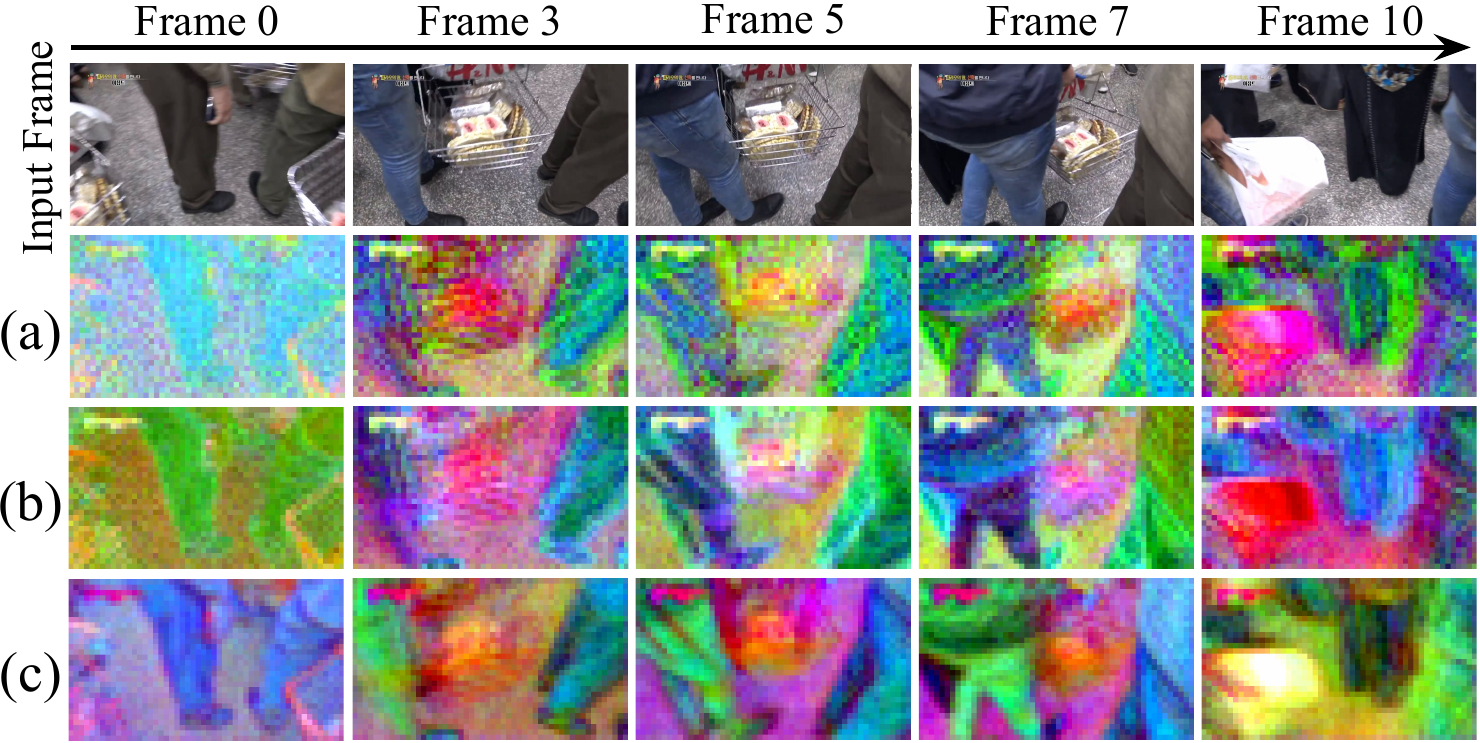}
\end{minipage}
 
\caption{\textbf{Additional latent PCA visualizations (continued).}
    Principal components of the full $C{+}C'$ latent at f16t8p2, encoded from $480{\times}832{\times}81$ videos, for \textbf{(a)} a single encoder, \textbf{(b)} ours without $\mathcal{L}_\text{align}$, and \textbf{(c)} ours, with the input frames on top. The three leading components are computed per clip and mapped to RGB, so colors are not comparable across panels.}
\label{fig:pca_supp_2}
\end{figure}